\documentclass[pdflatex,sn-basic,iicol]{sn-jnl}

\usepackage[dvipsnames]{xcolor}

\usepackage{graphicx}%
\usepackage{multirow}%
\usepackage{amsmath,amssymb,amsfonts}%
\usepackage{amsthm}%
\usepackage{mathrsfs}%
\usepackage[title]{appendix}%
\usepackage{textcomp}%
\usepackage{manyfoot}%
\usepackage{booktabs}%
\usepackage{algorithm}%
\usepackage{algorithmicx}%
\usepackage{algpseudocode}%
\usepackage{listings}%

\usepackage{arydshln}
\usepackage{subcaption}
\let\cline\cmidrule
\usepackage{amssymb}
\usepackage{pifont}
\newcommand{\cmark}{\ding{51}}%
\newcommand{\xmark}{\ding{55}}%
\usepackage{stmaryrd}
\usepackage{pdfpages}

\newcommand{\review}[1]{\textcolor{black}{#1}}
\newcommand{\red}[1]{\textcolor{Maroon}{#1}}
\newcommand{\green}[1]{\textcolor{OliveGreen}{#1}}

\theoremstyle{thmstyleone}%
\theoremstyle{thmstyletwo}%

\theoremstyle{thmstylethree}%

\begin{document}

\title[Self-Supervised Learning for Text Recognition]{Self-Supervised Learning for Text Recognition: A Critical Survey}


\author*[]{\fnm{Carlos} \sur{Penarrubia}}\email{carlos.penarrubia@ua.es}

\author[]{\fnm{Jose J.} \sur{Valero-Mas}}\email{jjvalero@dlsi.ua.es}
\equalcont{These authors contributed equally to this work.}

\author[]{\fnm{Jorge} \sur{Calvo-Zaragoza}}\email{jcalvo@dlsi.ua.es}
\equalcont{These authors contributed equally to this work.}

\affil[]{\orgdiv{Pattern Recognition and Artificial Intelligence Group}, \orgname{University of Alicante}, \orgaddress{\city{Alicante}, \country{Spain}}}


\abstract{Text Recognition (TR) refers to the research area that focuses on retrieving textual information from images, a topic that has seen significant advancements in the last decade due to the use of Deep Neural Networks (DNN). However, these solutions often necessitate vast amounts of manually labeled or synthetic data. Addressing this challenge, Self-Supervised Learning (SSL) has gained attention by utilizing large datasets of unlabeled data to train DNN, thereby generating meaningful and robust representations. Although SSL was initially overlooked in TR because of its unique characteristics, recent years have witnessed a surge in the development of SSL methods specifically for this field. This rapid development, however, has led to many methods being explored independently, without taking previous efforts in methodology or comparison into account, thereby hindering progress in the field of research. This paper, therefore, seeks to consolidate the use of SSL in the field of TR, offering a critical and comprehensive overview of the current state of the art. We will review and analyze the existing methods, compare their results, and highlight inconsistencies in the current literature. This thorough analysis aims to provide general insights into the field, propose standardizations, identify new research directions, and foster its proper development.}




\maketitle

\section{Introduction}
\label{sec:Introduction}
Text Recognition (TR) is the research area within computer vision that involves automatically recognizing the text present in a given image \citep{sharma2020character}. This field is crucial for automatically retrieving textual information from our environment through optical systems and digital image processing. Considering the nature of the images, TR is broadly divided into two main branches: Scene Text Recognition (STR) and Handwritten Text Recognition (HTR). STR focuses on recognizing text in natural scenes \citep{chen2021text}, such as logo texts \citep{li2022seetek,hubenthal2023image}, billboards \citep{yu2021english}, or shop fronts \citep{zhang2020street}, while HTR deals with recognizing text from handwritten documents, such as historical manuscripts \citep{narang2020ancient,alkendi2024advancements} and personal notes in electronic devices \citep{BezerraZanchettinToseeliPirlo:Book:2017,ghosh2022advances}.

\begin{figure*}[!t]
  \centering
  \includegraphics[width=\linewidth]{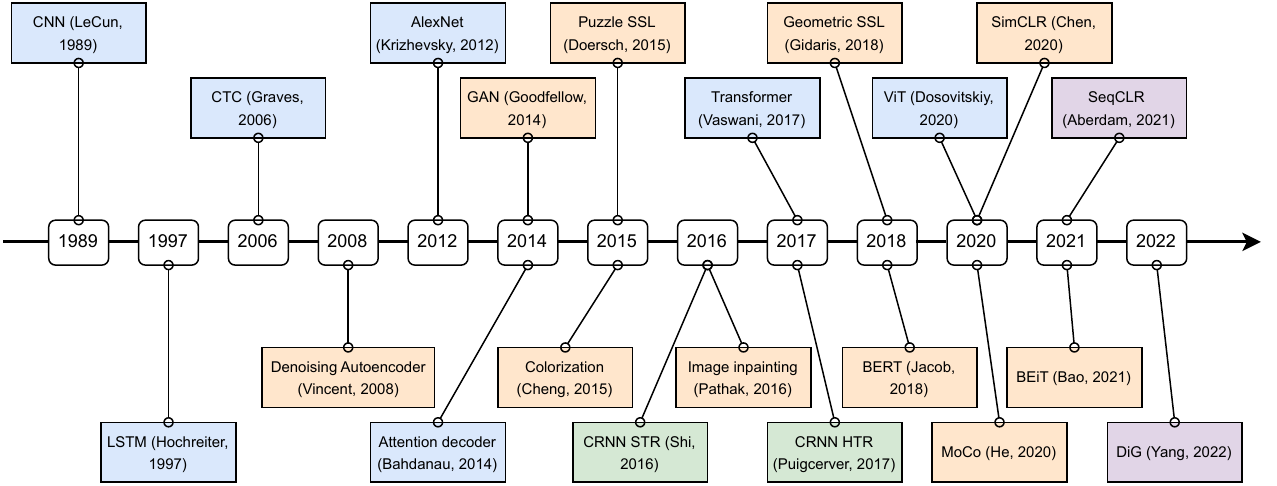}
    \caption{Timeline with relevant contributions within the context of this survey, including milestones related to Deep Learning (blue boxes), Text Recognition (green boxes), Self-Supervised Learning (orange boxes), and Self-Supervised Learning in Text Recognition (purple boxes).}
  \label{fig:time_line}
\end{figure*}

In the last decade, with the emergence of Deep Neural Networks (DNN) \citep{krizhevsky2012imagenet,dong2021survey}, there has been significant development and enhancement of capabilities in TR \citep{khan2021deep,long2021scene,ansari2022handwritten}. This increase has also been possible thanks to the creation of large labeled datasets \citep{nikitha2020handwritten}. However, obtaining manually labeled data requires significant resources and time \citep{nikolaidou2022survey}, considering that each specific domain typically requires several datasets for training and benchmarking \citep{zhou2022domain}. As an alternative, tools for synthetic data generation have been developed, but this kind of data is not as effective as the real one, since there is always a domain shift that limits the performance of models in real environments \citep{zhang2019sequence,kang2020unsupervised,baek2021if}.

To alleviate these problems, different solutions have emerged such as data augmentation \citep{luo2020learn,atienza2021data}, Semi-Supervised Learning \citep{liu2020semitext,gao2021semi,aberdam2022multimodal}, or Self-Supervised Learning (SSL), which constitutes the focus of this work. SSL represents an unsupervised learning paradigm that has gained much attention and popularity due to its ability to learn general and meaningful representations from unlabeled data \citep{ozbulak2023know}. This is achieved by employing what is termed a ``pretext task'', which consists of leveraging collections of unlabeled data to train DNN in a supervised manner \citep{balestriero2023cookbook}. This supervision is done by automatically creating labels for the originally unlabeled data \citep{jing2020self}.

Due to its remarkable success in the related field of natural language processing \citep{qiu2020pre}, SSL has been increasingly applied in the area of computer vision. There, the main research area has been that of image classification, where SSL has led to great advances in the recognition capabilities \citep{balestriero2023cookbook}. However, SSL took longer to be explored in TR mainly because of its particular formulation: the output is a sequence and not a single category, which prevents, or limits to some extent, the application of many SSL methods developed for image classification. This can be supported by the fact that, while for image classification there is a sustained development of methods from 2015 \citep{agrawal2015learning,wang2015unsupervised,doersch2015unsupervised,radford2015unsupervised}, the first SSL method for TR was proposed in 2021 \citep{aberdam2021sequence}, as shown in Fig. \ref{fig:time_line}.

\begin{figure}[!t]
  \centering
  \includegraphics[width=\linewidth]{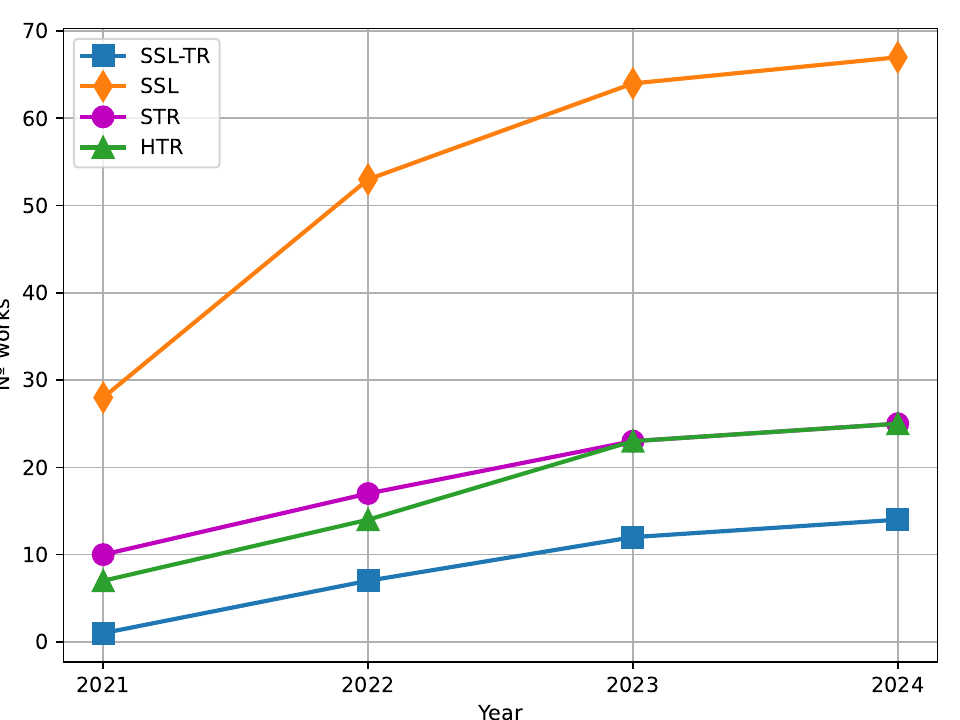}
  \caption{Cumulative number of methodological contributions for SSL, HTR, STR and SSL-TR (from the first SSL-TR method \citep{aberdam2021sequence}).}
  \label{fig:num_works}
\end{figure}

More recently, there has been an explosion of development of SSL for TR. As shown in Fig. \ref{fig:num_works}, since the first SSL-TR method, the development of new ones shows a similar pace to that of SSL, STR and HTR fields. However, this rapid development in an emerging field has caused many strategies to have been explored in parallel, without taking into account previous or simultaneous efforts, and making it difficult to establish an appropriate comparison to determine the state of the art of the field. Furthermore, recent surveys of both TR \citep{chen2021text,long2021scene,naiemi2022scene,gupta2022traditional,ansari2022handwritten,wang2023survey,alkendi2024advancements} and SSL \citep{liu2021self,albelwi2022survey,balestriero2023cookbook,ozbulak2023know} do not collect these, thereby preventing the research community to leverage the contributions and benefits demonstrated in works exploring SSL for TR.

This article arises from the need to compile and organize all the SSL methodologies devised for TR. We aim to summarize the development of this field, describing the main rationale behind each method, highlighting the strengths and weaknesses, and properly organizing the field within a modern SSL taxonomy. Furthermore, another key contribution of this work is that it compares the performance of all strategies and points out the existing inconsistencies from the lack of standardizations, exposing the differences and allowing a better understanding of these comparisons. These issues, accentuated by the rapid development, indicate that such protocols have not been properly established and may lead to partial or misleading conclusions. This analysis will serve to draw precise conclusions about SSL for TR, study solutions and standardizations for this emerging field, and identify new lines of research that address the needs discussed throughout the article. In short, the present work provides an integral and critical view of the current state of SSL for TR, facilitating a better understanding and encouraging the formal development of this field to create more robust and effective methods in the future.

The remainder of the work continues as follows: a general formulation of TR is provided in Section \ref{sec:Text_Recognition_foundations}, along with an outline of the most common DNN architectures considered for this purpose; the existing SSL methodologies for TR are categorized and explained in detail in Section \ref{sec:SSL_taxonomy_for_TR}; then, Section \ref{sec:benchmarking} introduces the different approaches for both training and evaluating these models; after that, Section \ref{sec:Results_comparison} compares and deeply analyzes the results obtained by the existing works in the related literature; finally, we conclude the work in Section \ref{sec:conclusions}, identifying current trends and future directions for SSL in TR.

\section{Fundamentals of Text Recognition}
\label{sec:Text_Recognition_foundations}

Before delving into the main use of SSL for TR, we first outline in this section the core principles behind TR approaches. More precisely, we will provide a formal definition of the task and describe the common DNN employed to address it.

\subsection{Problem Formulation}
Text recognition, whether for STR or HTR, involves decoding text images into their corresponding symbolic representations. Typically, these are sequences of characters (strings).

Formally, for an image $i \in \mathcal{I}$ with $\mathcal{I} \in \mathbb{R}^{c \times H \times W}$, where $c$ is the number of channels, $H$ is the height, and $W$ is the width, the goal is to retrieve the sequence $\mathbf{s} = [s_{1}, \ldots, s_{\left | \mathbf{s} \right |}] \in \Sigma^{*}$, with $\Sigma$ representing the \review{symbol alphabet and $\Sigma^{*}$ all its possible combinations}. Then, the recognition problem can be formalized as predicting the most probable string $\mathbf{\hat{s}}$ given the image $i$:

\begin{equation}
\label{eq:map}
    \mathbf{\hat{s}} = \underset{\mathbf{s}\in \Sigma^{*}}{\text{arg max}}~P(\mathbf{s} | i)
\end{equation}

Finding $\mathbf{\hat{s}}$ according to Eq.~\ref{eq:map} is known to be NP-hard \citep{de2014most}. Instead, practical solutions to TR approximate this probability. While historically addressed in different ways, the current state of the art for this challenge is defined by the so-called \textit{end-to-end} formulation, which does not break it down into several steps. Instead, there is a single model that directly decodes a string from a given input image \citep{shi2016end}. It must be noted, however, that TR pipelines often comprise pre-processing (e.g., image enhancement, distortion correction, etc.) and/or post-processing (e.g., rescoring based on language models) stages to improve overall performance. Nevertheless, these processes are not discussed in this survey as they are outside the scope of the SSL perspective.

To approximate Eq.~\ref{eq:map}, modern approaches for TR are mostly based on DNN, which learn from a dataset $\mathcal{D} \subset \mathcal{I} \times \Sigma^{*}$. The neural architectures considered state-of-the-art for this purpose are described below. 

\subsection{Neural architectures for TR}
In order to better understand how the reviewed SSL methods operate, it is necessary to know the basic approach of TR methods. In this section, we specifically focus on the set of approaches for TR that are considered in the SSL-TR literature. We refer the reader to STR and HTR literature for more specific architectures of each task \citep{ansari2022handwritten,wang2023survey}.

\begin{figure*}[!tbh]
  \centering
  \includegraphics[width=\linewidth]{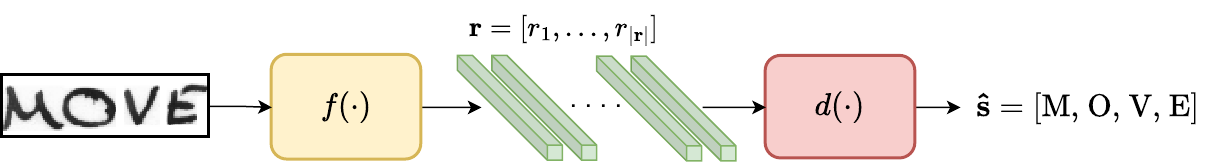}
  \caption{Encoder-decoder pipeline for TR. The encoder $f(\cdot)$ produces the sequence of features representations $\mathbf{r}$, while the decoder $d(\cdot)$ produces the sequence of predicted symbols $\mathbf{\hat{s}}$.}
  \label{fig:encoder_decoder}
\end{figure*}

Broadly speaking, the state of the art for TR considers the use of encoder-decoder architectures (see Fig. \ref{fig:encoder_decoder}). Next, we break down this architecture into the encoder and decoder modules.

\subsubsection{Encoder}
Given an input image $i$, the encoder $f(\cdot)$ is responsible for extracting the sequence of features $\mathbf{r} = \left [r_1, ..., r_{\left | \mathbf{r} \right |} \right ]$, where $r \in \mathbb{R}^T$ and $T$ is the feature embedding size. In the literature two main architectures are typically used: (i) those based on Convolutional Recurrent Neural Networks (CRNN) \citep{shi2016end,puigcerver2017multidimensional}; and (ii) those based on the Vision Transformers (ViT) \citep{dosovitskiy2020image}:

\begin{itemize}
    \item \textbf{CRNN}. In this type of encoder, a Convolutional Neural Network (CNN) visual feature extractor \citep{lecun1989backpropagation} $\text{cnn}: \mathbb{R}^{c\times H\times W}\rightarrow\mathbb{R}^{c^{\prime }\times H^{\prime}\times W^{\prime}}$ learns the visual characteristics of the images, where $c^{\prime}$ is the number of feature maps of size $H^{\prime}\times W^{\prime}$. Then, a map-to-sequence operation $\text{mts}:\mathbb{R}^{c^{\prime}\times H^{\prime}\times W^{\prime}}\rightarrow\mathbb{R}^{W^{\prime}\times T^{\prime}}$ reshapes the features into a sequence of so-called frames, \review{where $T^{\prime}$ is the feature dimension}. Finally, a Recurrent Neural Network (RNN) \citep{sherstinsky2020fundamentals} $\text{rnn}: \mathbb{R}^{W^{\prime}\times T^{\prime}}\rightarrow\mathbb{R}^{W^{\prime}\times T}$, where $T$ is the output feature size, interprets the sequence of frames, leading to the final representation $\mathbf{r} = \left [r_1, \ldots, r_{W^{\prime}}  \right ]$. In this case, the encoder can be expressed as:

    \begin{equation}
        f(\cdot) = \text{rnn}(\text{mts}(\text{cnn}(\cdot)))
        \label{eq:crnn_encoder}
    \end{equation}

    \item \textbf{ViT}. This encoder is based on the Transformer architecture \citep{vaswani2017attention}. It splits the input image into a sequence of non-overlapping patches of size $H^{\prime}\times W^{\prime}$, then linearly embeds and adds a positional embedding to each, leading to $\mathbf{p_0} = \left [ p_{0_{1}}, \ldots, p_{0_{n}} \right ] $, where $n=(W \cdot H)/(W^{\prime} \cdot H^{\prime})$ and $p \in \mathbb{R}^T$. \review{Note that, depending on the implementation, the positional encoding can be both 1D or 2D. After,} these embedded patches are processed by $L$ Transformer \review{encoder} blocks. Each block consists of a Normalization Layer (NL) \citep{ba2016layer}, Multi-Head Self-Attention (MSA), Multi-Layer Perceptron (MLP), and residual connections \citep{he2016deep}. Then, the final representation $\mathbf{r} = \left [r_1, \ldots, r_n  \right ]$ is computed as:
    
    \begin{equation}
    \begin{split}
        \mathbf{a}_{l} = \text{MSA}(\text{NL}(\mathbf{p}_{l-1})) + \mathbf{p}_{l-1}, \quad l=1,\ldots, L \\ \mathbf{p}_{l} = \text{MLP}(\text{NL}(\mathbf{a}_{l})) + \mathbf{a}_{l},\quad l=1, \ldots, L \\ \mathbf{r} = \mathbf{p}_L
    \end{split}
    \label{eq:vit}
    \end{equation}

    \item \textbf{\review{Hybrid encoder}}. \review{This type of encoder combines CNN with Transformer encoder. In this case, similar to a CRNN, the image is first processed by a CNN which extracts the feature map. After adding the positional encoding, a flatten operation is performed $\text{fl}:\mathbb{R}^{c^{\prime}\times H^{\prime}\times W^{\prime}}\rightarrow\mathbb{R}^{\left( H^{\prime}\cdot W^{\prime}\right)\times c^{\prime}}$, obtaining a sequence of features which will be processed by $L$ Transformer encoder blocks, leading to the final representation $\mathbf{r} = \left [r_1, \ldots, r_{H^{\prime}\cdot W^{\prime}} \right ]$.}
\end{itemize}

\subsubsection{Decoder}
As depicted in Fig. \ref{fig:encoder_decoder}, given a sequence of feature representation $\mathbf{r}$, the decoder $d(\cdot)$ must retrieve the sequence of \review{symbols} $\mathbf{s} = [s_{1}, \ldots, s_{\left | \mathbf{s} \right |}] \in \Sigma^{*}$. In the literature, there are three main decoders: that based on the Connectionist Temporal Classification (CTC) function \citep{graves2006connectionist}, the Attention decoder (Att) \citep{bahdanau2014neural,shi2016robust,cheng2017focusing}, and the Transformer Decoder (TD) \citep{vaswani2017attention}. Typcally, CTC and Att are the common decoders for the CRNN encoder, with TD more typically employed with ViT. However, there have been recent works that hybridize these architectures \citep{lee2020recognizing,diaz2021rethinking}, showing the effectiveness of the different combinations. Each of these decoders works as follows:

\begin{itemize}
    \item \textbf{CTC}. Although in the literature CTC is often referred to as a decoder, strictly speaking it is a training procedure \citep{graves2006connectionist}. CTC addresses the challenge of aligning variable-length input sequences ($\mathbf{r}$) with variable-length output sequences ($\mathbf{s}$), without requiring explicit alignment information during training. This is achieved by allowing the network to output a special ``\textit{blank}'' symbol along with the other symbols---i.e. ${\Sigma}' = \Sigma \cup \left \{ \textit{blank} \right \}$---subsequently enabling the production of sequences with repeated symbols and arbitrary lengths. 
    During prediction, CTC is no longer used. Instead, a certain heuristic is followed to produce the actual output sequence $\mathbf{\hat{s}}$. For instance, the greedy policy selects the most \review{probable symbol} of each frame, then merging consecutive repeated symbols and removing all ``\textit{blank}'' symbols, leading to $\mathbf{\hat{s}}$.

    \item \textbf{Att}. This decoder is autoregressive, i.e., it produces the next token $\hat{s}_j$ iteratively taking as input the encoded representations $\mathbf{r}$ from the input sequence and the previous predictions $\mathbf{ \hat{s}}=[\hat{s}_1, \ldots, \hat{s}_{j-1}]$. At each step, the decoder computes attention scores between its current state and the input representations, yielding a context vector that captures the most relevant information from the input. This context vector, along with the decoder's current state, guides the generation of the next token. To implement this type of decoder a Long Short-Term Memory (LSTM) cell \citep{hochreiter1997long} is typically used.
    
    \item \textbf{TD}. \review{The main feature of this architecture is that, apart from containing the same parts of a Transformer encoder, it also contains the cross-attention module, which computes the cross-attention between two input sequences. Initially, this architecture was conceived to perform autoregressive decoding based on the encoded input feature sequence $\mathbf{r}$ and the previous generated sequence prediction $\mathbf{\hat{s}}=[\hat{s}_1, \ldots, \hat{s}_{j-1}]$. However, Non-Autoregressive Transformer decoders \citep{chen2020non} are being currently developed. These decoders predict the entire sequence at the same time based on $\mathbf{r}$ and a character position vector, which encodes positional information ensuring that the decoder knows the relative or absolute position of each token to be predicted. Nevertheless, in the context of SSL-TR this last approach has not been used.}
    
\end{itemize}

\section{Taxonomy of SSL methodologies for TR}
\label{sec:SSL_taxonomy_for_TR}

In the field of SSL, the methods can be broadly categorized into two general approaches \citep{ozbulak2023know}: 
\begin{enumerate}
    \item \textbf{Discriminative:} Paradigm that aims to obtain the meaningful representations by discriminating between different possible categories associated to an input \citep{smys2020survey}. Note that, since in SSL there is no annotated data, the supervision signal comes from the data itself, proposing different pretext tasks for the encoder to solve.
    \item \textbf{Generative:} Family of approaches that consists in learning a generative distribution from the given data collection. By means of this task, the model learns fundamental and characteristic patterns and structures of the data, thus achieving meaningful embeddings \citep{harshvardhan2020comprehensive}.
\end{enumerate}

The rest of this section describes, in a detailed manner, the different categories of SSL and explains, in chronological order, the different existing proposals for TR. For each method, we will emphasize the problems that the respective authors detected in the state of the art, how they proposed to solve them, and the conclusions drawn. A summary of the strategies to be reviewed is provided in Table \ref{tab:methods_summary}. Note that, since the development of SSL in the field of TR is still recent, not all subcategories of general SSL methods have been explored. These missing categories will be explicitly discussed in Section \ref{sec:conclusions}.

\begin{table*}[!h]
\renewcommand{\arraystretch}{1.15}
\centering
\caption{List of the methods reviewed in this survey, organized by SSL category and working principle. Methods that belong to the same set are ordered chronologically.}
\label{tab:methods_summary}
\begin{tabular}{lllll}
\toprule[1pt]
 & \multicolumn{3}{l}{\textbf{Discriminative}} &  \\
 &  & \multicolumn{2}{l}{\textit{Contrastive}} &  \\
 &  &  & Sequence-to-sequence Contrastive Learning (SeqCLR) \citep{aberdam2021sequence} &  \\
 &  &  & Perceiving stroke-semantic context (PerSec) \citep{liu2022perceiving} &  \\
 &  &  & Scene Text Recognition with Contrastive Predictive Coding (STR-CPC) \citep{jiang2022scene} &  \\
 &  &  & Character Contrastive Learning (ChaCo) \citep{zhang2022chaco} &  \\
 &  &  & Character Movement Task to assist Contrastive Learning (CMT-Co) \citep{zhang2022cmt} &  \\
 &  &  & Relational Contrastive Learning for Scene Text Recognition (RCLSTR) \citep{zhang2023relational} &  \\
 &  & \multicolumn{2}{l} {\textit{Geometric}} &  \\
 &  &  & Flip \citep{penarrubia2024spatial} &  \\
 &  & \multicolumn{2}{l}{\textit{Puzzle}} &  \\
 &  &  & Sorting \citep{penarrubia2024spatial} &  \\
 &  & \multicolumn{2}{l}{\textit{Distillation}} &  \\
 &  &  & Character-to-Character Distillation (CCD) \citep{guan2023self} &  \\ \midrule
 & \multicolumn{3}{l}{\textbf{Generative}} &  \\
 &  & \multicolumn{2}{l}{\textit{Colorization}} &  \\
 &  &  & Similarity-Aware Normalization (SimAN) \citep{luo2022siman} &  \\
 &  & \multicolumn{2}{l}{\textit{Masked Image Modeling}} &  \\
 &  &  & Text-Degradation Invariant Autoencoder (Text-DIAE) \citep{souibgui2023text} &  \\
 &  &  & Dual Masked Autoencoder (Dual-MAE) \citep{qiao2023decoupling} &  \\
 &  &  & MaskOCR \citep{lyu2023maskocr} &  \\ \midrule
 & \multicolumn{3}{l}{\textbf{Hybrid}} &  \\
 &  &  & Discriminative and Generative (DiG) \citep{yang2022reading} &  \\
 &  &  & Symmetric Superimposition Modeling (SSM) \citep{gao2024self} &  \\
\bottomrule[1pt]
\end{tabular}
\end{table*}

\subsection{Discriminative approaches}
Despite leveraging the same underlying assumption, discriminative SSL can be further divided into the following subcategories according to their working principle \citep{ozbulak2023know}: contrastive learning, geometric and puzzle solvers, distillation, clustering, and information maximization. As aforementioned, not all this principles have been considered in the task of TR. Below, we describe, in chronological order of publication, those subcategories for which SSL methods have been posed for this domain.

\subsubsection{Contrastive learning}

The contrastive learning framework aims to pull closer embeddings of similar samples and push away those of dissimilar elements inside the representation space \citep{jaiswal2020survey}. This learning paradigm dates back to the early works of \cite{bromley1993signature,hadsell2006dimensionality}. These works, by employing siamese neural networks and labeled data, enforce representations of negative pairs to exceed a distance threshold between them. Since then, this idea has been considered in other works \citep{valero2024overview}.

In SSL, where labeled data is not used, the models are trained using the Information Noise Contrastive Estimation (InfoNCE) loss \citep{oord2018representation}, which is a modification of the NCE loss \citep{gutmann2010noise}, defined as:

\begin{equation}
\begin{split}
    & \text{InfoNCE}_{z, z^+, \{z^-\}}  = \\ & -\log\left ( \frac{\text{exp}(\text{sim}(z, z^+)/\tau)}{\sum_{z^-}\text{exp}(\text{sim}(z, z^-)/\tau)} \right )
\end{split}
\end{equation}
\noindent where $z$ denotes the representation of a sample, $z^+$ one positive (similar) pair, $\{z^-\}$ a set of negative (dissimilar) samples, $\tau$ a temperature hyperparameter, and $\text{sim}(\cdot, \cdot)$ is a similarity function (typically, dot product or cosine similarity). This loss function was proposed for the first time in the Contrastive Predictive Coding (CPC) method \citep{oord2018representation}, which is developed for audio and uses InfoNCE loss to discern the prediction of future latent representation of the same signal from those coming from different signals.

\begin{figure}[!tbh]
  \centering
  \includegraphics[width=\linewidth]{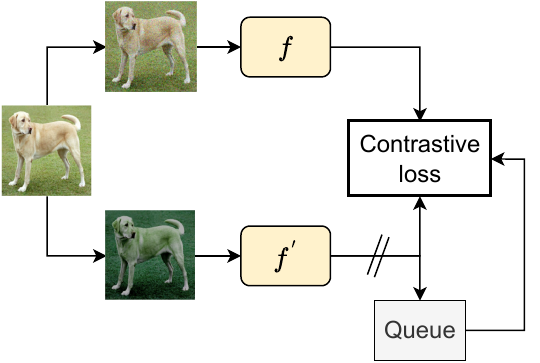}
  \caption{Example of a common contrastive learning framework (MoCo \citep{he2020momentum}) where two augmented views from the same image are created and the negative views are taken from a collected queue.}
  \label{fig:Contrastive}
\end{figure}

In the field of computer vision, \review{particularly in image classification, positive pairs consist of different views of the same image created by stochastic data augmentation, while negative pairs belong to past views of other images}. As shown in Fig. \ref{fig:Contrastive}, this idea was exploited by the general Momentum Contrast (MoCo) method \citep{he2020momentum}, which employs a teacher-student architecture and a queue to collect negative pairs. Note that, teacher-student refers to a dual-branch architecture in which the student side is updated via backpropagation and the teacher branch directly draws the parameters from the student using an exponential moving average.

Later, SimCLR \citep{chen2020simple} proposed the use of a siamese network, a projection head that performs nonlinear transformations on the encoded features, the batch itself as the set of negative samples, and the cosine distance as similarity measure. Authors demonstrated that the projection head heavily helps the model to extract higher-level abstract representations that capture complex patterns in the input data \citep{gupta2022understanding,bordes2023surprisingly}. Actually, the idea of the projection head is later incorporated in an improved version of the MoCo method, known as MoCo v2 \citep{chen2020improved}.

It is also worth highlighting the works of \cite{wei2020co2,zheng2021ressl}, termed as Relational Contrastive Learning which, in addition to InfoNCE, they use the Kullback-Leibler (KL) divergence to capture higher-order relationships among multiple entities in data. Mathematically, the KL divergence is a measure of how a probability distribution $P$ diverges from a second reference probability distribution $Q$, and is computed as:

\begin{equation}
    D_{\text{KL}}(P||Q) = \sum_x P(x) \text{log}\left(\frac{P(x)}{Q(x)}\right)
\end{equation}

While the aforementioned ideas were applied for CNN-based encoders, MoCo v3 \citep{chen2021empirical} adapts previous contrastive learning frameworks---i.e. MoCo and MoCo v2---for ViT encoders, also incorporating the query head proposed by \cite{grill2020bootstrap} to the framework. This is a head that only exists in the student branch and performs nonlinear transformations, which leads to better representations.

\begin{figure}[!tbh]
  \centering
  \includegraphics[width=\linewidth]{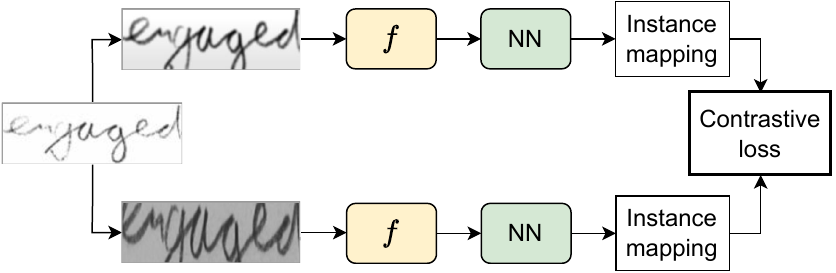}
  \caption{Graphical overview of SeqCLR \citep{aberdam2021sequence}. Two different augmentations are performed for a given image. The views are embedded by the encoder $f$ and a NN projection head. Once the instance mapping function groups frames into instances, the contrastive loss is applied.}
  \label{fig:seqclr_diagram}
\end{figure}

Within the previous context, the first method that applied SSL to TR is \emph{Sequence-to-sequence Contrastive Learning} (\textbf{SeqCLR}) \citep{aberdam2021sequence}, which adapts the idea of SimCLR for pre-training CRNN architectures. Unlike in image classification, this work considers that the input image cannot be treated as an atomic input, but the sequential nature of TR must be taken into account. Through an instance mapping function, SeqCLR establishes the instance level as the unit to perform the contrastive loss, which can be a frame (frame level), an average of several frames (subword level), or the average of all frames (word level) produced by a CRNN encoder. Therefore, the contrastive loss is applied at the considered instance level, as shown in Fig. \ref{fig:seqclr_diagram}. In addition, they also investigate the use of different types of projection heads (MLP, LSTM, or none) and the influence of horizontal cropping inside contrastive learning.

SeqCLR demonstrated for the first time that contrastive learning is applicable to the field of TR. In more specific terms, authors showed that the best results are obtained when considering subwords as instances, yet there is no clear trend in the rest of the setting (projection head or decoder). They also devised that the use of aggressive horizontal croppings can lead to misalignment problems between positive and negative pairs that perturbs the contrastive learning.

\begin{figure}[!tbh]
  \centering
  \includegraphics[width=\linewidth]{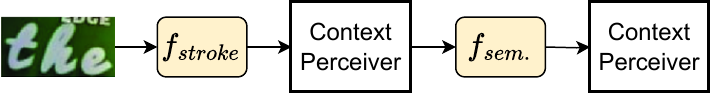}
  \caption{Graphical overview of PerSec \citep{liu2022perceiving}. The entire encoder is composed of $f_{\text{stroke}}$ and $f_{\text{sem}}$. Then, PerSec applies its methodology to both levels of representation.}
  \label{fig:PerSec_diagram}
\end{figure}

Subsequently, \emph{Perceiving stroke-semantic context} (\textbf{PerSec}) \citep{liu2022perceiving} identifies three fundamental problems in SeqCLR: (i) the positive pair, formed by two augmentations at the same location, can suffer from instance misalignment; (ii) negative pairs are selected across different samples, disrupting semantic continuity within an image; and (iii) all contrastive operations are performed on high-level features. Based on these assumptions, they develop a \emph{context perceiver}, which consists of a context aggregator and a quantizer. In the context aggregator, a masking strategy and a windowed MSA are used, which forces to learn the value of each masked patch by its local context \citep{baevski2020wav2vec}. Furthermore, the quantizer operates in a similar way to a Discrete Variational Autoencoder \citep{rolfe2016discrete}, making each patch represented by an embedding from a discrete codebook, thus providing higher abstraction of the features. Finally, contrastive learning is applied within the same image, both reducing the distance between masked patches and their respective quantized representations as well as maximizing the distance to these latter elements of other masked patches. Following a hierarchy in the contrastive learning, this framework is applied to both the output of the encoder $f_{\text{sem}}$ and the representations produced in the middle of the encoder $f_{\text{stroke}}$, as depicted in Fig. \ref{fig:PerSec_diagram}.

Evaluations on CNN-based and ViT-based encoders demonstrate the importance of considering both visual and semantic information, which is provided in this case by hierarchical contrastive learning. Additionally, it is shown that, in general terms, ViT yields better performance, demonstrating the high capacity of this type of architecture for feature learning in SSL methods.

The \emph{Scene Text Recognition with Contrastive Predictive Coding} (\textbf{STR-CPC}) \citep{jiang2022scene} method points out that existing contrastive learning methods focus on instance discrimination but neglect the correlation of the visual sequence within text instances, thus losing such sequential information. To mitigate this issue, they propose the use of CPC \citep{oord2018representation}. However, authors demonstrate that its direct use is not possible, since the use of deeper CNN for TR results in the information overlap problem within adjacent frames in the feature map as each frame incorporates information about adjacent ones, and so the prediction becomes trivial. To solve this problem, they propose a Widthwise Causal Convolution Neural Network. which decreases information overlap, and a progressive training strategy to solve the architectural discrepancy between the CNN used for pre-training and the eventual CNN used for TR. Therefore, they demonstrate the importance of taking into account the inherent sequentiality of TR.

\begin{figure}[!tbh]
  \centering
  \includegraphics[width=0.4\linewidth]{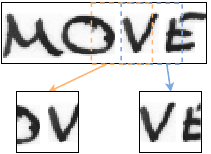}
  \caption{Character Unit Crop module in ChaCo \citep{zhang2022chaco}. From a handwritten text image, two overlapped crops are considered.}
  \label{fig:ChaCo_diagram}
\end{figure}

\cite{zhang2022chaco} argue that the fundamental unit of recognition in TR is the character and, therefore, contrastive learning must be done taking this assumption into account. To do so, they propose \emph{Character Contrastive Learning} (\textbf{ChaCo}), which uses the Character Unit Crop module to generate two random crops of an image ensuring that they contain overlapping parts. By adjusting some hyperparameters, the size of the crops is made similar to the average size of a letter, as shown in Fig. \ref{fig:ChaCo_diagram}. As a contrastive learning framework, they use MoCo v2 to pretrain the visual feature extractor---i.e., a CNN. Their experimentation reveals the importance of information at the character level, especially focusing on the fact that the overlap between positive pairs should not be high so that the encoder learns to distinguish between common and uncommon parts.

More recently, \cite{zhang2022cmt} claim that performing contrastive learning at the frame or subword levels may lead to errors. Given that each frame might contain information from several consecutive characters or an over-segmented subcharacter, the model may be somehow misled by perceiving information from these semantic cues. To address this issue, they introduce \emph{Character Movement Task to Assist Contrastive Learning} (\textbf{CMT-Co}), \review{which enhances word-level contrastive learning (semantic information) by incorporating a character-level pretext task named \emph{Character Movement Task}. The last} task is to move a character by a certain number of pixels, being the model meant to estimate this displacement. The characters to be moved are located using the vertical projection distribution. As contrastive learning, MoCo v2 is employed. The final loss function is a weighted sum of the two tasks with which the visual feature extractor (CNN) is pre-trained. Furthermore, they also propose a data augmentation strategy suitable for handwritten text. This work, therefore, highlights the importance of using data augmentation techniques that best fit the nature of the data and reinforces the importance of focusing on both visual features (at the character level) and semantic features (at the word level).

Finally, \cite{zhang2023relational} point out that textual relations are restricted to the finite size of the dataset, which usually causes the problem of overfitting because of lexical dependencies. To address this problem, they propose \emph{Relational Contrastive Learning for Scene Text Recognition} (\textbf{RCLSTR}), which enriches textual relations through ``rearrangement'', ``hierarchy'', and ``interaction'', resulting in a richer contrastive mechanism. For ``rearrangement'', images are horizontally divided and blended to compute their embeddings with different context relations. For ``hierarchy'', relational contrastive loss is computed in frame, subword, and word levels. For ``interaction'', they employ the symmetric KL divergence to ensure a correlation between the different levels. This divergence is a modification of the standard KL divergence that addresses its asymmetry by computing: 

\begin{equation}
\begin{split}
    & D_{\text{SymKL}}(P||Q) =\\ &\;\;\frac{1}{2} D_{\text{KL}}(P||Q) +  \frac{1}{2} D_{\text{KL}}(Q||P)
\end{split}
\end{equation}

\begin{figure}[!tbh]
  \centering
  \includegraphics[width=0.7\linewidth]{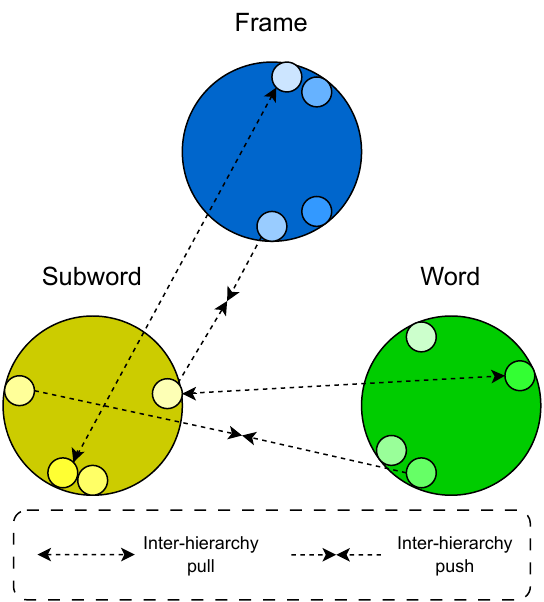}
  \caption{Cross-hierarchy consistency in RCLSTR  \citep{zhang2023relational}. Contrastive learning is performed in each level (frame, subword and word) and between them the correlation is maintained thanks to the Symmetric KL divergence.}
  \label{fig:RCLSTR_diagram}
\end{figure}

The experiments carried out demonstrate that it is important not only to take into account the different levels that encapsulate visual and contextual information, but also to maintain a correlation and coherence between these levels, as shown in Fig. \ref{fig:RCLSTR_diagram}.

Throughout this subsection, it has been evident that contrastive learning for TR significantly differs from its application in image classification. In this latter task, the unit for performing the contrastive process is straightforward, the image itself. However, in TR, identifying the appropriate unit for contrastive learning remains an open question. Various approaches have been explored: some studies use subwords \citep{aberdam2021sequence}, others use characters \citep{zhang2022chaco}, patches \citep{liu2022perceiving}, or a combination of different levels \citep{zhang2022cmt,zhang2023relational}. Consequently, this subcategory within TR faces unique issues, such as misalignment between positive and negative pairs, determining the correct instance level, and integrating different levels of information such as visual, sequential or semantic.

\subsubsection{Geometric transformation}
\begin{figure}[!tbh]
  \centering
  \includegraphics[width=0.8\linewidth]{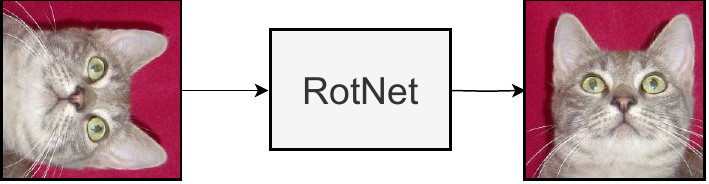}
  \caption{RotNet \citep{gidaris2018unsupervised}. The task consists in predicting the rotation performed to a given image.}
  \label{fig:Rotation}
\end{figure}

Geometric self-supervised learning is a family of approaches that learn useful representations by leveraging the geometric properties and structures inherent in the data \citep{jing2020self}. The pretext task typically consists in predicting a geometric transformation, such as rotations \citep{gidaris2018unsupervised}, applied to the input image (see Fig. \ref{fig:Rotation}). In this way, the encoder learns to understand the underlying spatial and structural relationships of the data. Despite its simplicity, this approach has been demonstrated to be very effective in different scenarios \citep{novotny2018self,zhang2019aet,chen2019self,feng2019self,yamaguchi2021image,liang2021source}.

\cite{penarrubia2024spatial} discuss that HTR is rich in spatial information due to the font as well as the stroke style. Given that text is neither horizontally nor vertically symmetric, they propose \textbf{Flip} as a geometric transformation, which is more meaningful in this context. Then, the pretext task consist in predicting which flip transformation has been performed among the possibilities $\{\text{nothing}, \text{horizontal}, \text{vertical}, \text{both} \}$ applied to a text image. They experimentally demonstrate that this transformation helps the CNN focus on the intrinsic features of characters and words, thereby resulting more effective than the common rotation transformation.

\subsubsection{Puzzle solvers}
\begin{figure}[!tbh]
  \centering
  \includegraphics[width=0.8\linewidth]{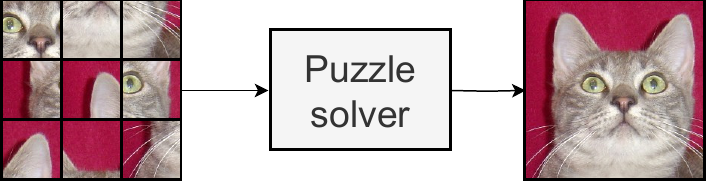}
  \caption{Diagram of Jigsaw puzzle in computer vision. The task consist in predicting the original arrangement among disordered patches.}
  \label{fig:Puzzle}
\end{figure}

In computer vision, the ``puzzle'' pretext task relates to predicting the relative positions of some disordered patches in which a given image has been divided \citep{doersch2015unsupervised,noroozi2016unsupervised} (see Fig. \ref{fig:Puzzle}). This task has shown its effectiveness in the early attempts of SSL, since the relative position among patches of an image encompasses rich spatial and contextual information. Note that, despite being one of the earliest pretext tasks of this type, its use is still common by state-of-the-art approaches \citep{carlucci2019domain,pang2020solving,misra2020self,yang2022fully,baykal2022exploring}.

\begin{figure}[!tbh]
  \centering
  \includegraphics[width=\linewidth]{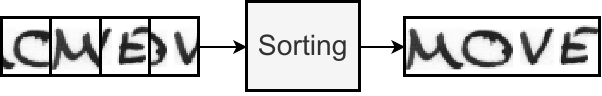}
  \caption{Diagram of Sorting method for HTR \citep{penarrubia2024spatial}. For a given shuffled input image, the networks must predict the correct order of patches.}
  \label{fig:Sorting_diagram}
\end{figure}

Considering the work of \cite{penarrubia2024spatial}, the \textbf{Sorting} method forces a model to infer the correct order of a shuffled set of vertical patches (see Fig. \ref{fig:Sorting_diagram}). This method underscores the sequential nature of text. However, the experiments showed that, among the proposed methods based on spatial context---i.e. geometric and puzzle solvers---, this proposal showed to be the least effective.

\subsubsection{Distillation}
\begin{figure}[!tbh]
  \centering
  \includegraphics[width=\linewidth]{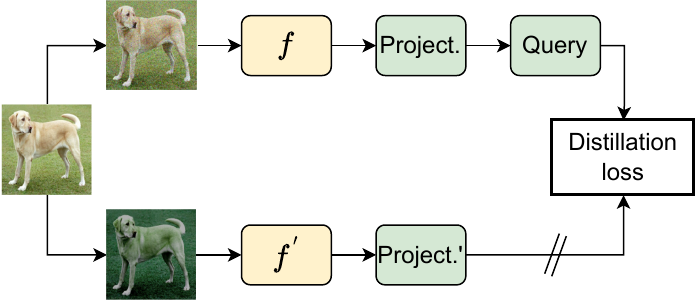}
  \caption{Diagram of the distillation framework for SSL. The framework consists of a teacher-student architecture and a query head that helps preventing neural collapse and leads to better performance.}
  \label{fig:Distillation}
\end{figure}

In dual-branch architectures, neural collapse refers to the problem where an encoder produces constant or dummy (non-informative) representations. While contrastive learning prevents neural collapse by means of negative examples, the works of BYOL \citep{grill2020bootstrap} and SimSiam \citep{chen2021exploring} demonstrated that there are other ways to avoid this issue. These works rely on two main aspects, as shown in Fig. \ref{fig:Distillation}: (i) a teacher-student framework, that employs the stop gradient operation on the teacher branch; and (ii) the use of a prediction head to produce asymmetry between the two branches. Both works proved that the combination of these aspects prevents the network collapse. Moreover, the work of DINO \citep{caron2021emerging} extended these ideas to the application in ViT, also proposing a local-to-global feature alignment and a more complex loss function. In this case, the collapse is avoided using the so-called centering operation, which consists in adding a bias term to the teacher's prediction branch.

The work of \emph{Character-to-character distillation} (\textbf{CCD}) \citep{guan2023self} studied the use of this distillation framework in STR. Authors argue that dividing the images horizontally for instance discrimination limits the aggressiveness of the data augmentation, such as hard geometric augmentations. Furthermore, it does not take into account the structure of the characters since their semantic information is lost when subdividing or mixing characters.  To avoid these problems, \cite{guan2023self} propose using a distillation framework---using the loss function proposed by DINO. Furthermore, to enhance character alignment, they propose the use of a self-supervised character segmentation head to compute the patch heads, thus highlighting the character structure. Note that patch head is a neural scheme that performs nonlinear transformations in the embedded patches. Overall, CCD demonstrates the benefit of employing distillation by itself, the benefit of hard geometric augmentations during training (successful because of the distillation framework), and the enhancement of using their character segmentation head, not only for TR but for other related tasks such as text segmentation.

\subsection{Generative approaches}
With the emergence of ViT architectures, the recent years have seen an increasing interest in generative approaches. Actually, image generation learning has a broader scope \citep{vincent2008extracting} and has shown high capacity for training DNN for a disparate set of tasks such as denoising \citep{zhang2022survey,chen2023auto}, image enhancement \citep{ali2023vision}, or segmentation \citep{minaee2021image}, among many others.

We divide the methods of this paradigm into four subcategories: image colorization, Masked Image Modelling, image inpainting, and Generative Adversarial Networks (GAN). Below we describe each subcategory in which an SSL method for TR has been developed, and in turn, presenting these methods in chronological order.

\subsubsection{Image colorization}
\begin{figure}[!tbh]
  \centering
  \includegraphics[width=\linewidth]{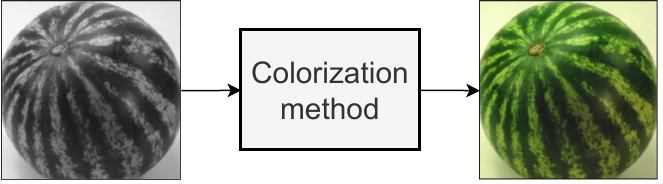}
  \caption{Diagram of the colorization framework. Given a gray scale image, the task is to compute its colored version.}
  \label{fig:Colorization}
\end{figure}

The image colorization is one of the oldest generative pretext task. It consists in predicting the original colored image of a grayscaled input \citep{cheng2015deep,larsson2016learning,larsson2017colorization}, as shown in Fig. \ref{fig:Colorization}. This leads to meaningful representations because, in natural images, objects and entities are usually associated with specific coloring \citep{zhang2016colorful}. Therefore, the model must capture the underlying semantics and structure of the visual content, which helps understanding and distinguishing different parts of the image \citep{caron2020unsupervised,anwar2020image}.

\begin{figure}[!tbh]
  \centering
  \includegraphics[width=0.8\linewidth]{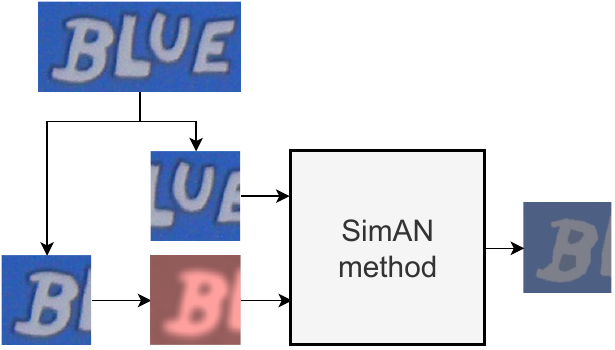}
  \caption{Diagram of SimAN method \citep{luo2022siman}. The original format of an augmented crop is recovered using a non-overlapped and non-augmented one from the same image.}
  \label{fig:SimAN_diagram}
\end{figure}

\cite{luo2022siman} analyze that some constant properties are maintained for the entire image in STR, such as color, texture, stroke width, etc. Based on these assumptions, they propose \emph{Similarity-Aware Normalization} (\textbf{SimAN}), whose pretext task is to recover the original format of an augmented crop of an image by means of another non-overlapped crop from the same image before augmentations (see Fig. \ref{fig:SimAN_diagram}). This makes the task more complex than simple colorization, leading to richer representations in the context of TR. To achieve this, SimAN employs a Similarity-Aware Normalization module in which the local style information of the non-augmented crop is extracted and that of the augmented crop is removed. Then, the original image is reconstructed by using an attention procedure. In addition, this method also takes advantage of the field of GAN \citep{creswell2018generative,aggarwal2021generative,gui2021review} since, by using a discriminator with adversarial loss, SimAN increases the quality of the reconstruction.  This work demonstrates that the Similarity-Aware Normalization module, in addition to extracting adequate features for TR, is capable of handling other tasks such as data synthesis, text image editing, and font interpolation.

\subsubsection{Masked Image Modeling}
\begin{figure}[!tbh]
  \centering
  \includegraphics[width=\linewidth]{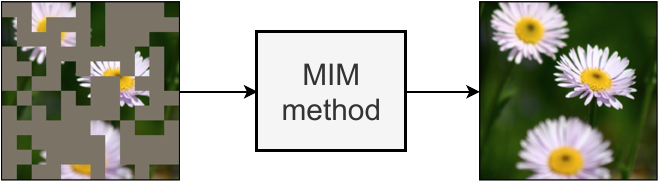}
  \caption{MIM framework that consists in predicting the masked patches of an input image.}
  \label{fig:MIM}
\end{figure}

Within the generative SSL paradigm, Masked Image Modeling (MIM) comes to the fore in the last few years. As shown in Fig. \ref{fig:MIM}, this framework predicts the masked patches of an input image. Since this paradigm was initially proposed for Natural Language Processing (i.e., Masked Language Modeling \citep{qiu2020pre}), it is not surprising that Bidirectional Encoder representation from Image Transformers (BEiT) \citep{bao2021beit}, first SSL task within this category, imitated the well-known Bidirectional Encoder Representations from Transformers (BERT) \citep{devlin2018bert}. BEiT proposes to predict the category of discrete visual tokens that are masked. The discrete visual tokens are generated by a pre-trained discrete variational autoencoder \citep{ramesh2021zero}. Later, Simple Framework for Masked Image Modeling (SimMIM) \citep{xie2022simmim} proposed to predict directly on pixels, showing competitive results without the need for an offline tokenizer. Furthermore, Masked Autoencoder (MAE) \citep{he2022masked}, an asymmetric autoencoder framework, showed that such architectures are efficient feature learners \review{also predicting masked patches at pixel level}. Context Autoencoder (CAE) \citep{chen2024context} showed the enhancement of using a contextual regressor and aligning the representation given by this unit of the masked patches with those produced by the encoder itself.

\begin{figure}[!tbh]
  \centering
  \includegraphics[width=\linewidth]{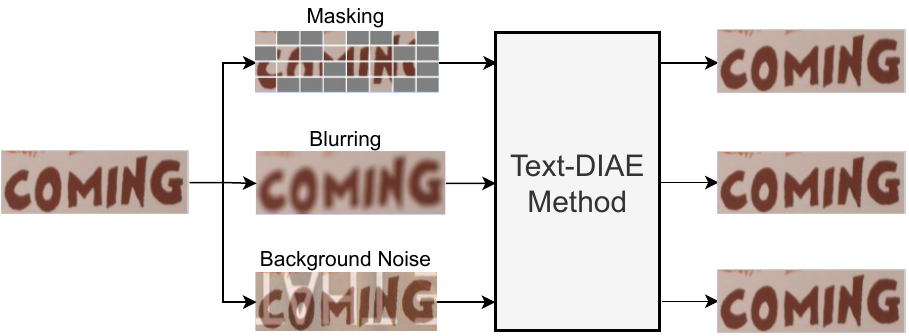}
  \caption{Diagram of the Text-DIAE method \citep{souibgui2023text}, which focuses on reconstructing the degraded input images.}
  \label{fig:textdiae_diagram}
\end{figure}

\emph{Text-Degradation Invariant Autoencoder} (\textbf{Text-DIAE}) \citep{souibgui2023text} is the first method that fully relies on MIM to pre-train a ViT encoder for TR. It leverages a denoising autoencoder for different perturbations: masking, blur, and noise. The method reconstructs at pixel level the degraded input images with an encoder-decoder pipeline (see Fig. \ref{fig:textdiae_diagram}). It should be noted that, after pre-training, the decoder is discarded and only the encoder is kept for the downstream task. In their experiments, authors show that the degradation that provides the greatest benefit for TR is masking; however, they also report that other degradations are useful for other tasks such as image enhancement.

\begin{figure}[!htb]
  \centering
  \begin{subfigure}{\linewidth}
    \centering
    \includegraphics[width=0.4\linewidth]{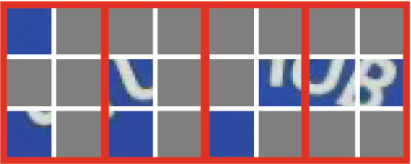}
    \caption{Intra-window masking.}
    \label{fig:DualMAE_intra}
  \end{subfigure}
  \hfill
  \begin{subfigure}{\linewidth}
    \centering
    \includegraphics[width=0.4\linewidth]{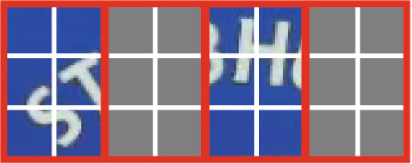}
    \caption{Window masking.}
    \label{fig:DualMAR_window}
  \end{subfigure}
  \caption{Diagram of Dual-MAE masking strategies. Figure taken from \cite{qiao2023decoupling}.}
  \label{fig:DualMAE_intra_DualMAR_window}
\end{figure}

Later, \cite{qiao2023decoupling} argue that MIM is more suitable for TR since the instance discrimination used in contrastive learning is not flexible for this task. Instead, MIM helps the model build the intra-characters and inter-character dependencies, as a character can be recognized by the corresponding visual clues and context information from other characters. Based on the last assumptions, they propose the \emph{Dual Mask Autoencoder} (\textbf{Dual-MAE}) method, which employs two new masking strategies: (i) the intra-window masking (Fig. \ref{fig:DualMAE_intra}), that limits the self-attention to each windows a masks a percentage of patches inside it; and (ii) the window masking (Fig. \ref{fig:DualMAR_window}), that masks entire windows. The former one is intended to obtain the information from the visual content, whereas the latter one takes it from the context. In addition, authors propose two neural architectures based on MAE to leverage these masking strategies: one that uses Siamese networks \review{(Dual-MAE-Siam)} and another that uses distillation \review{(Dual-MAE-Dist)}. Their experiments demonstrate the clear benefit of combining both types of masking strategies. Then again, it is highlighted that it is important to take into account both visual and semantic information for TR.

Up to this point, all methods have focused on just pre-training the encoder \review{or a part of it}. \textbf{MaskOCR} \citep{lyu2023maskocr} represents a framework that addresses this limitation, i.e., it pre-trains both the encoder and the decoder to improve the overall performance of the model on TR tasks. The strategy first pre-trains the encoder using CAE. In a second step, the encoder is frozen and, by means of synthetic generation, words with missing characters are generated, for which the encoder-decoder must fill such gaps. It is worth noting that the proposed method to pre-train the decoder is not completely self-supervised, since it requires synthetically-generated labeled data. However, they showed that incorporating a pre-trained decoder increases the performance of the model, especially for fine-tuning scenarios with scarce labeled data.

\subsection{Hybrid approaches}
In this section we describe those methods that address the self-supervised challenge from several perspectives, instead of relying on a single working principle.

The \emph{Discriminative and Generative} (\textbf{DiG}) method \citep{yang2022reading} combines both generative and contrastive learning into a unified framework. Authors choose SimMIM and MoCo v3 as MIM and contrastive frameworks, respectively, and consider a weighted sum of their loss functions. The benefit of combining both methods is reported experimentally.

\begin{figure}[!tbh]
  \centering
  \includegraphics[width=\linewidth]{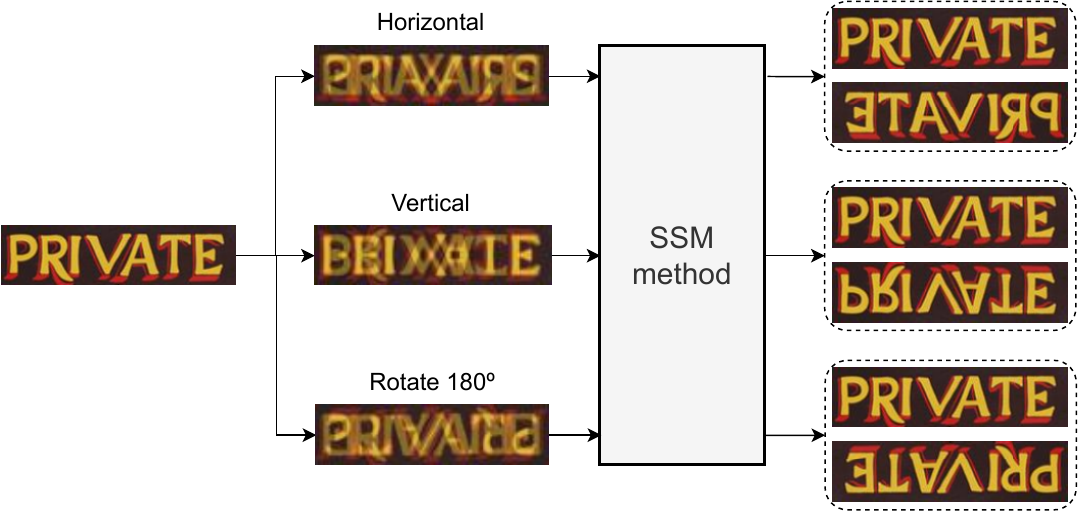}
  \caption{Diagram of SSM method \citep{gao2024self}. From a symmetrically superimposed image, the model must reconstruct both the original image and the imposed one.}
  \label{fig:SSM_diagram}
\end{figure}

\cite{gao2024self} point out that existing methods focus primarily on learning robust visual features of characters, overlooking the linguistic relationship between characters such as spelling. Then, they propose \emph{Symmetric Superimposition Modeling} (\textbf{SSM}), which forces the model to reconstruct two images (original and distorted) from a superimposed input (see Fig. \ref{fig:SSM_diagram}). They consider different ways of generating the distortion, such as horizontal flip, vertical flip, or 180º rotation. Furthermore, inspired by the work of \cite{tao2023siamese}, they perform both contrastive learning and mean squared error loss between the patches generated by the ViT. Therefore, the final loss function is a weighted sum of these three individual components. This way, authors demonstrate the benefit of combining different categories of SSL.

\section{Benchmarking}
\label{sec:benchmarking}
Once the different SSL approaches for TR have been reviewed, this section delves into how their evaluation is addressed. This involves describing the most common datasets for STR and HTR in the literature, as well as the evaluation protocols and metrics considered.

\subsection{Datasets}
The datasets considered, and the way of using them, are quite different depending on the specific task (STR or HTR), and so we elaborate on this issue in different sections. In all cases, it is important to emphasize that we are intentionally not describing all the possible datasets for each task but rather those that have been considered in the SSL-TR literature. We encourage the reader to look for more specific references to learn about more datasets.

\subsubsection{Datasets for STR}
As a field, STR has a particular way of employing existing datasets. The common pipeline of supervised learning for STR consists in training the model on a large synthetic dataset and then evaluating the trained model on several benchmark datasets \citep{baek2019wrong}. Analogously, SSL methods first pre-train the \review{recognition scheme---either in its entirety or part of it---}on one dataset and then fine-tune the whole model in another one. The idea is that the pre-trained model should be able to obtain more discriminative representations. Then, just like in the supervised scenario, the model is evaluated in a set of benchmarks.

\begin{figure}[!t]
  \centering
  \begin{subfigure}{\linewidth}
    \centering
    \includegraphics[width=\linewidth]{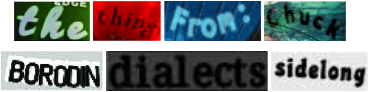}
        \caption{Examples from synthetic datasets.}
    \label{fig:Sinthetic_str}
  \end{subfigure}
  \hfill
  \begin{subfigure}{\linewidth}
    \centering
    \includegraphics[width=\linewidth]{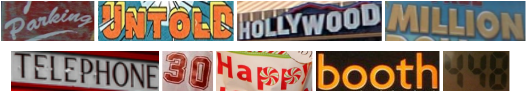}
    \caption{Examples from real datasets.}
    \label{fig:Real_str}
  \end{subfigure}
  \caption{Comparison of synthetic and real STR datasets for STR.}
  \label{fig:Sinthetic_str_Real_str}
\end{figure}

In the first proposed methods, the datasets for pre-training and fine-tuning are the same synthetic dataset \citep{aberdam2021sequence,luo2022siman,liu2022perceiving,souibgui2023text}. Later, the same procedure is extended to the use of real unlabeled datasets for pre-training and real labeled datasets for fine-tuning \citep{yang2022reading,qiao2023decoupling,lyu2023maskocr,gao2024self}. Figure \ref{fig:Sinthetic_str_Real_str} illustrates both synthetic and real datasets for STR, while Table \ref{tab:training_str} provides a summary. We describe below their most important characteristics:

\begin{itemize}
    \item \textbf{SynthText (ST)} \citep{gupta2016synthetic} is a synthetically generated dataset, originally designed for scene text detection. It comprises over 8M text boxes, of which 5.5M do not contain non-alphanumeric characters.
    \item \textbf{MJSynth (MJ)} \citep{jaderberg2014synthetic} is a synthetic dataset containing 8.9M images each containing a single word rendered with various fonts, sizes, colors, orientations, backgrounds and distortions to simulate real-world conditions.
    \item \textbf{Synthetic Text Data (STD)} is an established combination of ST and MJ, resulting in a dataset of approximately 17M of samples.
    \item \textbf{Modified-SynthText (Mod-ST)} is a modified version of ST employed by \cite{gomez2018single}. It contains 4M cropped scene text images generated synthetically. The main difference with the previous ones is that Mod-ST avoids random flipping and rotation of text images, which can disrupt the sequential order of characters.
    
    \item \textbf{UTI-100M} is a real unlabeled dataset proposed by \cite{liu2022perceiving}. It contains 100M images captured by mobile camera devices, covering 5 scenes: street view, receipt, product package, book, and poster, which are mainly written in English, Chinese, Japanese, and Korean.

    \item \textbf{Unlabeled Real Data (URD)} \citep{yang2022reading} consists of 15.77M unlabeled text images (termed CC-OCR) from Conceptual Captions Dataset (CC) \citep{yang2021tap}, which are exclusively used for self-supervised learning in the STR domain.

    \item \textbf{Annotated Real Data (ARD)} consists of 2.78M annotated text images from TextOCR \citep{singh2021textocr} (0.71M) and Open Images Dataset v5 (2.07M).

    \item \textbf{Real-U} \citep{baek2021if} is real unlabeled dataset containing 4.2M of text images collected from Book32 \citep{iwana2016judging}, TextVQA \citep{singh2019towards}, and ST-VQA \citep{biten2019scene}.

    \item \textbf{Real-L} \citep{baek2021if} includes 276K training images and 63K validation images gathered from several datasets: SVT \citep{wang2011end}, IIIT \citep{mishra2012scene}, IC13 \citep{karatzas2013icdar}, IC15 \citep{karatzas2015icdar}, COCO-Text \citep{veit2016coco}, RCTW \citep{shi2017icdar2017}, Uber \citep{zhang2017uber}, ArT \citep{chng2019icdar2019}, LSVT \citep{sun2019icdar}, MLT19 \citep{nayef2019icdar2019}, ReCTS \citep{zhang2019icdar}.

\end{itemize}

\begin{table}[!htb]
\centering
\caption{Summary of the STR datasets for training in the related SSL literature.}
\label{tab:training_str}
\begin{tabular}{llcccl}
\toprule[1pt]
 & \textbf{Name} & \textbf{Type} & \textbf{Labeled} & \textbf{Images} &  \\ \cmidrule{2-5}
 & ST & synthetic & \cmark & 8M &  \\
 & MJ & synthetic & \cmark & 8.9M &  \\
 & STD & synthetic & \cmark & 17M &  \\
 & Mod-ST & synthetic & \cmark & 4M &  \\
 & UTI-100M & real & \xmark & 100M &  \\
 & URD & real & \xmark & 17.77M &  \\
 & ARD & real & \cmark & 2.78M &  \\
 & Real-U & real & \xmark & 4.2M &  \\
 & Real-L & real & \cmark & 276K &  \\
\bottomrule[1pt]
\end{tabular}
\end{table}

In terms of evaluation, the STR literature generally uses 6 reference datasets to measure and compare the effectiveness of the trained models. Note that in this paper we focus on those benchmark datasets that have been established as the standard for evaluation in SSL-TR literature, and we have left out those more specific to certain works or areas---i.e. Chinese datasets, Arabic datasets, occluded scene text etc. Depending on the difficulty and geometric layout of the texts, the SSL-STR benchmark datasets are classified as \emph{regular} or \emph{irregular}.

\begin{figure}[!htb]
  \centering
  \begin{subfigure}{\linewidth}
    \centering
    \includegraphics[width=\linewidth]{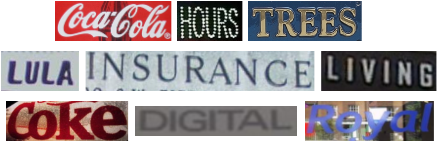}
        \caption{Examples from \emph{regular} STR benchmark datasets.}
    \label{fig:regular_str_bench}
  \end{subfigure}
  \hfill
  \begin{subfigure}{\linewidth}
    \centering
    \includegraphics[width=\linewidth]{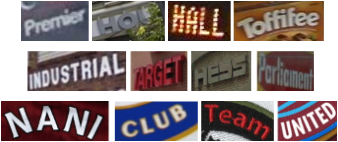}
    \caption{Examples from \emph{irregular} STR benchmark datasets.}
    \label{fig:irregular_str_bench}
  \end{subfigure}
  \caption{Examples of both \emph{regular} and \emph{irregular} STR benchmark datasets.}
  \label{fig:regular_str_bench_irregular_str_bench}
\end{figure}

The \emph{regular} datasets are those whose images contain the letters arranged horizontally with regular spaces between them, as depicted in Fig. \ref{fig:regular_str_bench}. Due to these regularities, it is usually easier to attain high recognition rates. The regular datasets considered in the literature are the following:

\begin{itemize}
    \item \textbf{IIIT5K-Words (IIIT)} \citep{mishra2012scene} contains 5K word images collected from the web, with a split of 2K for training and 3K for evaluation. The dataset includes a variety of fonts, colors, and backgrounds.

    \item \textbf{Street View Text (SVT)} \citep{wang2011end} includes 257 images for training and 647 images for evaluation, extracted from street view photos, focusing on challenging scenarios like low resolution, blurriness, and varying lighting conditions.

    \item \textbf{ICDAR 2013 (IC13)} \citep{karatzas2013icdar} is an updated version of the ICDAR 2003 dataset \cite{lucas2005icdar} with more challenging text samples and annotations. It contains 848 images for training and 1095 images for evaluation, out of them 1015 and 857 respectively stand for words with non-alphanumeric characters and with less than 3 characters.
    
\end{itemize}

The \emph{irregular} datasets are those that contain letters arranged in different geometric shapes, such as curved and arbitrarily rotated or distorted texts, as illustrated in Fig. \ref{fig:irregular_str_bench}. As a result, these datasets tend to be more challenging for STR. The common irregular datasets are the following:

\begin{itemize}
    \item \textbf{ICDAR 2015 (IC15)} \citep{karatzas2015icdar}, which is known for its complex images. This dataset includes incidental scene text images taken by Google Glass. It has 4468 images for training and 2077 images for evaluation, of which 1811 do not contain non-alphanumeric characters and are not extremely rotated, perspective-shifted, or curved images.

    \item \textbf{SVT Perspective (SP)} \citep{phan2013recognizing} is created from Google Street View and contains 645 images for evaluation.

    \item \textbf{CUTE80 (CT)} \citep{risnumawan2014robust} is collected from natural scenes and contains 288 cropped text images for evaluation. 

\end{itemize}

\begin{table}[!htb]
\centering
\caption{Summary of the STR datasets used for evaluation in the related SSL literature.}
\label{tab:bechmark_str}
\begin{tabular}{llccl}
\toprule[1pt]
 & \textbf{Benchmark} & \textbf{Test images} & \textbf{Description} &  \\ \cmidrule{2-4}
 & IIIT & 3000 & Regular &  \\
 & SVT & 647 & Regular &  \\
 & IC13 & 1095/1015/857 & Regular &  \\
 & SP & 645 & Irregular &  \\
 & IC15 & 2077/1811 & Irregular &  \\
 & CT & 288 & Irregular & \\
\bottomrule[1pt]
\end{tabular}
\end{table}

A summary of the datasets considered for STR evaluation is shown in Table \ref{tab:bechmark_str}. Given that these datasets are considered for evaluating the SSL mechanisms, the training partitions are either used for validation or simply discarded, unless mentioned otherwise.

\subsubsection{Datasets for HTR}
The way of using datasets for HTR is usually simpler. Typically, in the classic supervised learning scenario, there is a labeled dataset with predefined partitions, which makes comparisons easier. Therefore, the model is just trained with the training set and evaluated with the test set. Figure \ref{fig:iam_images_cvl_images_rimes_images} provides examples from HTR datasets.

In the case of employing SSL, the model is first pre-trained and then fine-tuned with the training set. Quite often, fine-tuning is considered only with a reduced percentage of the training set, in order to see how the pre-trained model is able to take better advantage of limited data. In this sense, there is currently no standardized protocol, and this issue will be further discussed in Section \ref{What_is_wrong_with_HTR}.

Below we describe the HTR datasets that are most frequently used in the context of SSL:

\begin{figure}[!htb]
  \centering
  \begin{subfigure}{\linewidth}
    \centering
    \includegraphics[width=\linewidth]{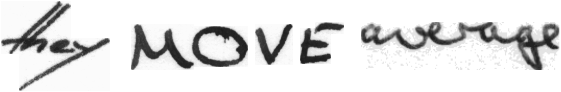}
    \caption{Examples from IAM dataset.}
    \label{fig:iam_images}
  \end{subfigure}
  \hfill
  \begin{subfigure}{\linewidth}
    \centering
    \includegraphics[width=\linewidth]{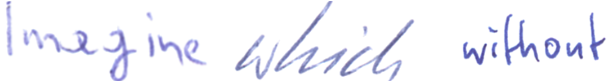}
    \caption{Examples from CVL dataset.}
    \label{fig:cvl_images}
  \end{subfigure}
  \hfill
  \begin{subfigure}{\linewidth}
    \centering
    \includegraphics[width=\linewidth]{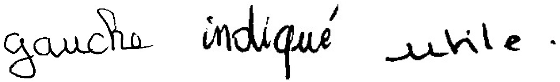}
    \caption{Examples from RIMES dataset.}
    \label{fig:rimes_images}
  \end{subfigure}
  \caption{Examples of datasets typically considered in SSL.}
  \label{fig:iam_images_cvl_images_rimes_images}
\end{figure}

\begin{itemize}
    \item \textbf{IAM} dataset \citep{marti2002iam} is one of the most widely used benchmarks for HTR. This dataset contains 115,320 isolated and labeled words from 1,539 scanned handwritten English pages from 657 different writers.

    \item \textbf{CVL} dataset \citep{kleber2013cvl} is a comprehensive collection of modern handwritten text. The dataset consists of 1 German and 6 English texts. In total, 310 writers appear in the dataset, 27 of which wrote 7 texts and 283 writers who write 5 texts.

    \item \textbf{RIMES} dataset \citep{grosicki2009icdar} is primarily used for the recognition of French handwritten text. It contains 12,723 annotated pages from a total of 1,300 writers. 
    
\end{itemize}

It is important to remark that these datasets are used for SSL at the word level, unlike the conventional case of HTR where it may be more common at the line level. A quantitative summary of the datasets is shown in Table \ref{tab:htr_datasets}.

\begin{table}[!htb]
\centering
\caption{Summary of the HTR datasets in the related SSL literature.}
\label{tab:htr_datasets}
\begin{tabular}{llcccl}
\toprule[1pt]
 & \textbf{Name} & \textbf{Images} & \textbf{Writers} & \textbf{Language} &  \\ \cline{2-5}
 & IAM & 79275 & 657 & Eng &  \\
 & CVL & 99904 & 310 & Eng, Ger &  \\
 & RIMES & 66480 & 1300 & Fr &  \\
\bottomrule[1pt]
\end{tabular}
\end{table}

\subsection{SSL evaluation protocols}
SSL is originally devised to pre-train the encoder with unsupervised data, and then fine-tune it together with a prediction head (in our case a decoder) for any target downstream task with labeled data \citep{balestriero2023cookbook}. Following this original approach, two main protocols are established in the SSL-TR literature, graphically illustrated in Fig. \ref{fig:SSL_protocols_1_SSL_protocols_2}:

\begin{figure*}[!t]
  \centering
  \begin{subfigure}{0.49\linewidth}
    \centering
    \includegraphics[width=0.55\linewidth]{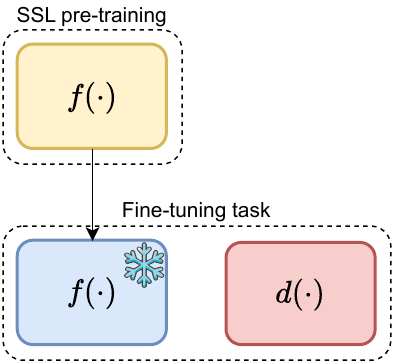}
    \caption{Quality evaluation protocol.}
    \label{fig:SSL_protocols_1}
  \end{subfigure}
  \hfill
  \begin{subfigure}{0.49\linewidth}
    \centering
    \includegraphics[width=0.55\linewidth]{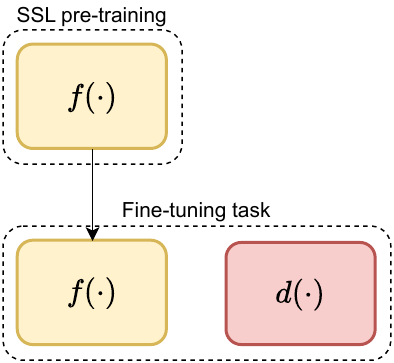}
    \caption{Fine-tuning evaluation protocol.}
    \label{fig:SSL_protocols_2}
  \end{subfigure}
  \caption{SSL evaluation protocols. The quality evaluation freezes the pre-trained part while the Fine-tuning evaluation fine-tuned the whole DNN starting from the pre-trained parts. Note that in this case the pre-trained part is the full encoder, but it does not have to. Perhaps it can be only the CNN in a CRNN architecture or both the encoder and decoder.}
  \label{fig:SSL_protocols_1_SSL_protocols_2}
\end{figure*}

\begin{itemize}
    \item \textbf{Quality evaluation} in SSL involves a key protocol where pre-trained layers are frozen and only the remaining parts of the model are fine-tuned, as shown (see Fig. \ref{fig:SSL_protocols_1}). Widely acknowledged in the literature, this protocol directly assesses the quality of the learned representations from unlabeled data. Effective SSL methods are expected to produce representations that accurately capture significant and discriminative features of the data. Under this protocol, SSL techniques showcase their capability to generalize across diverse datasets and tasks, reducing reliance on labeled data. Moreover, quality evaluation offers insights into the underlying mechanisms of the SSL models, promoting a deeper comprehension of their behavior and guiding the development towards more robust techniques. Note that here we refer to ``pre-trained layers'' and not to the encoder. This is because, as explained above, there are different architectures in TR, and these in turn are made up of different parts. Therefore, not all SSL methods pre-train and freeze the same layers. For example, there are methods that only pre-train the CNN for a CRNN encoder \citep{zhang2022chaco,zhang2023relational}, whereas others pre-train both the CNN and the RNN \citep{aberdam2021sequence}. Similarly, instead of pre-train only the ViT encoder in a fully transformer architecture \citep{yang2022reading,qiao2023decoupling}, some authors choose to pre-train both the encoder and the decoder \citep{lyu2023maskocr}. So far there is no clear and specific guide in this protocol due to the great heterogeneity of the architectures and approaches. Unfortunately, this makes a generic benchmarking for quality evaluation difficult because it is unfair to compare models with different frozen layers.

    \item \textbf{Fine-tuning evaluation} consists in fine-tuning the whole DNN using the pre-training as initial weights, as illustrated in Fig. \ref{fig:SSL_protocols_2}. This protocol leverages the transfer learning capacity of the pre-trained layers. In addition, it is also intended to demonstrate how the rest of the untrained layers, or directly the decoder, benefit from starting with significant and meaningful representations, greatly reducing the need for large volumes of data and showcasing the ability to leverage both unlabeled and labeled data. Precisely for this reason, this protocol is especially useful to demonstrate the ability of SSL methods to address tasks in which there is a scarcity of data. Furthermore, due to the heterogeneity in architectures and pre-training strategies, this protocol is the one that provides the fairest benchmarking. This is because each method decides which parts to pre-train and which parts to initialize randomly for the downstream task. Ultimately, this allows showing which method is able to take the most from the available data.
\end{itemize}

\subsection{Evaluation Metrics}
Once the TR models have been trained, it is necessary to evaluate them in the downstream task (HTR or STR). It is important to note that these evaluation metrics are primarily derived from the HTR and STR fields. The implementation of SSL does not alter how the eventual TR models are evaluated but rather affects the training protocols used to develop these models (previous section). In the literature, three main metrics are considered for this purpose:

\begin{itemize}
    \item \textbf{Character Error Rate (CER):} This metric is defined as the average number of editing operations (insertions, deletions, or substitutions)---i.e., Levenshtein distance---needed to align the model's predicted string with the ground truth, normalized by the length of the latter. Therefore, this metric does not treat the prediction as a binary outcome (correct or incorrect), but rather considers the specific errors made within that sequence. Note that the lower the value, the better. Formally, it is defined as:
    \begin{equation}
        \text{CER (\%)} = \frac{\sum_{i=1}^{\left | S \right |} \text{ED}\left ( \mathbf{\hat{s}}_i , \mathbf{s}_i \right )}{\sum_{i=1}^{\left | S \right |} \left | \mathbf{s}_i \right | }
    \end{equation}
    \noindent where $\text{ED} \left ( \cdot, \cdot \right )$ is the Levenshtein distance, $S$ is the test set of data, and $\mathbf{s}_i$ and $\mathbf{\hat{s}}_i$ are the ground-truth and predicted sequences, respectively.

    \item \textbf{Word Accuracy (WAcc):} This metric evaluates the recognition accuracy at the word level. It measures the proportion of correctly recognized words out of the total number of words in the ground truth. Formally, it is defined as:
    \begin{equation}
        \text{WAcc (\%)} = \frac{\sum_{i=1}^{\left | S \right |} \llbracket\text{ED}\left ( \mathbf{\hat{s}}_i , \mathbf{s}_i \right ) = 0 \rrbracket}{\left | S \right |}
    \end{equation}
    \noindent where $\llbracket\cdot\rrbracket\rightarrow\left\{0,1\right\}$ is the Iverson bracket, which returns $1$ when the condition is met and $0$ otherwise.

    \item \textbf{Single Edit Distance (ED1):} This metric is somewhat placed between CER and WAcc. It measures the accuracy of word recognition by allowing for a single edit operation, making it more lenient than WAcc but stricter than CER. This metric treats the output as a binary outcome, distinguishing between words that are nearly correct from those that clearly incorrect, providing a nuanced evaluation of recognition performance. Formally, it is defined as:
    \begin{equation}
        \text{ED1 (\%)} = \frac{\sum_{i=1}^{\left | S \right |} \llbracket\text{ED}\left ( \mathbf{\hat{s}}_i , \mathbf{s}_i \right ) \leq 1 \rrbracket}{\left | S \right |}
    \end{equation}
    
\end{itemize}

\section{Comparative analysis of performance}
\label{sec:Results_comparison}
In this section, we describe, compare, and analyze the results obtained by the different SSL methods reviewed. In order to provide a global vision, Table \ref{tab:architecture_details} provides a summary of the architectural details and pre-trained parts of the methods to be compared. Moreover, we point out the inconsistencies found with the aim of providing clearer conclusions and encouraging fairer comparisons. Despite the great similarities between HTR and STR, such as architectures or problem formulation, the datasets and the evolution of each field are different. Therefore, we will separately elaborate on the comparative issue. 

\begin{table*}[t]
\centering
\caption{Implementation details of the methods compared, along with their pre-trained parts and whether they present STR or HTR results. \review{For IC13 and IC15 datasets, the test size is included as super index whenever specified in the original papers}.}
\label{tab:architecture_details}
\begin{tabular}{llcccccc}
\toprule[1pt]
 & \multirow{2}{*}{\textbf{Method}} & \multirow{2}{*}{\textbf{Encoder}} & \multirow{2}{*}{\textbf{Decoder}} & \multirow{2}{*}{\textbf{Pre-trained}} & \multicolumn{2}{c}{\textbf{Datasets}} & \multicolumn{1}{l}{} \\ \cline{6-7}
 &  &  &  &  & \textbf{STR} & \textbf{HTR} &  \\ \cline{2-7}
 & SeqCLR & CRNN & CTC, Att & CRNN & IIIT IC03 $\text{IC13}^{1015}$ & IAM CVL RIMES &  \\ \cline{2-7}
 & SimAN & CRNN & CTC, Att & CNN & \begin{tabular}[c]{@{}c@{}}IIIT $\text{IC13}^{1015}$ SVT\\ IC15 CT SP\end{tabular} & - &  \\ \cline{2-7}
 & PerSec & CNN, ViT & CTC, Att & CNN, ViT & \begin{tabular}[c]{@{}c@{}}IIIT IC13 SVT\\ IC15 CT SP\end{tabular} & IAM CVL &  \\ \cline{2-7}
 & DiG & ViT & TD & ViT & \begin{tabular}[c]{@{}c@{}}IIIT $\text{IC13}^{1015}$ SVT\\ $\text{IC15}^{1811}$ CT SP\end{tabular} & IAM CVL &  \\ \cline{2-7}
 & STR-CPC & CRNN & CTC, Att & CNN & \begin{tabular}[c]{@{}c@{}}IIIT IC13 SVT\\ IC15 CT SP\end{tabular} & - &  \\ \cline{2-7}
 & ChaCo & CRNN & CTC, Att & CNN & - & IAM CVL RIMES &  \\ \cline{2-7}
 & CMT-Co & CRNN & CTC, Att & CNN & - & IAM CVL RIMES &  \\ \cline{2-7}
 & Text-DIAE & ViT & TD & ViT & IIIT $\text{IC13}^{1015}$ & IAM CVL &  \\ \cline{2-7}
 & CCD & ViT & TD & ViT & \begin{tabular}[c]{@{}c@{}}IIIT $\text{IC13}^{1015}$ SVT\\ $\text{IC15}^{1811}$ CT SP\end{tabular} & - &  \\ \cline{2-7}
 & Dual-MAE & ViT & TD & ViT & \begin{tabular}[c]{@{}c@{}}IIIT $\text{IC13}^{857,1015}$ SVT\\ $\text{IC15}^{1811,2077}$ CT SP\end{tabular} & - &  \\ \cline{2-7}
 & MaskOCR & ViT & TD & ViT+TD & \begin{tabular}[c]{@{}c@{}}IIIT $\text{IC13}^{857}$ SVT\\ $\text{IC15}^{1811}$ CT SP\end{tabular} & - &  \\ \cline{2-7}
 & RCLSTR & CRNN & CTC, Att & CNN & \begin{tabular}[c]{@{}c@{}}IIIT $\text{IC13}^{1015}$ SVT\\ $\text{IC15}^{2077}$ CT SP\end{tabular} & - &  \\ \cline{2-7}
 & Flip & CRNN & CTC, TD & CNN & - & IAM CVL RIMES &  \\ \cline{2-7}
 & Sorting & CRNN & CTC, TD & CRNN & - & IAM CVL RIMES &  \\ \cline{2-7}
 & SSM & ViT & TD & ViT & \begin{tabular}[c]{@{}c@{}}IIIT $\text{IC13}^{1015}$ SVT\\ $\text{IC15}^{2077}$ CT SP\end{tabular} & - & \\
\bottomrule[1pt]
\end{tabular}
\end{table*}

\subsection{Analysis of SSL performance for STR}
Although the use of SSL for TR is relatively recent, the emergence of new approaches has been increasing especially for the task of STR. Unfortunately, many of the existing methods are not compared or referenced among themselves. In this section we fill that gap with a quantitative comparison of the reported performances. Note that the results and number of parameters are taken from the original papers. In the cases where the number of parameters is not reported, we carefully followed the \review{architectural} description from the paper \review{to calculate the number of parameters when enough hyperparameters were reported}.

In order to enable a comprehensive comparison, \review{Table} \ref{tab:str_synth_fintuning} \review{shows the performance of all pre-trained method's models for the six common STR benchmarks when they are fine-tuned with synthetic data}. It can be observed that for \review{the regular datasets} the modern techniques often achieve near-perfect accuracy. Furthermore, substantial improvements over the years are observed for \review{the irregular datasets}, which are known for its challenging nature due to their irregular image complexities. Note that the training data for each method's model is shown in Table \ref{tab:datasets_for_methods}.


\begin{table*}[t]
\centering
\caption{\review{Comparison in terms of WAcc of the methods presenting results under the fine-tuning evaluation protocol for the six common STR benchmarks. After pre-training, all method's models are fine-tuned with synthetic data. More specific details about data configurations can be found in Table \ref{tab:datasets_for_methods}. Inside each architectural type the methods are arranged chronologically. For each model it is assumed TD unless specified in parentheses. Symbol $\dagger$ indicates that test size is not specified in the reference work.}}
\label{tab:str_synth_fintuning}
\begin{tabular}{lllccccccccccc}
\toprule[1pt]
\textbf{} & \multirow{3}{*}{\textbf{Model}} &  & \multicolumn{4}{c}{\textbf{Regular}} &  & \multicolumn{4}{c}{\textbf{Irregular}} &  &  \\ \cline{4-7} \cline{9-12}
\multicolumn{2}{l}{} &  & \multirow{2}{*}{\textbf{IIIT}} & \multirow{2}{*}{\textbf{SVT}} & \multicolumn{2}{c}{\textbf{IC13}} &  & \multirow{2}{*}{\textbf{SP}} & \multicolumn{2}{c}{\textbf{IC15}} & \multirow{2}{*}{\textbf{CT}} &  &  \\ \cline{6-7} \cline{10-11}
\multicolumn{2}{l}{} &  &  &  & 857 & 1015 &  &  & 1811 & 2077 &  & \textbf{Avg} & \textbf{Par.} \\ \cline{1-2} \cline{4-14} 
\multicolumn{2}{l}{CNN-based} &  &  &  &  &  &  &  &  &  &  &  &  \\
 & SeqCLR (ctc) &  & 80.9 & - & - & 86.3 &  & - & - & - & - & - & 48M \\
 & SeqCLR (att) &  & 82.9 & - & - & 87.9 &  & - & - & - & - & - & 49M \\
 & SimAN (att) &  & 87.5 & 80.6 & - & 89.9 &  & 68.3 & \multicolumn{2}{c}{71.4} & 66.2 & 77.3 & 49M \\
 & PerSec-CNN (ctc) &  & 82.2 & 83.1 & \multicolumn{2}{c}{87.9} &  & 70.4 & \multicolumn{2}{c}{62.3} & 63.5 & 74.9 & - \\
 & PerSec-CNN (att) &  & 84.2 & 82.4 & \multicolumn{2}{c}{88.9} &  & 73.6 & \multicolumn{2}{c}{68.2} & 68.4 & 77.6 & - \\
 & RCLSTR (att) &  & 86.0 & 83.2 & - & 91.1 &  & 74.9 & - & 69.2 & 67.9 & 78.7 & 49M \\
 \hdashline
\multicolumn{2}{l}{ViT-based} &  & \multicolumn{1}{l}{} & \multicolumn{1}{l}{} & \multicolumn{1}{l}{} & \multicolumn{1}{l}{} & \multicolumn{1}{l}{} & \multicolumn{1}{l}{} & \multicolumn{1}{l}{} & \multicolumn{1}{l}{} & \multicolumn{1}{l}{} & \multicolumn{1}{l}{} & \multicolumn{1}{l}{} \\ 
 & PerSec-ViT (ctc) &  & 83.7 & 83.0 & \multicolumn{2}{c}{$89.7^\dagger$} &  & 71.4 & \multicolumn{2}{c}{$64.6^\dagger$} & 65.2 & 76.3 & - \\
 & PerSec-ViT (att) &  & 85.2 & 84.9 & \multicolumn{2}{c}{$89.2^\dagger$} &  & 75.9 & \multicolumn{2}{c}{$70.9^\dagger$} & 69.1 & 79.2 & - \\
 & $\text{DiG}_{\text{Tiny}}$ &  & 95.8 & 92.9 & - & 96.4 &  & 87.4 & 84.8 & - & 86.1 & 90.6 & 20M \\
 & $\text{DiG}_{\text{Small}}$ &  & 96.7 & 93.4 & - & 97.1 &  & 90.1 & 87.1 & - & 88.5 & 92.2 & 36M \\
 & $\text{DiG}_{\text{Base}}$ &  & 96.7 & 94.6 & - & 96.9 &  & 91.0 & 87.1 & - & 91.3 & 92.9 & 52M \\
 & Text-DIAE &  & 86.1 & - & - & 92.0 &  & - & - & - & - & - & 68M \\
 & $\text{CCD}_{\text{Tiny}}$ &  & 96.5 & 93.4 & - & 96.3 &  & 89.8 & 85.2 & - & 89.2 & 92.6 & 20M \\
 & $\text{CCD}_{\text{Small}}$ &  & 96.8 & 94.4 & - & 96.6 &  & 91.3 & 87.3 & - & 92.4 & 93.6 & 36M \\
 & $\text{CCD}_{\text{Base}}$ &  & 97.2 & 94.4 & - & 97.0 &  & 91.8 & 87.6 & - & 93.3 & 94.0 & 52M \\
 & Dual-MAE-Siam &  & 96.7 & 94.6 & 97.2 & 95.8 &  & 89.6 & 87.1 & 83.2 & 91.0 & 91.5 & 36M \\
 & Dual-MAE-Dist &  & 97.0 & 93.5 & 97.7 & 96.4 &  & 90.9 & 86.9 & 83.3 & 91.6 & 91.5 & 36M \\
 & $\text{MaskOCR}_{\text{Small}}$ &  & 95.8 & 94.0 & 97.7 & - &  & 90.2 & 87.5 & - & 89.2 & 93.0 & 31M \\
 & $\text{MaskOCR}_{\text{Base}}$ &  & 95.8 & 94.9 & 98.1 & - &  & 89.8 & 87.5 & - & 90.3 & 93.1 & 97M \\
 & $\text{SSM}_{\text{Tiny}}$ &  & 96.5 & 94.4 & - & 96.3 &  & 89.3 & - & 85.6 & 89.9 & 92.8 & 20M \\
 & $\text{SSM}_{\text{Small}}$ &  & 97.4 & 94.6 & - & 96.7 &  & 91.3 & - & 86.8 & 94.8 & 93.8 & 36M \\
\bottomrule[1pt]
\end{tabular}
\end{table*}

The results report a clear dominance of methods that are developed for ViT architectures over those that use CNN. This dominance is heavily attributed to the low inductive bias of ViT architectures, especially useful in SSL, that allow them to capture dependencies and contextual information more effectively than CNN. Although ViTs are known to be more data-demanding, this is in turn palliated by the use of SSL. Moreover, ViT's self-attention mechanisms allows processing and understanding global patterns within the data, which is particularly beneficial for tasks requiring high-level semantic understanding, such as TR. 

Concerning methodologies, from the first method (SeqCLR) to the latest included in this review (SSM), WAcc has been increasing in a very pronounced way in a short period of time. For instance, there has been an improvement of around 15 units of WAcc in the IIIT dataset (Table  \ref{tab:str_synth_fintuning}). This rapid evolution might be explained by the fact that the research line of SSL for STR greatly benefits from all the previous and current developments that appear for image classification (already reflected in Section \ref{sec:SSL_taxonomy_for_TR}).

\begin{table*}[!t]
\centering
\caption{Datasets used for pre-training and fine-tuning by the different SSL-STR methods. This data configurations are the considered for the results illustrated in Table \ref{tab:str_synth_fintuning} and Table \ref{tab:str_real_fintuning}, and Fig. \ref{fig:real_vs_synth}. The differences between data configurations give rise to unfair comparisons and these are addressed in Section \ref{sec:What_is_wrong_with_STR}.}
\label{tab:datasets_for_methods}
\begin{tabular}{lllcccl}
\toprule[1pt]
 & \multirow{2}{*}{\textbf{Method}} &  & \multirow{2}{*}{\textbf{Pre-training (SSL)}} & \multicolumn{2}{c}{\textbf{Fine-tuning}} &  \\ \cline{5-6}
 &  &  &  & \textbf{Synthetic} & \textbf{Real} &  \\ \cline{2-2} \cline{4-6}
 & SeqCLR &  & ST (8M) & ST (8M) & - &  \\
 & SimAN &  & ST (8M) & ST (8M) & - &  \\
 & PerSec &  & STD (17M) & STD (17M) & - &  \\
 & DiG &  & STD+URD (32.77M) & STD (17M) & ARD (2.78M) &  \\
 & Text-DIAE &  & Mod-ST (4M) & Mod-ST (4M) & - &  \\
 & RCLSTR &  & ST (5.5) & ST (5.5) & - &  \\
 & STR-CPC &  & Real-U (4.2M) & - & Real-L (276K) &  \\
 & MaskOCR &  & STD+URD (32.77M) & STD (17M) & ARD (2.78M) &  \\
 & CCD &  & STD+URD (32.77M) & STD (17M) & ARD (2.78M) &  \\
 & Dual-MAE &  & STD+ARD (19.78M) & STD (17M) & ARD (2.78M) &  \\
 & SSM &  & URD (15.77) & STD (17M) & ARD (2.78M) & \\
\bottomrule[1pt]
\end{tabular}
\end{table*}

\subsubsection{What is wrong when comparing SSL methods for STR?}
\label{sec:What_is_wrong_with_STR}

So far, the comparison and analysis of the methods has focused specifically on the results and architectural details provided by the different method's models. Therefore, certain differences with respect to the data used during training have been omitted. In order to give a much more complete and realistic picture, in this section we highlight the inconsistencies in comparing different SSL methodologies for STR. It is worth noting that this section is not intended to dismiss the conclusions drawn above, but rather to nuance and provide much more complete conclusions that allow us to reach a greater degree of understanding and also promote more accurate benchmarking.

It is well known that errors can arise when comparing models due to data issues in STR \citep{baek2019wrong}, and the case of SSL is no exception. From the earliest methods to the most recent ones, the use of training data has evolved and expanded, leading to distorted comparisons between different models. For instance, Table \ref{tab:datasets_for_methods} presents the data used in the state of the art for pre-training and fine-tuning. Early works, such as SeqCLR and SimAN, use 8M of synthetic data for both pre-training and fine-tuning, while later methods like DiG, CCD, Dual-MAE, MaskOCR, and SSM employ larger datasets, utilizing unlabeled real data for pre-training and larger amounts of synthetic data for \review{the synthetic} fine-tuning. This variation in data usage complicates direct comparisons and highlights the need for standardized evaluation protocols.

It is not surprising that larger amounts of training data favor higher performance. A clear example of this is Text-DIAE. Its performance is more limited, as shown in Table \ref{tab:str_synth_fintuning}, despite being a ViT trained with a similar method (MIM) to the other ViTs. One possible cause for this is that it has been trained with only 4M data, compared to the 32.77M of pre-training data and 17M of fine-tuning data used in the DiG, MaskOCR, or CCD methods. Furthermore, the lower performance of CNN architectures may also be influenced by the fact that they have generally been trained with smaller amounts of data. This disparity in training data volume further emphasizes the need for standardized evaluation protocols to ensure fair comparisons across different models and methodologies.

Our analysis reveals that, although the most recent methods define the current state of the art, they benefit from larger amounts of data, resulting in more robust and generalizable models. Therefore, older methods should not be overlooked, as they offer very competitive performances and provide valuable knowledge and insights that contribute to the development and understanding of SSL in STR. Recognizing the contributions of these earlier methods is essential for advancing the field and ensuring a comprehensive understanding of its evolution.

\subsubsection{\review{Self-Supervised vs. Supervised Learning in STR}}

\review{Based on the above and with the aim of providing a fair analysis of the actual contribution of the different SSL-STR methods, we now analyze the improvement they provide over those models trained from scratch when all other conditions are maintained---i.e. architecture and training data.}

\begin{figure}[!tb]
  \centering
  \begin{subfigure}{\linewidth}
    \centering
    \includegraphics[width=\linewidth]{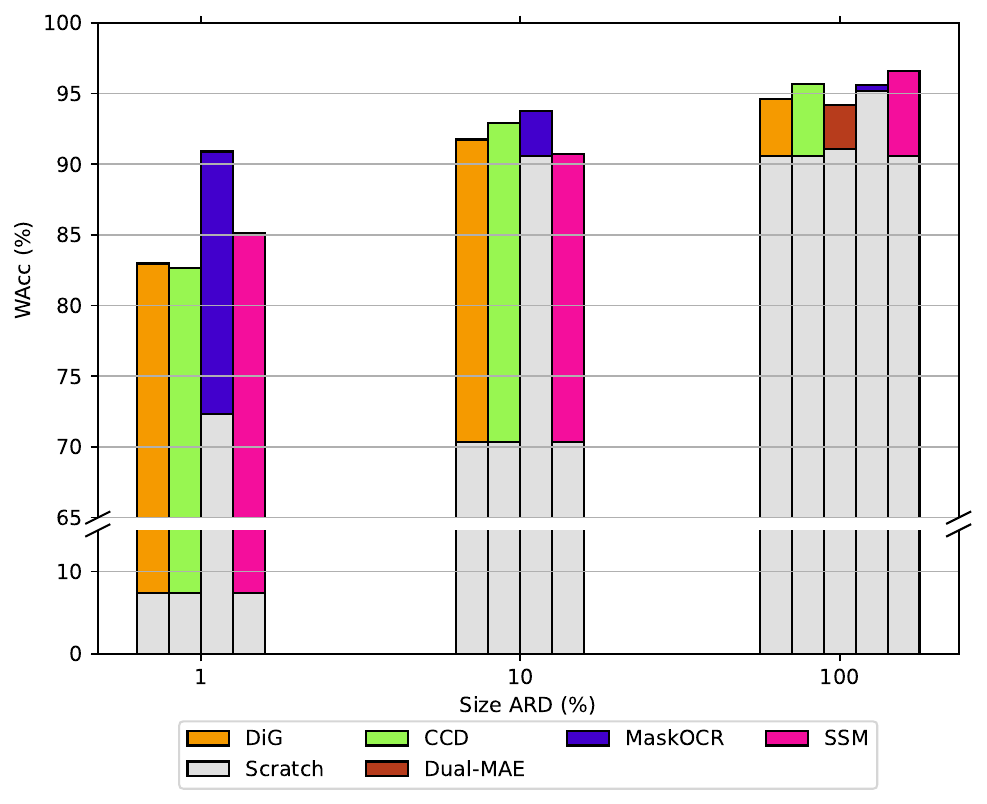}
    \caption{\review{Fine-tuning on the ARD dataset using 1\%, 10\% and 100\% of the collection.}}
    \label{fig:sup_vs_ssl_ard}
  \end{subfigure}
  \hfill
  \begin{subfigure}{\linewidth}
    \centering
    \includegraphics[width=\linewidth]{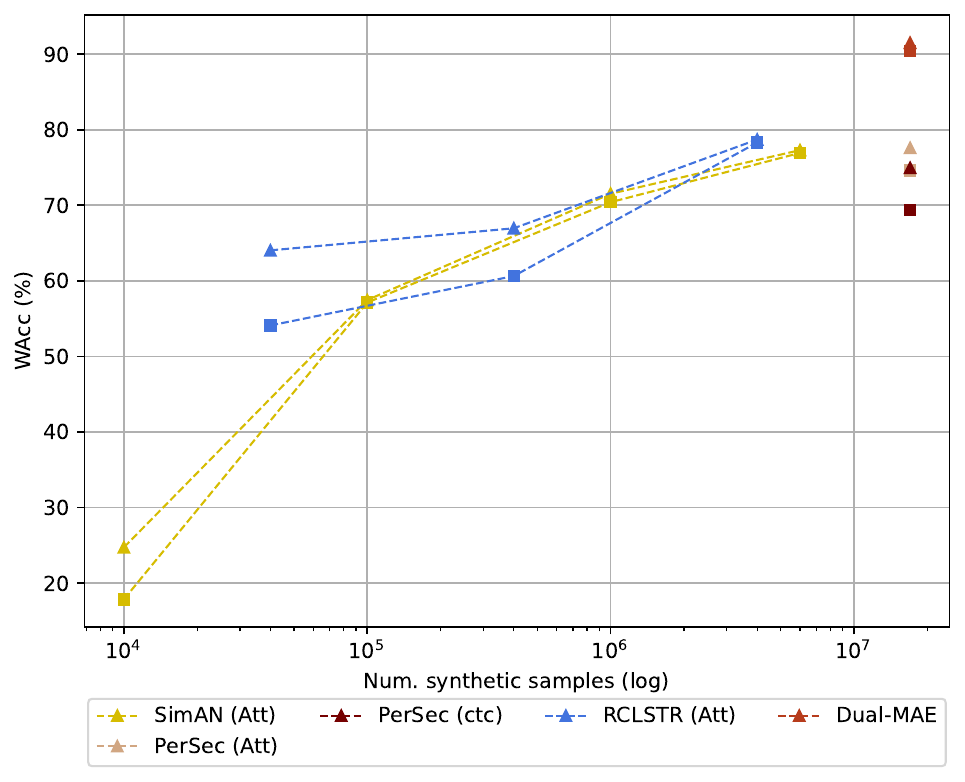}
    \caption{\review{Fine-tuning on different quantities of synthetic data. The triangle and square markers respectively indicate pre-trained and from scratch models. Dashed lines connect different sizes of data used for the same model.}}
    \label{fig:sup_vs_ssl_synth}
  \end{subfigure}
  \caption{\review{Comparison in terms of WAcc of the methods that present results under the fine-tuning evaluation protocol and those starting from scratch for the average of the six common STR benchmarks. Dual-MAE represents the biggest value between Dual-MAE-Siam and Dual-MAE-Dist.}}
  \label{fig:sup_vs_ssl_synth_sup_vs_ssl_real}
\end{figure}

\review{First, Fig. \ref{fig:sup_vs_ssl_synth_sup_vs_ssl_real} illustrates the improvement observed across the average of the six common STR benchmarks when the pre-training methods are applied to the recognition schemes compared to those trained from scratch. In Fig. \ref{fig:sup_vs_ssl_ard} these models were fine-tuned using different percentages of the ARD real data collection. As shown, there is a notable difference between the models trained from scratch and those initialized with pre-trained weights, with the gap being especially pronounced when only a small percentage of the dataset is used. This insight aligns with previous findings, which suggests that SSL proves to be remarkably effective in scenarios with limited annotated data, thus reducing labeling costs.}

\review{In contrast, the schemes in Fig. \ref{fig:sup_vs_ssl_synth} were fine-tuned using varying amounts of synthetic data instead of real data. While pre-training still outperforms scenarios in which the models are trained from scratch with the same amount of synthetic data, the improvements observed in this case are more modest than those when using real data. This highlights that the true potential of SSL is best exploited when fine-tuning on real data. Starting with a model that already deeply understands the visual and semantic patterns on a given data distribution allows for greater efficiency when using real data, capitalizing the more complexity and variability it offers.}

\subsubsection{Synthetic vs. real data}

\begin{figure}[!tb]
  \centering
  \includegraphics[width=\linewidth]{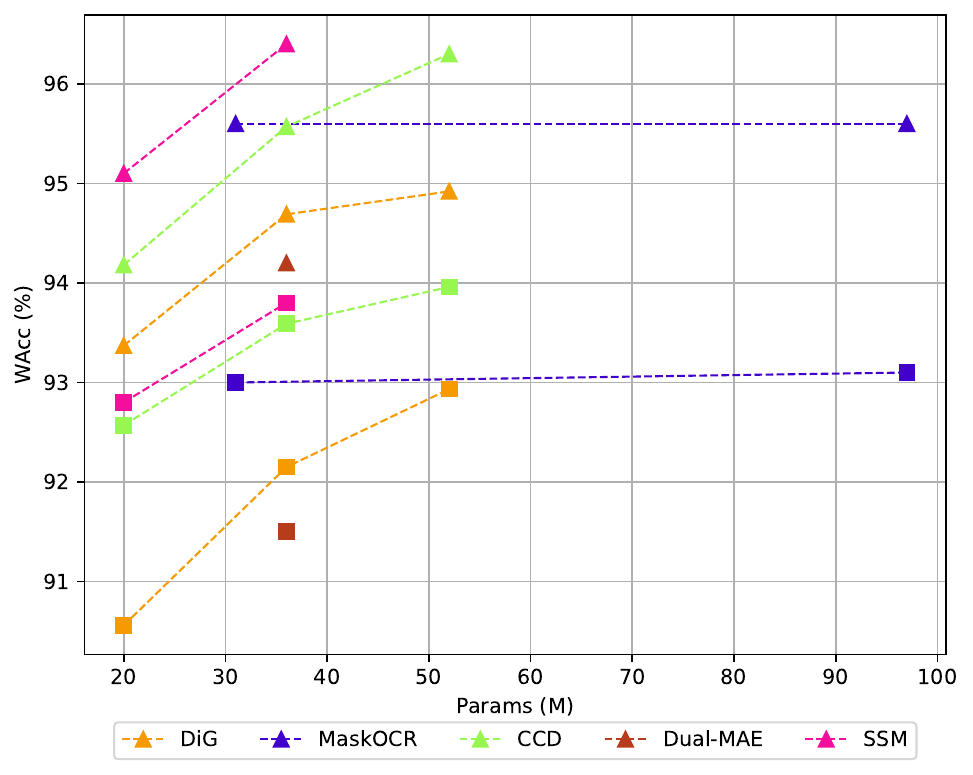}
  \caption{Comparison in terms of WAcc of the methods presenting results under the fine-tuning evaluation protocol for the average of the STR benchmarks, using either real or synthetic data for fine-tuning. More specific details about \review{results and data configurations are found in Table \ref{tab:str_real_fintuning} and Table \ref{tab:str_synth_fintuning}, and Table \ref{tab:datasets_for_methods}}. Square and triangle markers respectively indicate models fine-tuned with synthetic and real data. Dashed lines connect different sizes of the same model.}
  \label{fig:real_vs_synth}
\end{figure}

\review{As previously mentioned,} one of the main purposes of SSL is to leverage large amounts of unlabeled real data in order to avoid the intensive labor of annotation or resorting to alternatives such as \review{extensive training in} synthetic data. This is first discussed in the work of \cite{baek2021if}, which claims that synthetic data often lacks the variability and complexity present in real STR data\review{, insight that is also reinforced by the results presented in the previous section of this work}. Figure \ref{fig:real_vs_synth} \review{shows the average performance on the six common STR benchmarks for the models that, after pre-training, are fine-tuned on either synthetic or real data}. \review{In that Figure it is illustrated that the previous idea} is supported by the operation of all SSL methods, for which a better performance is clearly attained by using real data \review{in the fine-tuning}, even at the expense of using much less data \review{than synthetic ones, as shown in Table \ref{tab:datasets_for_methods}. For more comprehensive information about the results when fine-tuning on real data see Table \ref{tab:str_real_fintuning}.} 

As can be observed, the use of SSL for STR effectively achieves the goal of improving the models' ability to generalize under real-world conditions, while also reducing the need for annotated data, which is expensive and difficult to obtain.

\begin{table*}[t]
\centering
\caption{\review{Comparison in terms of WAcc of the methods presenting results under the fine-tuning evaluation protocol for the six common STR benchmarks. After pre-training, all method's models are fine-tuned with real data. More specific details about data configurations is found in Table \ref{tab:datasets_for_methods}. Inside each architectural type the methods are arranged chronologically. For each model it is assumed TD unless specified other architectural configuration in parenthesis. In STR-CPC, CRNN and TRBA indicate their different architectural configurations.}}
\label{tab:str_real_fintuning}
\begin{tabular}{lllccccccccccc}
\toprule[1pt]
\textbf{} & \multirow{3}{*}{\textbf{Model}} &  & \multicolumn{4}{c}{\textbf{Regular}} &  & \multicolumn{4}{c}{\textbf{Irregular}} &  &  \\ \cline{4-7} \cline{9-12}
 &  &  & \multirow{2}{*}{\textbf{IIIT}} & \multirow{2}{*}{\textbf{SVT}} & \multicolumn{2}{c}{\textbf{IC13}} &  & \multirow{2}{*}{\textbf{SP}} & \multicolumn{2}{c}{\textbf{IC15}} & \multirow{2}{*}{\textbf{CT}} &  &  \\ \cline{6-7} \cline{10-11}
 &  &  &  &  & 857 & 1015 &  &  & 1811 & 2077 &  & \textbf{Avg} & \textbf{Par.} \\ \cline{1-2} \cline{4-14} 
\multicolumn{2}{l}{CNN-based} &  &  &  &  &  &  &  &  &  &  &  &  \\
 & STR-CPC (CRNN) &  & 84.7 & 79.6 & \multicolumn{2}{c}{88.7} &  & 64.3 & \multicolumn{2}{c}{64.7} & 68.1 & 77.0 & 8M \\
 & STR-CPC (TRBA) &  & 93.4 & 88.1 & \multicolumn{2}{c}{93.6} &  & 79.5 & \multicolumn{2}{c}{76.9} & 88.2 & 87.2 & 50M \\ \hdashline
\multicolumn{2}{l}{ViT-based} &  &  &  &  &  &  &  &  &  &  &  &  \\
 & $\text{DiG}_{\text{Tiny}}$ &  & 96.4 & 94.4 & - & 96.2 &  & 90.2 & 87.4 & - & 94.1 & 93.1 & 20M \\
 & $\text{DiG}_{\text{Small}}$ &  & 97.7 & 96.1 & - & 97.3 &  & 91.6 & 88.6 & - & 96.2 & 94.6 & 36M \\
 & $\text{DiG}_{\text{Base}}$ &  & 97.6 & 96.5 & - & 97.6 &  & 92.9 & 88.9 & - & 96.5 & 95.0 & 52M \\
 & $\text{CCD}_{\text{Tiny}}$ &  & 97.1 & 96.0 & - & 97.5 &  & 91.6 & 87.5 & - & 95.8 & 94.2 & 20M \\
 & $\text{CCD}_{\text{Small}}$ &  & 98.0 & 96.4 & - & 98.3 &  & 92.7 & 90.3 & - & 98.3 & 95.6 & 36M \\
 & $\text{CCD}_{\text{Base}}$ &  & 98.0 & 97.8 & - & 98.3 &  & 96.1 & 91.6 & - & 98.3 & 96.3 & 52M \\
 & Dual-MAE-Siam &  & 97.8 & 96.6 & 97.2 & 97.2 &  & 93.0 & 90.2 & 89.0 & 96.5 & 94.2 & 36M \\
 & Dual-MAE-Dist &  & 97.8 & 96.3 & 97.1 & 97.2 &  & 93.0 & 89.7 & 88.7 & 97.9 & 94.0 & 36M \\
 & $\text{MaskOCR}_{\text{Small}}$ &  & 98.0 & 96.9 & 97.8 & - &  & 94.9 & 90.2 & - & 96.2 & 95.6 & 31M \\
 & $\text{MaskOCR}_{\text{Base}}$ &  & 98.0 & 96.9 & 98.2 & - &  & 94.6 & 90.1 & - & 95.8 & 95.6 & 97M \\
 & $\text{SSM}_{\text{Tiny}}$ &  & 98.1 & 96.1 & - & 97.8 &  & 92.6 & - & 89.0 & 96.5 & 95.1 & 20M \\
 & $\text{SSM}_{\text{Small}}$ &  & 98.9 & 98.0 & - & 98.5 &  & 95.0 & - & 90.8 & 98.3 & 96.4 & 36M \\
\bottomrule[1pt]
\end{tabular}
\end{table*}

\subsubsection{Reaching a performance ceiling}
\label{sec:Reaching_a_Performance_Ceiling}
Despite the remarkable progress achieved in the field, it appears that results are approaching a performance glass ceiling concerning the six traditional STR benchmarks. The datasets, once considered highly challenging a decade ago, now seem somewhat easy to address. This might be explained by the appearance of new architectures such as the Transformer and the development and application of SSL, which allows to take advantage of real unlabeled data. Therefore, these benchmarks may no longer sufficiently represent the evolving complexities of real-world text recognition tasks \citep{jiang2023revisiting}. For example, considering the most recent methods in Table \ref{tab:str_real_fintuning}, there is hardly any difference in performance when fine-tuning with real data.

Given the advancements and the saturation in performance on existing benchmarks, one takeaway of our analysis is that the focus must be shifted towards more comprehensive and challenging datasets, such as Union14M-Benchmark \citep{jiang2023revisiting}. Union14M-Benchmark offers a broader spectrum of TR challenges, capturing a wider variety of real-world scenarios and complexities. This shift is essential to continue pushing the boundaries of what STR systems can achieve, ensuring models that remain robust and effective in increasingly diverse and unpredictable environments.


\subsection{Analysis of SSL performance for HTR}

Although to a lesser extent, the development of SSL also impacted the field of HTR significantly. Table \ref{tab:iam_cvl_rimes_results} shows the SSL-TR methods that present results for IAM, CVL \review{and RIMES} datasets. For instance, it can be appreciated that the performance on the well-known IAM dataset ranges from the first model (SeqCLR) to the best-performing one (DiG) by approximately 14 units of WAcc.

\begin{table*}[!t]
\centering
\caption{\review{Comparison in terms of WAcc of the methods that report results under the fine-tuning evaluation protocol for IAM, CVL and RIMES datasets considering different percentages of labeled data.}}
\label{tab:iam_cvl_rimes_results}
\begin{tabular}{lllccccccccccclc}
\toprule[1pt]
 & \multirow{2}{*}{\textbf{Method}} &  & \multicolumn{3}{c}{\textbf{IAM}} &  & \multicolumn{3}{c}{\textbf{CVL}} &  & \multicolumn{3}{c}{\textbf{RIMES}} &  &  \\ \cline{4-6} \cline{8-10} \cline{12-14}
 &  &  & 5 & 10 & 100 &  & 5 & 10 & 100 &  & 5 & 10 & 100 &  & \textbf{Par.} \\ \cline{1-2} \cline{4-16} 
\multicolumn{2}{l}{CTC decoder} &  &  &  &  &  &  &  &  &  &  &  &  &  &  \\
 & SeqCLR &  & 31.2 & 44.9 & 76.7 &  & 66.0 & 71.0 & 77.0 &  & 61.8 & 71.9 & 90.1 &  & 48M \\
 & PerSec-CNN &  & - & - & 77.9 &  & - & - & 78.1 &  & - & - & - &  & - \\
 & PerSec-ViT &  & - & - & 78.0 &  & - & - & 78.8 &  & - & - & - &  & - \\
 & ChaCo &  & 44.3 & 53.1 & 79.8 &  & 65.6 & 70.3 & 78.1 &  & 61.3 & 73.0 & 89.2 &  & 48M \\
 & CMT-Co &  & 47.7 & 56.0 & 80.1 &  & 71.9 & 74.5 & 78.2 &  & 67.1 & 75.4 & 90.1 &  & 48M \\
 & Sorting &  & 56.4 & 64.7 & 79.0 &  & 31.8 & 34.7 & 78.5 &  & 74.1 & 71.5 & 90.0 &  & 48M \\
 & Flip &  & 61.7 & 67.4 & 79.1 &  & 40.6 & 54.6 & 85.5 &  & 79.8 & 84.2 & 91.1 &  & 48M \\ \hdashline
\multicolumn{2}{l}{Att decoder} &  &  &  &  &  &  &  &  &  &  &  &  &  &  \\
 & SeqCLR &  & 40.3 & 52.3 & 79.9 &  & 73.1 & 74.8 & 77.8 &  & 70.9 & 77.0 & 92.5 &  & 49M \\
 & PerSec-CNN &  & - & - & 80.8 &  & - & - & 80.2 &  & - & - & - &  & - \\
 & PerSec-ViT &  & - & - & 81.8 &  & - & - & 80.8 &  & - & - & - &  & - \\
 & ChaCo &  & 46.7 & 55.5 & 81.4 &  & 68.9 & 74.0 & 78.2 &  & 68.4 & 77.0 & 90.6 &  & 49M \\
 & CMT-Co &  & 50.4 & 55.8 & 81.9 &  & 73.6 & 76.2 & 78.7 &  & 68.8 & 77.6 & 91.2 &  & 49M \\ \hdashline
\multicolumn{2}{l}{TD decoder} &  &  &  &  &  &  &  &  &  &  &  &  &  &  \\
 & $\text{DiG}_{\text{Tiny}}$ &  & - & - & 83.2 &  & - & - & 87.4 &  & - & - & - &  & 20M \\
 & $\text{DiG}_{\text{Small}}$ &  & - & - & 85.7 &  & - & - & 90.5 &  & - & - & - &  & 36M \\
 & $\text{DiG}_{\text{Base}}$ &  & - & - & 87.0 &  & - & - & 91.3 &  & - & - & - &  & 52M \\
 & Text-DIAE &  & 49.6 & 58.7 & 80.0 &  & 47.9 & 68.5 & 87.3 &  & - & - & - &  & 68M \\
 & Sorting &  & 59.2 & 69.2 & 80.8 &  & 43.4 & 54.2 & 86.2 &  & 80.4 & 83.9 & 91.7 &  & 50M \\
 & Flip &  & 62.5 & 69.8 & 80.8 &  & 52.1 & 66.3 & 91.3 &  & 79.0 & 83.9 & 89.7 &  & 50M \\ \bottomrule[1pt]
\end{tabular}
\end{table*}


Furthermore, unlike in STR, the current HTR datasets are still challenging, since they present considerable room for improvement. When using a small percentage of the data for fine-tuning, this is even more pronounced. Table \ref{tab:iam_cvl_rimes_results} shows the methods that present the results when using 5\% and 10\% of labeled data from the IAM dataset, barely exceeding a performance of 50\%.


\review{It is also worth highlighting that the results reported show} a dominance of architectures that use Att or TD as decoders. Incorporating a language model can be especially beneficial since it allows a more explicit understanding of the word being recognized as a whole. This is evidenced in the work of \cite{penarrubia2024spatial}, where simply replacing CTC with TD considerably increases the WAcc.

\subsubsection{What is wrong when comparing SSL methods for HTR?}
\label{What_is_wrong_with_HTR}
Once again, strict comparison in HTR can lead to erroneous conclusions due to the disparity data conditions in which the different proposals are trained. In SeqCLR, the first SSL method for HTR, the data settings are not explicitly provided, \review{neither in PerSec nor in Text-DIAE}. This leads to other methods using different data conditions for training. For example, DiG mentions that it merges the IAM and CVL sets for training the model. However, ChaCo mentions that, after consulting the authors of SeqCLR, only the training sets corresponding to each evaluated dataset are used for pre-training. Therefore, they reproduce that configuration. This is followed by the works of \cite{zhang2022cmt,penarrubia2024spatial}.

It is obvious that merging different datasets for pre-training has a significant impact on the performance of the model. While DiG uses 146,805 images for training, in works such SeqCLR and such ChaCo around 74000 are used for the IAM and CVL models. This makes the comparison with available results unfair, since merging datasets introduces a greater diversity and amount of data, thus allowing a model to learn better. Therefore, it is not known how much of the improvement offered by DiG is due to the use of ViT and the SSL method itself and how much is due to the greater amount of data, so the current method that defines the baseline to be outperformed is not so indicative. It is, therefore, crucial that future research efforts explicitly detail their training configurations. Furthermore, a standard protocol describing how to use the datasets should be established to ensure a fair and reproducible evaluation.

\subsubsection{\review{Self-Supervised vs. Supervised Learning in HTR}}

\begin{figure}[!htb]
  \centering
  \begin{subfigure}{\linewidth}
    \centering
    \includegraphics[width=\linewidth]{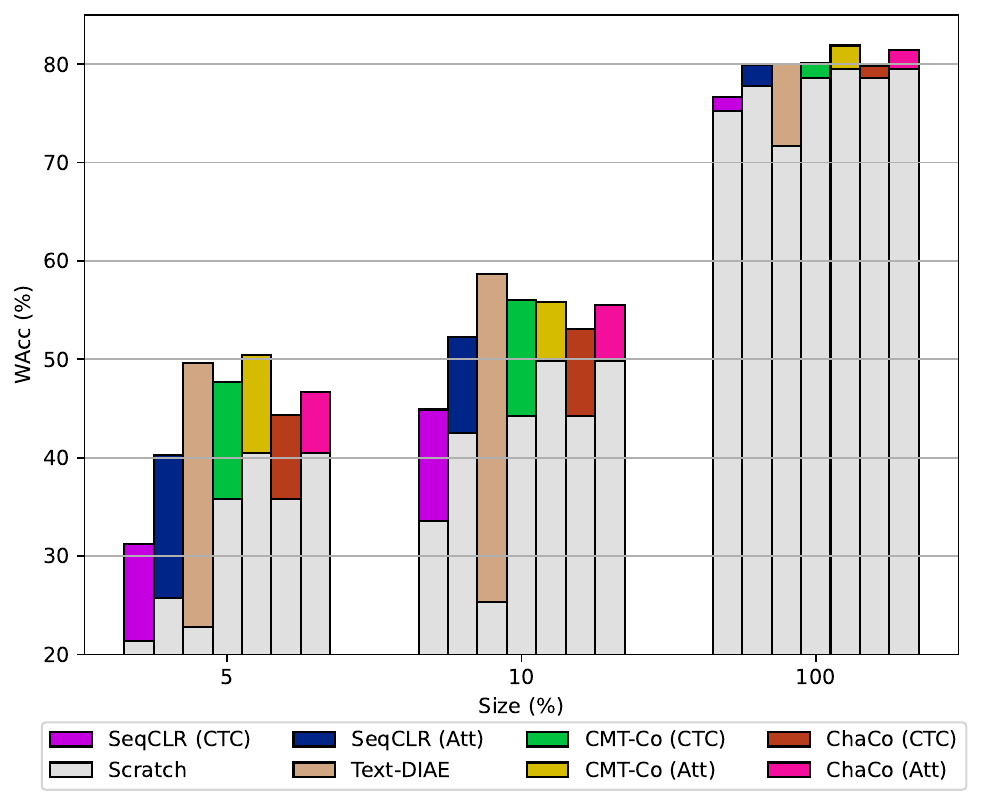}
    \caption{\review{IAM dataset.}}
    \label{fig:iam_ssl_vs_scratch}
  \end{subfigure}
  \hfill
  \begin{subfigure}{\linewidth}
    \centering
    \includegraphics[width=\linewidth]{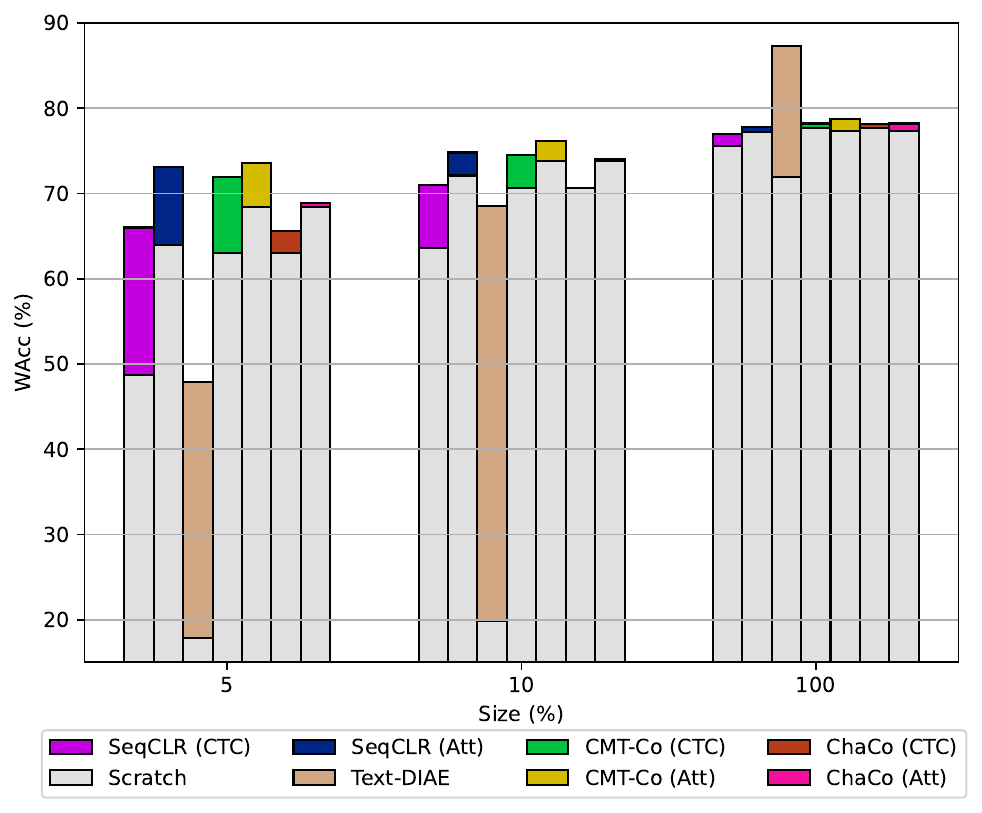}
    \caption{\review{CVL dataset.}}
    \label{fig:cvl_ssl_vs_scratch}
  \end{subfigure}
  \caption{\review{Comparison in terms of WAcc of the methods that present results under the fine-tuning evaluation protocol and starting from scratch for IAM and CVL datasets when considering 5\%, 10\% and 100\% of data.}}
  \label{fig:htr_ssl_vs_scratch}
\end{figure}

\review{Since each model differs in both architecture and training data, this section focuses on comparing the improvements achieved by different methods over their corresponding architectures trained from scratch.}

\review{Figure \ref{fig:htr_ssl_vs_scratch} illustrates the performance gains provided by each method compared to training the same model from scratch on the IAM and CVL datasets, when such information is available. Notably, methods leveraging pre-trained models demonstrate a substantial improvement. Moreover, also in the HTR domain, this improvement becomes even more pronounced as the amount of training data decreases. This finding is particularly valuable for HTR, given the vast diversity of languages---both modern and historical---as well as script variations, making it a crucial step toward reducing the high costs associated with manual annotation.}

\subsection{\review{Comparison with general Self-Supervised Learning methods}}
\label{sec:comparison_general-purpose}


\review{This survey collects and analyzes the SSL methods specifically created for TR. However, as introduced in Section \ref{sec:Introduction}, SSL has a much longer history, with a great development of general SSL methods in fields such as computer vision or natural language processing. One alternative would be to directly use the pre-trained models obtained from these fields and adapt them through large-scale fine-tuning to STR or HTR. In this section, we will analyze the comparison of general pre-trained models against SSL-TR methods.}

\begin{table*}[t]
\centering
\caption{\review{Comparison of large-scale pre-trained models starting from general SSL methods against the last three state-of-the-art SSL-STR methods. In first part of the Table the WAcc of the large-scale pre-trained models is shown. In the second part, the relative WAcc gain between the SSL-STR models from Table \ref{tab:str_real_fintuning} and the general models listed above is reported. For both scenarios, the quantity of fine-tuning data is reported (\emph{Bench} refers to using the training set of the six STR benchmarks). For MaskOCR, the best value between the small and the base model is chosen.}}
\label{tab:general_ssl_vs_str_ssl}
\begin{tabular}{llcclccccccccc}
\hline
 & \multirow{3}{*}{\textbf{Model}} & \multicolumn{2}{c}{\multirow{2}{*}{\textbf{\begin{tabular}[c]{@{}c@{}}Fine-tuning\\ data\end{tabular}}}} &  & \multicolumn{4}{c}{\textbf{Regular}} &  & \multicolumn{4}{c}{\textbf{Irregular}} \\ \cline{6-9} \cline{11-14} 
\textbf{} &  & \multicolumn{2}{c}{} &  & \multirow{2}{*}{\textbf{IIIT}} & \multirow{2}{*}{\textbf{SVT}} & \multicolumn{2}{c}{\textbf{IC13}} &  & \multirow{2}{*}{\textbf{SP}} & \multicolumn{2}{c}{\textbf{IC15}} & \multirow{2}{*}{\textbf{CT}} \\ \cline{3-4} \cline{8-9} \cline{12-13}
\textbf{} &  & \textbf{Synth} & \textbf{Real} &  &  &  & 857 & 1015 &  &  & 1811 & 2077 &  \\ \hline
\multicolumn{2}{l}{\emph{General}} &  &  &  & \multicolumn{9}{c}{WAcc} \\ \cline{6-14} 
 & $\text{TrOCR}_{\text{Base}}$ & 722M & Bench &  & 93.4 & 95.2 & 98.4 & 97.4 &  & 92.1 & 86.9 & 81.2 & 90.6 \\
 & $\text{TrOCR}_{\text{Large}}$ & 722M & Bench &  & 94.1 & 96.1 & 98.4 & 97.3 &  & 93.0 & 88.1 & 84.1 & 95.1 \\
 & DTrOCR & 10B & - &  & 98.4 & 96.9 & 98.8 & 97.8 &  & 95.0 & 92.3 & 90.4 & 97.6 \\
 & DTrOCR & 10B & 1.5M &  & 99.6 & 98.9 & 99.1 & 99.4 &  & 98.6 & 93.5 & 93.2 & 99.1 \\ \hline \hline
\multicolumn{2}{l}{\emph{SSL-STR}} &  &  &  & \multicolumn{9}{c}{$\Delta$WAcc} \\ \cline{6-14} 
 & \multirow{4}{*}{MaskOCR} & \multirow{4}{*}{-} & \multirow{4}{*}{2.78M} &  & \green{+}4.6 & \green{+}1.7 & \multicolumn{1}{r}{\red{-}0.2} & - &  & \green{+}2.8 & \green{+}3.3 & - & \green{+}5.6 \\
 &  &  &  &  & \green{+}3.9 & \green{+}0.8 & \multicolumn{1}{r}{\red{-}0.2} & - &  & \green{+}1.9 & \green{+}2.1 & - & \green{+}1.1 \\
 &  &  &  &  & \red{-}0.4 & 0 & \red{-}0.6 & - &  & \red{-}0.1 & \red{-}2.1 & - & \red{-}1.4 \\
 &  &  &  &  & \red{-}1.6 & \red{-}2 & \red{-}0.9 & - &  & \red{-}3.7 & \red{-}3.3 & - & \red{-}2.9 \\ \hdashline 
 & \multirow{4}{*}{$\text{CCD}_{\text{Base}}$} & \multirow{4}{*}{-} & \multirow{4}{*}{2.78M} &  & \green{+}4.6 & \green{+}2.6 & - & \green{+}0.9 &  & \green{+}4 & \green{+}4.7 & - & \green{+}7.7 \\
 &  &  &  &  & \green{+}3.9 & \green{+}1.7 & - & \green{+}1 &  & \green{+}3.1 & \green{+}3.5 & - & \green{+}3.2 \\
 &  &  &  &  & \red{-}0.4 & \green{+}0.9 & - & \green{+}0.5 &  & \green{+}1.1 & \red{-}0.7 & - & \green{+}0.7 \\
 &  &  &  &  & \red{-}1.6 & \red{-}1.1 & - & \red{-}1.1 &  & \red{-}2.5 & \red{-}1.9 & - & \red{-}0.8 \\ \hdashline
 & \multirow{4}{*}{$\text{SSM}_{\text{Small}}$} & \multirow{4}{*}{-} & \multirow{4}{*}{2.78M} &  & \green{+}5.5 & \green{+}2.8 & - & \green{+}1.1 &  & \green{+}2.9 & - & \green{+}9.6 & \green{+}7.7 \\
 &  &  &  &  & \green{+}4.8 & \green{+}1.9 & - & \green{+}1.2 &  & \green{+}2 & - & \green{+}6.7 & \green{+}3.2 \\
 &  &  &  &  & \green{+}0.5 & \green{+}1.1 & - & \green{+}0.7 &  & 0 & - & \green{+}0.4 & \green{+}0.7 \\
 &  &  &  &  & \red{-}0.7 & \red{-}0.9 & - & \red{-}0.9 &  & \red{-}3.6 & - & \red{-}2.4 & \red{-}0.8 \\ \hline
\end{tabular}
\end{table*}

\review{Specifically, we compare the performance of TrOCR \citep{li2023trocr} and DTrOCR \citep{fujitake2024dtrocr}. TrOCR employs an encoder-decoder architecture, testing two image-based encoders---BEiT \citep{bao2021beit} and DeiT \citep{touvron2021training}---and two language models as decoders---RoBERTa \citep{liu2019roberta} and MiniLM \citep{wang2020minilm}---both fine-tuned on hundreds of millions of TR synthetic samples. In contrast, DTrOCR leverages a Generative Pre-trained Transformer \citep{yenduri2023generative} model, further fine-tuned on several billion synthetic samples to adapt it also for TR. Then, Table \ref{tab:general_ssl_vs_str_ssl} presents a comparison of these models with the three most recent state-of-the-art SSL-STR methods.}

\review{A key insight from this comparison is that SSL-STR methods achieve comparable, or even superior, performance to the large-scale pre-trained models. Despite the latter being derived from general SSL models and fine-tuned on massive synthetic datasets, SSL-STR methods can reach similar levels of accuracy with less than 3M real samples, greatly reducing the computational and data requirements for training. This highlights the critical importance of developing SSL techniques specifically tailored for TR. Furthermore, it reinforces the superiority of real data over synthetic elements, even when synthetic datasets are scaled to enormous sizes.}

\review{Regarding the SSL-HTR methods, it must be noted that the large-scale pre-trained methods we compare with are devised to work at the line level, while the ones assessed in this work do so at the word level. Furthermore, the large pre-trained models recognize only alphanumeric characters, thus modifying the ground-truth of the HTR datasets, while in HTR it is common to keep the original charset of the dataset. All this makes it impossible to establish a fair comparison. These issues will be further discussed in Section \ref{sec:Future_research_of_SSL_for_HTR}.}

\section{Conclusions}
\label{sec:conclusions}
We have reviewed the existing literature \review{on} Self-Supervised Learning (SSL) applied to Text Recognition (TR) with the aim of describing, analyzing, and comparing the different existing alternatives in the field. Additionally, we have highlighted inconsistencies in the comparisons of these methods to promote more precise benchmarking and enable the scientific community to more clearly identify the real improvements provided by new approaches. Below we discuss different topics that are yet to be highlighted.

\subsection{\review{Why develop Self-Supervised methods for TR?}}


\review{In this work, we traced the evolution of SSL methods for TR and observed that early advancements were largely influenced by developments in other fields, such as image classification. However, continuous progress in state-of-the-art SSL-TR models has highlighted the importance of addressing the unique characteristics of TR images. For example, recent methods have increasingly focused on leveraging both the visual information of individual characters and the semantic relationships or sequential information among characters within an image, giving rise to a specialized subfield of SSL tailored specifically for TR.}

\review{These SSL-TR methods have demonstrated remarkable effectiveness in optimizing the use of fine-tuning datasets, which is especially beneficial in data-scarce scenarios, greatly reducing the manual cost of labeling. Furthermore, they have shown a significant advantage when fine-tuned on real datasets, as they better capture the intrinsic and distinctive patterns of real-world TR images. This allows them to fully exploit the complexity and diversity present in such datasets.}

\review{This analysis culminates in Section \ref{sec:comparison_general-purpose}, where we compare large-scale pre-trained models---originating from general SSL approaches---with the latest state-of-the-art SSL-STR models. Notably, despite the former being fine-tuned on hundreds of millions or even billions of synthetic samples, SSL-STR methods achieve comparable performance while requiring significantly fewer training images---less than 3M. By starting from models pre-trained with TR images we avoid the massive use of synthetic data to adapt general SSL models to the specific TR domain, thus greatly reducing the computational cost. This analysis underscores the value of developing TR-specific SSL techniques, as they are uniquely suited to extract the underlying information embedded in TR images, ultimately enhancing in TR models the efficiency of training data and efficacy in real-world scenarios.}

\subsection{Current trends}
In this subsection, we review the historical development of SSL, highlighting significant milestones and trends that have shaped the field. We also discuss how these trends are reflected in the evolution of SSL methods for TR.

SSL has a long history of development. In its early days, discriminative learning was primarily used, such as those based on spatial context \citep{doersch2015unsupervised,noroozi2016unsupervised} or geometric pretext tasks \citep{gidaris2018unsupervised}. Generative approaches like inpainting also emerged early and demonstrated their effectiveness \citep{pathak2016context,xiang2023deep}. Then, there was a paradigm shift towards joint-embedding architectures that search for similarities between views of the same image. Such frameworks avoid the collapse problem in various ways as, for instance, the use of negative pairs in contrastive learning \citep{he2020momentum,chen2020simple}, regularizing terms in information-maximization \citep{zbontar2021barlow,bardes2021vicreg,bardes2022vicregl,tong2023emp}, or different architectures in distillation schemes \citep{grill2020bootstrap,chen2021exploring,caron2021emerging}. Recently, with the emergence of Vision Transformers (ViT), Masked Image Modeling (MIM) has gained considerable attention \citep{zhang2022survey}, also in combination with other paradigms \citep{ozbulak2023know}.

This evolution of SSL is also reflected in the field of TR. Early methods, such as SeqCLR \citep{aberdam2021sequence}, PerSec \citep{liu2022perceiving}, and SimAN \citep{luo2022siman}, were framed within the contrastive and generative learning fields, which represented the state of the art at the time. Other works later explored the use of MIM and ViT architectures \citep{souibgui2023text,lyu2023maskocr,guan2023self}. The most recent works explore the combination of different SSL working principles \citep{yang2022reading,gao2024self}.

\subsection{Open questions}
This subsection addresses the gaps and challenges within the current SSL landscape for TR. We identify unexplored SSL categories, the need for a solid theoretical understanding, issues related to efficiency, and the importance of standardized benchmarking protocols.

\subsubsection{Unexplored SSL categories}
Despite the significant progress in this field over a short period, many SSL areas remain unexplored for TR \citep{penarrubia2024spatial}. One might think that, due to the recent dominance of MIM, other fields hold little interest. However, as previously mentioned, there is a current trend to combine different paradigms. Thus, even if new proposals do not strictly surpass the current state of the art, they should be disseminated as these contributions could be useful in the future to inspire new methods or even be combined with other methodologies.

Unexplored SSL subcategories for TR include clustering \citep{caron2018deep,caron2020unsupervised,qian2022unsupervised}, information-maximization \citep{zbontar2021barlow,bardes2021vicreg,tong2023emp}, use of GAN \citep{goodfellow2014generative,radford2015unsupervised,yu2021vector}, and inpainting \citep{pathak2016context,pajot2019unsupervised,elharrouss2020image}. For example, clustering, although being one of the oldest SSL methods and seemingly less relevant now, has been recently considered as a combination with contrastive learning \cite{zhang2021supporting,jiang2023group}. Therefore, it is crucial to explore and develop all SSL categories for TR.

\subsubsection{Theoretical understanding}
A solid theoretical foundation is essential to guide the development and application of SSL methods. While image classification has seen extensive research into how different paradigms learn data representations \citep{caron2021emerging,zhang2022does,garrido2022duality,chen2022bag,zhang2022dual,ben2023reverse}, TR remains largely unexplored in those terms. Note that, understanding how different models learn or what they focus on after applying different SSL categories is crucial for the overall advance of the field.

In TR, there are two main levels of information: visual (recognizing a character by its shape) and semantic (recognizing a character by its context). It would be beneficial to understand how a DNN focuses on or distinguishes between these pieces of information and how they influence SSL training. So far, only PerSec has shown the differences between these two levels of information by visualizing the embeddings produced in different network layers.

Additionally, theoretical understanding should help clarify contradictory conclusions drawn in different works. For example, in contrastive learning, the search for the unit to calculate the contrastive loss is a constant issue. SeqCLR focuses on subword level, but works like CMT-Co or ChaCo argue that subword level can break character or semantic continuity, suggesting character level is more suitable. Conversely, PerSec suggests negative pairs across different samples can break semantic continuity, while RCLSTR encourages rearrangement and contrastive loss calculation across samples. To validate these assumptions, studying how these issues influence representation quality remains as a clear need in the field.

\subsubsection{Efficient SSL}
As mentioned by \cite{ozbulak2023know}, current SSL methods require substantial computational resources. The latest trend of combining SSL methods to leverage their benefits further increases these complexities. For instance, MoCo v3 requires 625 TPU days and SimMIM claims to be 1.8 times more efficient than MoCo v3. DiG combines MoCo v3 with SimMIM, adding the computational costs of both methods. Executing such methods on large data volumes requires significant computational capacity. Additionally, TR methods often consider both visual and semantic information, resulting in more extensive and time-consuming methods.

Thus, it is essential to investigate and implement techniques for efficient training and evaluation \citep{ciga2021resource,baevski2023efficient} and value simpler methods requiring less computational power to democratize the use of SSL in TR, as done for specific SSL working principles \citep{wang2021solving,li2021efficient}.

\subsubsection{Standard evaluation protocols}
A significant challenge in comparing different SSL methods for TR is the inconsistency in the data used for training, a topic that we have substantially covered in this survey. Standardizing datasets and clear documentation on how they are used in the experiments is imperative. Some studies use a combination of multiple datasets for training, while others use isolated datasets, leading to incomparable results. Establishing common benchmarks and protocols will enable more accurate and fair comparisons, allowing to identify actual advancements in the field.

\subsection{Future research}

In this subsection, we outline potential areas for future research in SSL for both Scene Text Recognition (STR) and Handwritten Text Recognition (HTR). We discuss the challenges, opportunities, and potential impacts of advancing SSL methods in these fields.

\subsubsection{Future research of SSL for STR}
SSL leverages the data's inherent structure to create supervisory signals without relying on external labels. Due to this unique characteristic, it has been demonstrated that SSL does help in issues like data scarcity \citep{kraft2021overcoming,akrim2023self}, generalization capabilities \citep{bansal2020self,tendle2021study,huang2021towards,kim2021selfreg}, overfitting, or class imbalance \citep{liu2021imbalance,li2021imbalance,zhou2023novel}, which are prevalent in STR \citep{wan2020vocabulary,garcia2022out,yang2024class}. Then, SSL techniques can heavily alleviate such challenges by reducing out-of-vocabulary reliance \citep{zhang2022context}, addressing background noise or distortions in natural scenes \citep{zhuang2022text}.

As explained in Section \ref{sec:Reaching_a_Performance_Ceiling}, STR is currently reaching a performance ceiling due to recent developments and the simplicity of current benchmarking datasets. Works like that of \cite{jiang2023revisiting} have already highlighted this problem and proposed more challenging datasets. The Union14M-Benchmark dataset \citep{jiang2023revisiting}, used by the latest method recorded in this work (SSM), focuses on unexplored STR problems such as salient text, multi-word text, and incomplete text. It is particularly interesting to see how SSL can improve recognition under such circumstances. Moreover, using well-defined datasets would help alleviate the data issues mentioned above.

\subsubsection{Future research of SSL for HTR}
\label{sec:Future_research_of_SSL_for_HTR}
Following the STR trend, SSL methods developed for HTR have been evaluated for word-level recognition. However, the current state of HTR is line-level recognition \citep{voigtlaender2016handwriting,chowdhury2018efficient,yousef2020accurate,coquenet2022end,li2023trocr} and beyond. Therefore, it would be advisable to start evaluating how current methods perform at least at the line level, which is probably the most common setting in the HTR literature. Line-level recognition would introduce new challenges, enriching and complicating SSL application in this field. The variability in line lengths and aspect ratios, for instance, could make it difficult to use methods developed for STR, which typically rescale inputs to a fixed size. Additionally, semantic information would expand from word level (character set) to line level (word set). These challenges have yet to be explored, limiting the potential of current SSL methods in HTR.

Furthermore, HTR has recently transitioned to develop paragraph- or page-level recognition using end-to-end formulations \citep{singh2021full,kim2022ocr,dhiaf2023msdoctr,coquenet2023dan}. These typically use curriculum learning or synthetic data generation for model training. Therefore, this challenge could significantly benefit from SSL methods that enhance model learning and robustness by pre-training on real unlabeled data in a self-supervised manner.

While STR has seen the development of unlabeled real datasets for SSL, to our knowledge HTR has not yet explored this area. The current state of the art still relies on pre-training with synthetic data generation \citep{kang2020distilling,cascianelli2021learning,kang2022pay,pippi2023evaluating}. \review{However, throughout this work we have seen the superiority of real data over synthetic collections}. Therefore, efforts should also be made to create large datasets of unlabeled real data to enable the development of more robust HTR models using SSL.

\bmhead{Acknowledgements}
This work was partially funded by the Generalitat Valenciana through project CIGE/2023/216. The first author is supported by the University of Alicante through FPU Program (UAFPU22-19).

\section*{Declarations}
\begin{itemize}
\item \textbf{Competing interests:} The authors have no competing interests to declare that are relevant to the content of this article.
\item \textbf{Author contributions:} C.P. had the idea for the article, performed the literature search and data analysis, and drafted the work; J.J.V.-M. and J.C.-Z. provided guidance, discussed the approach, supervised the research, and critically revised the work.
\end{itemize}


\bibliography{sn-bibliography}

\begin{thebibliography}{207}
\providecommand{\natexlab}[1]{#1}
\providecommand{\url}[1]{{#1}}
\providecommand{\urlprefix}{URL }
\providecommand{\doi}[1]{\url{https://doi.org/#1}}
\providecommand{\eprint}[2][]{\url{#2}}
 \bibcommenthead

\bibitem[{Aberdam et~al(2021)Aberdam, Litman, Tsiper, Anschel, Slossberg,
  Mazor, Manmatha, and Perona}]{aberdam2021sequence}
Aberdam A, Litman R, Tsiper S, et~al (2021) Sequence-to-sequence contrastive
  learning for text recognition. In: Proceedings of the IEEE/CVF Conference on
  Computer Vision and Pattern Recognition, pp 15302--15312

\bibitem[{Aberdam et~al(2022)Aberdam, Ganz, Mazor, and
  Litman}]{aberdam2022multimodal}
Aberdam A, Ganz R, Mazor S, et~al (2022) Multimodal semi-supervised learning
  for text recognition. arXiv preprint arXiv:220503873

\bibitem[{Aggarwal et~al(2021)Aggarwal, Mittal, and
  Battineni}]{aggarwal2021generative}
Aggarwal A, Mittal M, Battineni G (2021) Generative adversarial network: An
  overview of theory and applications. International Journal of Information
  Management Data Insights 1(1):100004

\bibitem[{Agrawal et~al(2015)Agrawal, Carreira, and
  Malik}]{agrawal2015learning}
Agrawal P, Carreira J, Malik J (2015) Learning to see by moving. In:
  Proceedings of the IEEE international conference on computer vision, pp
  37--45

\bibitem[{Akrim et~al(2023)Akrim, Gogu, Vingerhoeds, and
  Sala{\"u}n}]{akrim2023self}
Akrim A, Gogu C, Vingerhoeds R, et~al (2023) Self-supervised learning for data
  scarcity in a fatigue damage prognostic problem. Engineering Applications of
  Artificial Intelligence 120:105837

\bibitem[{Albelwi(2022)}]{albelwi2022survey}
Albelwi S (2022) Survey on self-supervised learning: auxiliary pretext tasks
  and contrastive learning methods in imaging. Entropy 24(4):551

\bibitem[{Ali et~al(2023)Ali, Benjdira, Koubaa, El-Shafai, Khan, and
  Boulila}]{ali2023vision}
Ali AM, Benjdira B, Koubaa A, et~al (2023) Vision transformers in image
  restoration: A survey. Sensors 23(5):2385

\bibitem[{AlKendi et~al(2024)AlKendi, Gechter, Heyberger, and
  Guyeux}]{alkendi2024advancements}
AlKendi W, Gechter F, Heyberger L, et~al (2024) Advancements and challenges in
  handwritten text recognition: A comprehensive survey. Journal of Imaging
  10(1):18

\bibitem[{Ansari et~al(2022)Ansari, Kaur, Rakhra, Singh, and
  Singh}]{ansari2022handwritten}
Ansari A, Kaur B, Rakhra M, et~al (2022) Handwritten text recognition using
  deep learning algorithms. In: 2022 4th International Conference on Artificial
  Intelligence and Speech Technology (AIST), IEEE, pp 1--6

\bibitem[{Anwar et~al(2020)Anwar, Tahir, Li, Mian, Khan, and
  Muzaffar}]{anwar2020image}
Anwar S, Tahir M, Li C, et~al (2020) Image colorization: A survey and dataset.
  arXiv preprint arXiv:200810774

\bibitem[{Atienza(2021)}]{atienza2021data}
Atienza R (2021) Data augmentation for scene text recognition. In: Proceedings
  of the IEEE/CVF international conference on computer vision, pp 1561--1570

\bibitem[{Ba et~al(2016)Ba, Kiros, and Hinton}]{ba2016layer}
Ba JL, Kiros JR, Hinton GE (2016) Layer normalization. arXiv preprint
  arXiv:160706450

\bibitem[{Baek et~al(2019)Baek, Kim, Lee, Park, Han, Yun, Oh, and
  Lee}]{baek2019wrong}
Baek J, Kim G, Lee J, et~al (2019) What is wrong with scene text recognition
  model comparisons? dataset and model analysis. In: Proceedings of the
  IEEE/CVF international conference on computer vision, pp 4715--4723

\bibitem[{Baek et~al(2021)Baek, Matsui, and Aizawa}]{baek2021if}
Baek J, Matsui Y, Aizawa K (2021) What if we only use real datasets for scene
  text recognition? toward scene text recognition with fewer labels. In:
  Proceedings of the IEEE/CVF conference on computer vision and pattern
  recognition, pp 3113--3122

\bibitem[{Baevski et~al(2020)Baevski, Zhou, Mohamed, and
  Auli}]{baevski2020wav2vec}
Baevski A, Zhou Y, Mohamed A, et~al (2020) wav2vec 2.0: A framework for
  self-supervised learning of speech representations. Advances in neural
  information processing systems 33:12449--12460

\bibitem[{Baevski et~al(2023)Baevski, Babu, Hsu, and
  Auli}]{baevski2023efficient}
Baevski A, Babu A, Hsu WN, et~al (2023) Efficient self-supervised learning with
  contextualized target representations for vision, speech and language. In:
  International Conference on Machine Learning, PMLR, pp 1416--1429

\bibitem[{Bahdanau et~al(2014)Bahdanau, Cho, and Bengio}]{bahdanau2014neural}
Bahdanau D, Cho K, Bengio Y (2014) Neural machine translation by jointly
  learning to align and translate. arXiv preprint arXiv:14090473

\bibitem[{Balestriero et~al(2023)Balestriero, Ibrahim, Sobal, Morcos, Shekhar,
  Goldstein, Bordes, Bardes, Mialon, Tian et~al}]{balestriero2023cookbook}
Balestriero R, Ibrahim M, Sobal V, et~al (2023) A cookbook of self-supervised
  learning. arXiv preprint arXiv:230412210

\bibitem[{Bansal et~al(2020)Bansal, Kaplun, and Barak}]{bansal2020self}
Bansal Y, Kaplun G, Barak B (2020) For self-supervised learning, rationality
  implies generalization, provably. arXiv preprint arXiv:201008508

\bibitem[{Bao et~al(2021)Bao, Dong, Piao, and Wei}]{bao2021beit}
Bao H, Dong L, Piao S, et~al (2021) Beit: Bert pre-training of image
  transformers. arXiv preprint arXiv:210608254

\bibitem[{Bardes et~al(2021)Bardes, Ponce, and LeCun}]{bardes2021vicreg}
Bardes A, Ponce J, LeCun Y (2021) Vicreg: Variance-invariance-covariance
  regularization for self-supervised learning. arXiv preprint arXiv:210504906

\bibitem[{Bardes et~al(2022)Bardes, Ponce, and LeCun}]{bardes2022vicregl}
Bardes A, Ponce J, LeCun Y (2022) Vicregl: Self-supervised learning of local
  visual features. Advances in Neural Information Processing Systems
  35:8799--8810

\bibitem[{Baykal et~al(2022)Baykal, Ozcelik, and Unal}]{baykal2022exploring}
Baykal G, Ozcelik F, Unal G (2022) Exploring deshufflegans in self-supervised
  generative adversarial networks. Pattern Recognition 122:108244

\bibitem[{Ben-Shaul et~al(2023)Ben-Shaul, Shwartz-Ziv, Galanti, Dekel, and
  LeCun}]{ben2023reverse}
Ben-Shaul I, Shwartz-Ziv R, Galanti T, et~al (2023) Reverse engineering
  self-supervised learning. Advances in Neural Information Processing Systems
  36:58324--58345

\bibitem[{Bezerra et~al(2017)Bezerra, Zanchettin, Toselli, and
  Pirlo}]{BezerraZanchettinToseeliPirlo:Book:2017}
Bezerra BLD, Zanchettin C, Toselli AH, et~al (2017) Handwriting: recognition,
  development and analysis. Nova Science Publishers, Inc.

\bibitem[{Biten et~al(2019)Biten, Tito, Mafla, Gomez, Rusinol, Valveny,
  Jawahar, and Karatzas}]{biten2019scene}
Biten AF, Tito R, Mafla A, et~al (2019) Scene text visual question answering.
  In: Proceedings of the IEEE/CVF international conference on computer vision,
  pp 4291--4301

\bibitem[{Bordes et~al(2023)Bordes, Lavoie, Balestriero, Ballas, and
  Vincent}]{bordes2023surprisingly}
Bordes F, Lavoie S, Balestriero R, et~al (2023) A surprisingly simple technique
  to control the pretraining bias for better transfer: Expand or narrow your
  representation. arXiv preprint arXiv:230405369

\bibitem[{Bromley et~al(1993)Bromley, Guyon, LeCun, S{\"a}ckinger, and
  Shah}]{bromley1993signature}
Bromley J, Guyon I, LeCun Y, et~al (1993) Signature verification using a"
  siamese" time delay neural network. Advances in neural information processing
  systems 6

\bibitem[{Carlucci et~al(2019)Carlucci, D'Innocente, Bucci, Caputo, and
  Tommasi}]{carlucci2019domain}
Carlucci FM, D'Innocente A, Bucci S, et~al (2019) Domain generalization by
  solving jigsaw puzzles. In: Proceedings of the IEEE/CVF Conference on
  Computer Vision and Pattern Recognition, pp 2229--2238

\bibitem[{Caron et~al(2018)Caron, Bojanowski, Joulin, and
  Douze}]{caron2018deep}
Caron M, Bojanowski P, Joulin A, et~al (2018) Deep clustering for unsupervised
  learning of visual features. In: Proceedings of the European conference on
  computer vision (ECCV), pp 132--149

\bibitem[{Caron et~al(2020)Caron, Misra, Mairal, Goyal, Bojanowski, and
  Joulin}]{caron2020unsupervised}
Caron M, Misra I, Mairal J, et~al (2020) Unsupervised learning of visual
  features by contrasting cluster assignments. Advances in neural information
  processing systems 33:9912--9924

\bibitem[{Caron et~al(2021)Caron, Touvron, Misra, J{\'e}gou, Mairal,
  Bojanowski, and Joulin}]{caron2021emerging}
Caron M, Touvron H, Misra I, et~al (2021) Emerging properties in
  self-supervised vision transformers. In: Proceedings of the IEEE/CVF
  international conference on computer vision, pp 9650--9660

\bibitem[{Cascianelli et~al(2021)Cascianelli, Cornia, Baraldi, Piazzi, Schiuma,
  and Cucchiara}]{cascianelli2021learning}
Cascianelli S, Cornia M, Baraldi L, et~al (2021) Learning to read l’infinito:
  handwritten text recognition with synthetic training data. In: Computer
  Analysis of Images and Patterns: 19th International Conference, CAIP 2021,
  Virtual Event, September 28--30, 2021, Proceedings, Part II 19, Springer, pp
  340--350

\bibitem[{Chen et~al(2020{\natexlab{a}})Chen, Watanabe, Villalba, {\.Z}elasko,
  and Dehak}]{chen2020non}
Chen N, Watanabe S, Villalba J, et~al (2020{\natexlab{a}}) Non-autoregressive
  transformer for speech recognition. IEEE Signal Processing Letters
  28:121--125

\bibitem[{Chen and Guo(2023)}]{chen2023auto}
Chen S, Guo W (2023) Auto-encoders in deep learning—a review with new
  perspectives. Mathematics 11(8):1777

\bibitem[{Chen et~al(2019)Chen, Zhai, Ritter, Lucic, and
  Houlsby}]{chen2019self}
Chen T, Zhai X, Ritter M, et~al (2019) Self-supervised gans via auxiliary
  rotation loss. In: Proceedings of the IEEE/CVF conference on computer vision
  and pattern recognition, pp 12154--12163

\bibitem[{Chen et~al(2020{\natexlab{b}})Chen, Kornblith, Norouzi, and
  Hinton}]{chen2020simple}
Chen T, Kornblith S, Norouzi M, et~al (2020{\natexlab{b}}) A simple framework
  for contrastive learning of visual representations. In: International
  conference on machine learning, PMLR, pp 1597--1607

\bibitem[{Chen and He(2021)}]{chen2021exploring}
Chen X, He K (2021) Exploring simple siamese representation learning. In:
  Proceedings of the IEEE/CVF conference on computer vision and pattern
  recognition, pp 15750--15758

\bibitem[{Chen et~al(2020{\natexlab{c}})Chen, Fan, Girshick, and
  He}]{chen2020improved}
Chen X, Fan H, Girshick R, et~al (2020{\natexlab{c}}) Improved baselines with
  momentum contrastive learning. arXiv preprint arXiv:200304297

\bibitem[{Chen et~al(2021{\natexlab{a}})Chen, Jin, Zhu, Luo, and
  Wang}]{chen2021text}
Chen X, Jin L, Zhu Y, et~al (2021{\natexlab{a}}) Text recognition in the wild:
  A survey. ACM Computing Surveys (CSUR) 54(2):1--35

\bibitem[{Chen et~al(2021{\natexlab{b}})Chen, Xie, and He}]{chen2021empirical}
Chen X, Xie S, He K (2021{\natexlab{b}}) An empirical study of training
  self-supervised vision transformers. In: Proceedings of the IEEE/CVF
  international conference on computer vision, pp 9640--9649

\bibitem[{Chen et~al(2024)Chen, Ding, Wang, Xin, Mo, Wang, Han, Luo, Zeng, and
  Wang}]{chen2024context}
Chen X, Ding M, Wang X, et~al (2024) Context autoencoder for self-supervised
  representation learning. International Journal of Computer Vision
  132(1):208--223

\bibitem[{Chen et~al(2022)Chen, Bardes, Li, and LeCun}]{chen2022bag}
Chen Y, Bardes A, Li Z, et~al (2022) Bag of image patch embedding behind the
  success of self-supervised learning. arXiv preprint arXiv:220608954

\bibitem[{Cheng et~al(2015)Cheng, Yang, and Sheng}]{cheng2015deep}
Cheng Z, Yang Q, Sheng B (2015) Deep colorization. In: Proceedings of the IEEE
  international conference on computer vision, pp 415--423

\bibitem[{Cheng et~al(2017)Cheng, Bai, Xu, Zheng, Pu, and
  Zhou}]{cheng2017focusing}
Cheng Z, Bai F, Xu Y, et~al (2017) Focusing attention: Towards accurate text
  recognition in natural images. In: Proceedings of the IEEE international
  conference on computer vision, pp 5076--5084

\bibitem[{Chng et~al(2019)Chng, Liu, Sun, Ng, Luo, Ni, Fang, Zhang, Han, Ding
  et~al}]{chng2019icdar2019}
Chng CK, Liu Y, Sun Y, et~al (2019) Icdar2019 robust reading challenge on
  arbitrary-shaped text-rrc-art. In: 2019 International Conference on Document
  Analysis and Recognition (ICDAR), IEEE, pp 1571--1576

\bibitem[{Chowdhury and Vig(2018)}]{chowdhury2018efficient}
Chowdhury A, Vig L (2018) An efficient end-to-end neural model for handwritten
  text recognition. arXiv preprint arXiv:180707965

\bibitem[{Ciga et~al(2021)Ciga, Xu, and Martel}]{ciga2021resource}
Ciga O, Xu T, Martel AL (2021) Resource and data efficient self supervised
  learning. arXiv preprint arXiv:210901721

\bibitem[{Coquenet et~al(2022)Coquenet, Chatelain, and
  Paquet}]{coquenet2022end}
Coquenet D, Chatelain C, Paquet T (2022) End-to-end handwritten paragraph text
  recognition using a vertical attention network. IEEE Transactions on Pattern
  Analysis and Machine Intelligence 45(1):508--524

\bibitem[{Coquenet et~al(2023)Coquenet, Chatelain, and
  Paquet}]{coquenet2023dan}
Coquenet D, Chatelain C, Paquet T (2023) Dan: a segmentation-free document
  attention network for handwritten document recognition. IEEE transactions on
  pattern analysis and machine intelligence

\bibitem[{Creswell et~al(2018)Creswell, White, Dumoulin, Arulkumaran, Sengupta,
  and Bharath}]{creswell2018generative}
Creswell A, White T, Dumoulin V, et~al (2018) Generative adversarial networks:
  An overview. IEEE signal processing magazine 35(1):53--65

\bibitem[{Devlin et~al(2018)Devlin, Chang, Lee, and Toutanova}]{devlin2018bert}
Devlin J, Chang MW, Lee K, et~al (2018) Bert: Pre-training of deep
  bidirectional transformers for language understanding. arXiv preprint
  arXiv:181004805

\bibitem[{Dhiaf et~al(2023)Dhiaf, Rouhou, Kessentini, and
  Salem}]{dhiaf2023msdoctr}
Dhiaf M, Rouhou AC, Kessentini Y, et~al (2023) Msdoctr-lite: A lite transformer
  for full page multi-script handwriting recognition. Pattern Recognition
  Letters 169:28--34

\bibitem[{Diaz et~al(2021)Diaz, Qin, Ingle, Fujii, and
  Bissacco}]{diaz2021rethinking}
Diaz DH, Qin S, Ingle R, et~al (2021) Rethinking text line recognition models.
  arXiv preprint arXiv:210407787

\bibitem[{Doersch et~al(2015)Doersch, Gupta, and
  Efros}]{doersch2015unsupervised}
Doersch C, Gupta A, Efros AA (2015) Unsupervised visual representation learning
  by context prediction. In: Proceedings of the IEEE international conference
  on computer vision, pp 1422--1430

\bibitem[{Dong et~al(2021)Dong, Wang, and Abbas}]{dong2021survey}
Dong S, Wang P, Abbas K (2021) A survey on deep learning and its applications.
  Computer Science Review 40:100379

\bibitem[{Dosovitskiy et~al(2020)Dosovitskiy, Beyer, Kolesnikov, Weissenborn,
  Zhai, Unterthiner, Dehghani, Minderer, Heigold, Gelly
  et~al}]{dosovitskiy2020image}
Dosovitskiy A, Beyer L, Kolesnikov A, et~al (2020) An image is worth 16x16
  words: Transformers for image recognition at scale. arXiv preprint
  arXiv:201011929

\bibitem[{Elharrouss et~al(2020)Elharrouss, Almaadeed, Al-Maadeed, and
  Akbari}]{elharrouss2020image}
Elharrouss O, Almaadeed N, Al-Maadeed S, et~al (2020) Image inpainting: A
  review. Neural Processing Letters 51:2007--2028

\bibitem[{Feng et~al(2019)Feng, Xu, and Tao}]{feng2019self}
Feng Z, Xu C, Tao D (2019) Self-supervised representation learning by rotation
  feature decoupling. In: Proceedings of the IEEE/CVF Conference on Computer
  Vision and Pattern Recognition, pp 10364--10374

\bibitem[{Fujitake(2024)}]{fujitake2024dtrocr}
Fujitake M (2024) Dtrocr: Decoder-only transformer for optical character
  recognition. In: Proceedings of the IEEE/CVF winter conference on
  applications of computer vision, pp 8025--8035

\bibitem[{Gao et~al(2021)Gao, Chen, Wang, and Lu}]{gao2021semi}
Gao Y, Chen Y, Wang J, et~al (2021) Semi-supervised scene text recognition.
  IEEE Transactions on Image Processing 30:3005--3016

\bibitem[{Gao et~al(2024)Gao, Wang, Qu, Zhang, Wang, Xu, and Xie}]{gao2024self}
Gao Z, Wang Y, Qu Y, et~al (2024) Self-supervised pre-training with symmetric
  superimposition modeling for scene text recognition. arXiv preprint
  arXiv:240505841

\bibitem[{Garcia-Bordils et~al(2022)Garcia-Bordils, Mafla, Biten, Nuriel,
  Aberdam, Mazor, Litman, and Karatzas}]{garcia2022out}
Garcia-Bordils S, Mafla A, Biten AF, et~al (2022) Out-of-vocabulary challenge
  report. In: European Conference on Computer Vision, Springer, pp 359--375

\bibitem[{Garrido et~al(2022)Garrido, Chen, Bardes, Najman, and
  Lecun}]{garrido2022duality}
Garrido Q, Chen Y, Bardes A, et~al (2022) On the duality between contrastive
  and non-contrastive self-supervised learning. arXiv preprint arXiv:220602574

\bibitem[{Ghosh et~al(2022)Ghosh, Sen, Obaidullah, Santosh, Roy, and
  Pal}]{ghosh2022advances}
Ghosh T, Sen S, Obaidullah SM, et~al (2022) Advances in online handwritten
  recognition in the last decades. Computer Science Review 46:100515

\bibitem[{Gidaris et~al(2018)Gidaris, Singh, and
  Komodakis}]{gidaris2018unsupervised}
Gidaris S, Singh P, Komodakis N (2018) Unsupervised representation learning by
  predicting image rotations. arXiv preprint arXiv:180307728

\bibitem[{G{\'o}mez et~al(2018)G{\'o}mez, Mafla, Rusinol, and
  Karatzas}]{gomez2018single}
G{\'o}mez L, Mafla A, Rusinol M, et~al (2018) Single shot scene text retrieval.
  In: Proceedings of the European conference on computer vision (ECCV), pp
  700--715

\bibitem[{Goodfellow et~al(2014)Goodfellow, Pouget-Abadie, Mirza, Xu,
  Warde-Farley, Ozair, Courville, and Bengio}]{goodfellow2014generative}
Goodfellow I, Pouget-Abadie J, Mirza M, et~al (2014) Generative adversarial
  nets. Advances in neural information processing systems 27

\bibitem[{Graves et~al(2006)Graves, Fern{\'a}ndez, Gomez, and
  Schmidhuber}]{graves2006connectionist}
Graves A, Fern{\'a}ndez S, Gomez F, et~al (2006) Connectionist temporal
  classification: labelling unsegmented sequence data with recurrent neural
  networks. In: Proceedings of the 23rd international conference on Machine
  learning, pp 369--376

\bibitem[{Grill et~al(2020)Grill, Strub, Altch{\'e}, Tallec, Richemond,
  Buchatskaya, Doersch, Avila~Pires, Guo, Gheshlaghi~Azar
  et~al}]{grill2020bootstrap}
Grill JB, Strub F, Altch{\'e} F, et~al (2020) Bootstrap your own latent-a new
  approach to self-supervised learning. Advances in neural information
  processing systems 33:21271--21284

\bibitem[{Grosicki and El~Abed(2009)}]{grosicki2009icdar}
Grosicki E, El~Abed H (2009) Icdar 2009 handwriting recognition competition.
  In: 2009 10th International Conference on Document Analysis and Recognition,
  IEEE, pp 1398--1402

\bibitem[{Guan et~al(2023)Guan, Shen, Yang, Feng, Jiang, and
  Yang}]{guan2023self}
Guan T, Shen W, Yang X, et~al (2023) Self-supervised character-to-character
  distillation for text recognition. In: Proceedings of the IEEE/CVF
  International Conference on Computer Vision, pp 19473--19484

\bibitem[{Gui et~al(2021)Gui, Sun, Wen, Tao, and Ye}]{gui2021review}
Gui J, Sun Z, Wen Y, et~al (2021) A review on generative adversarial networks:
  Algorithms, theory, and applications. IEEE transactions on knowledge and data
  engineering 35(4):3313--3332

\bibitem[{Gupta et~al(2016)Gupta, Vedaldi, and Zisserman}]{gupta2016synthetic}
Gupta A, Vedaldi A, Zisserman A (2016) Synthetic data for text localisation in
  natural images. In: Proceedings of the IEEE conference on computer vision and
  pattern recognition, pp 2315--2324

\bibitem[{Gupta et~al(2022)Gupta, Ajanthan, Hengel, and
  Gould}]{gupta2022understanding}
Gupta K, Ajanthan T, Hengel Avd, et~al (2022) Understanding and improving the
  role of projection head in self-supervised learning. arXiv preprint
  arXiv:221211491

\bibitem[{Gupta and Jalal(2022)}]{gupta2022traditional}
Gupta N, Jalal AS (2022) Traditional to transfer learning progression on scene
  text detection and recognition: a survey. Artificial Intelligence Review
  55(4):3457--3502

\bibitem[{Gutmann and Hyv{\"a}rinen(2010)}]{gutmann2010noise}
Gutmann M, Hyv{\"a}rinen A (2010) Noise-contrastive estimation: A new
  estimation principle for unnormalized statistical models. In: Proceedings of
  the thirteenth international conference on artificial intelligence and
  statistics, JMLR Workshop and Conference Proceedings, pp 297--304

\bibitem[{Hadsell et~al(2006)Hadsell, Chopra, and
  LeCun}]{hadsell2006dimensionality}
Hadsell R, Chopra S, LeCun Y (2006) Dimensionality reduction by learning an
  invariant mapping. In: 2006 IEEE computer society conference on computer
  vision and pattern recognition (CVPR'06), IEEE, pp 1735--1742

\bibitem[{Harshvardhan et~al(2020)Harshvardhan, Gourisaria, Pandey, and
  Rautaray}]{harshvardhan2020comprehensive}
Harshvardhan G, Gourisaria MK, Pandey M, et~al (2020) A comprehensive survey
  and analysis of generative models in machine learning. Computer Science
  Review 38:100285

\bibitem[{He et~al(2016)He, Zhang, Ren, and Sun}]{he2016deep}
He K, Zhang X, Ren S, et~al (2016) Deep residual learning for image
  recognition. In: Proceedings of the IEEE conference on computer vision and
  pattern recognition, pp 770--778

\bibitem[{He et~al(2020)He, Fan, Wu, Xie, and Girshick}]{he2020momentum}
He K, Fan H, Wu Y, et~al (2020) Momentum contrast for unsupervised visual
  representation learning. In: Proceedings of the IEEE/CVF conference on
  computer vision and pattern recognition, pp 9729--9738

\bibitem[{He et~al(2022)He, Chen, Xie, Li, Doll{\'a}r, and
  Girshick}]{he2022masked}
He K, Chen X, Xie S, et~al (2022) Masked autoencoders are scalable vision
  learners. In: Proceedings of the IEEE/CVF conference on computer vision and
  pattern recognition, pp 16000--16009

\bibitem[{De~la Higuera and Oncina(2014)}]{de2014most}
De~la Higuera C, Oncina J (2014) The most probable string: an algorithmic
  study. Journal of Logic and Computation 24(2):311--330

\bibitem[{Hochreiter and Schmidhuber(1997)}]{hochreiter1997long}
Hochreiter S, Schmidhuber J (1997) Long short-term memory. Neural computation
  9(8):1735--1780

\bibitem[{Huang et~al(2021)Huang, Yi, Zhao, and Jiang}]{huang2021towards}
Huang W, Yi M, Zhao X, et~al (2021) Towards the generalization of contrastive
  self-supervised learning. arXiv preprint arXiv:211100743

\bibitem[{Hubenthal and Kumar(2023)}]{hubenthal2023image}
Hubenthal M, Kumar S (2023) Image-text pre-training for logo recognition. In:
  Proceedings of the IEEE/CVF Winter Conference on Applications of Computer
  Vision, pp 1145--1154

\bibitem[{Iwana et~al(2016)Iwana, Rizvi, Ahmed, Dengel, and
  Uchida}]{iwana2016judging}
Iwana BK, Rizvi STR, Ahmed S, et~al (2016) Judging a book by its cover. arXiv
  preprint arXiv:161009204

\bibitem[{Jaderberg et~al(2014)Jaderberg, Simonyan, Vedaldi, and
  Zisserman}]{jaderberg2014synthetic}
Jaderberg M, Simonyan K, Vedaldi A, et~al (2014) Synthetic data and artificial
  neural networks for natural scene text recognition. arXiv preprint
  arXiv:14062227

\bibitem[{Jaiswal et~al(2020)Jaiswal, Babu, Zadeh, Banerjee, and
  Makedon}]{jaiswal2020survey}
Jaiswal A, Babu AR, Zadeh MZ, et~al (2020) A survey on contrastive
  self-supervised learning. Technologies 9(1):2

\bibitem[{Jiang et~al(2023{\natexlab{a}})Jiang, Wang, Peng, Liu, and
  Jin}]{jiang2023revisiting}
Jiang Q, Wang J, Peng D, et~al (2023{\natexlab{a}}) Revisiting scene text
  recognition: A data perspective. In: Proceedings of the IEEE/CVF
  international conference on computer vision, pp 20543--20554

\bibitem[{Jiang et~al(2022)Jiang, Zhang, Du, Zhang, and Wu}]{jiang2022scene}
Jiang X, Zhang J, Du J, et~al (2022) Scene text recognition with
  self-supervised contrastive predictive coding. In: 2022 26th International
  Conference on Pattern Recognition (ICPR), IEEE, pp 1514--1521

\bibitem[{Jiang et~al(2023{\natexlab{b}})Jiang, Du, Hu, Xue, Ma, Wu, and
  Zhang}]{jiang2023group}
Jiang X, Du J, Hu P, et~al (2023{\natexlab{b}}) Group, contrast and recognize:
  A self-supervised method for chinese character recognition. In: International
  Conference on Document Analysis and Recognition, Springer, pp 411--427

\bibitem[{Jing and Tian(2020)}]{jing2020self}
Jing L, Tian Y (2020) Self-supervised visual feature learning with deep neural
  networks: A survey. IEEE transactions on pattern analysis and machine
  intelligence 43(11):4037--4058

\bibitem[{Kang et~al(2020{\natexlab{a}})Kang, Riba, Rusinol, Forn{\'e}s, and
  Villegas}]{kang2020distilling}
Kang L, Riba P, Rusinol M, et~al (2020{\natexlab{a}}) Distilling content from
  style for handwritten word recognition. In: 2020 17th International
  Conference on Frontiers in Handwriting Recognition (ICFHR), IEEE, pp 139--144

\bibitem[{Kang et~al(2020{\natexlab{b}})Kang, Rusinol, Forn{\'e}s, Riba, and
  Villegas}]{kang2020unsupervised}
Kang L, Rusinol M, Forn{\'e}s A, et~al (2020{\natexlab{b}}) Unsupervised writer
  adaptation for synthetic-to-real handwritten word recognition. In:
  Proceedings of the IEEE/CVF Winter Conference on Applications of Computer
  Vision, pp 3502--3511

\bibitem[{Kang et~al(2022)Kang, Riba, Rusi{\~n}ol, Forn{\'e}s, and
  Villegas}]{kang2022pay}
Kang L, Riba P, Rusi{\~n}ol M, et~al (2022) Pay attention to what you read:
  non-recurrent handwritten text-line recognition. Pattern Recognition
  129:108766

\bibitem[{Karatzas et~al(2013)Karatzas, Shafait, Uchida, Iwamura, i~Bigorda,
  Mestre, Mas, Mota, Almazan, and De~Las~Heras}]{karatzas2013icdar}
Karatzas D, Shafait F, Uchida S, et~al (2013) Icdar 2013 robust reading
  competition. In: 2013 12th international conference on document analysis and
  recognition, IEEE, pp 1484--1493

\bibitem[{Karatzas et~al(2015)Karatzas, Gomez-Bigorda, Nicolaou, Ghosh,
  Bagdanov, Iwamura, Matas, Neumann, Chandrasekhar, Lu
  et~al}]{karatzas2015icdar}
Karatzas D, Gomez-Bigorda L, Nicolaou A, et~al (2015) Icdar 2015 competition on
  robust reading. In: 2015 13th international conference on document analysis
  and recognition (ICDAR), IEEE, pp 1156--1160

\bibitem[{Khan et~al(2021)Khan, Sarkar, and Mollah}]{khan2021deep}
Khan T, Sarkar R, Mollah AF (2021) Deep learning approaches to scene text
  detection: a comprehensive review. Artificial Intelligence Review
  54:3239--3298

\bibitem[{Kim et~al(2021)Kim, Yoo, Park, Kim, and Lee}]{kim2021selfreg}
Kim D, Yoo Y, Park S, et~al (2021) Selfreg: Self-supervised contrastive
  regularization for domain generalization. In: Proceedings of the IEEE/CVF
  International Conference on Computer Vision, pp 9619--9628

\bibitem[{Kim et~al(2022)Kim, Hong, Yim, Nam, Park, Yim, Hwang, Yun, Han, and
  Park}]{kim2022ocr}
Kim G, Hong T, Yim M, et~al (2022) Ocr-free document understanding transformer.
  In: European Conference on Computer Vision, Springer, pp 498--517

\bibitem[{Kleber et~al(2013)Kleber, Fiel, Diem, and Sablatnig}]{kleber2013cvl}
Kleber F, Fiel S, Diem M, et~al (2013) Cvl-database: An off-line database for
  writer retrieval, writer identification and word spotting. In: 2013 12th
  international conference on document analysis and recognition, IEEE, pp
  560--564

\bibitem[{Kraft et~al(2021)Kraft, Pieczy{\'n}ski, and
  Siemionow}]{kraft2021overcoming}
Kraft M, Pieczy{\'n}ski D, Siemionow KK (2021) Overcoming data scarcity for
  coronary vessel segmentation through self-supervised pre-training. In: Neural
  Information Processing: 28th International Conference, ICONIP 2021, Sanur,
  Bali, Indonesia, December 8--12, 2021, Proceedings, Part III 28, Springer, pp
  369--378

\bibitem[{Krizhevsky et~al(2012)Krizhevsky, Sutskever, and
  Hinton}]{krizhevsky2012imagenet}
Krizhevsky A, Sutskever I, Hinton GE (2012) Imagenet classification with deep
  convolutional neural networks. Advances in neural information processing
  systems 25

\bibitem[{Larsson et~al(2016)Larsson, Maire, and
  Shakhnarovich}]{larsson2016learning}
Larsson G, Maire M, Shakhnarovich G (2016) Learning representations for
  automatic colorization. In: Computer Vision--ECCV 2016: 14th European
  Conference, Amsterdam, The Netherlands, October 11--14, 2016, Proceedings,
  Part IV 14, Springer, pp 577--593

\bibitem[{Larsson et~al(2017)Larsson, Maire, and
  Shakhnarovich}]{larsson2017colorization}
Larsson G, Maire M, Shakhnarovich G (2017) Colorization as a proxy task for
  visual understanding. In: Proceedings of the IEEE conference on computer
  vision and pattern recognition, pp 6874--6883

\bibitem[{LeCun et~al(1989)LeCun, Boser, Denker, Henderson, Howard, Hubbard,
  and Jackel}]{lecun1989backpropagation}
LeCun Y, Boser B, Denker JS, et~al (1989) Backpropagation applied to
  handwritten zip code recognition. Neural computation 1(4):541--551

\bibitem[{Lee et~al(2020)Lee, Park, Baek, Oh, Kim, and
  Lee}]{lee2020recognizing}
Lee J, Park S, Baek J, et~al (2020) On recognizing texts of arbitrary shapes
  with 2d self-attention. In: Proceedings of the IEEE/CVF Conference on
  Computer Vision and Pattern Recognition Workshops, pp 546--547

\bibitem[{Li et~al(2021{\natexlab{a}})Li, Yang, Zhang, Gao, Xiao, Dai, Yuan,
  and Gao}]{li2021efficient}
Li C, Yang J, Zhang P, et~al (2021{\natexlab{a}}) Efficient self-supervised
  vision transformers for representation learning. arXiv preprint
  arXiv:210609785

\bibitem[{Li et~al(2022)Li, Feh{\'e}rv{\'a}ri, Zhao, Macedo, and
  Appalaraju}]{li2022seetek}
Li C, Feh{\'e}rv{\'a}ri I, Zhao X, et~al (2022) Seetek: Very large-scale
  open-set logo recognition with text-aware metric learning. In: Proceedings of
  the IEEE/CVF Winter Conference on Applications of Computer Vision, pp
  2544--2553

\bibitem[{Li et~al(2021{\natexlab{b}})Li, Xue, Chaitanya, Luo, Ezhov, Wiestler,
  Zhang, and Menze}]{li2021imbalance}
Li H, Xue FF, Chaitanya K, et~al (2021{\natexlab{b}}) Imbalance-aware
  self-supervised learning for 3d radiomic representations. In: Medical Image
  Computing and Computer Assisted Intervention--MICCAI 2021: 24th International
  Conference, Strasbourg, France, September 27--October 1, 2021, Proceedings,
  Part II 24, Springer, pp 36--46

\bibitem[{Li et~al(2023)Li, Lv, Chen, Cui, Lu, Florencio, Zhang, Li, and
  Wei}]{li2023trocr}
Li M, Lv T, Chen J, et~al (2023) Trocr: Transformer-based optical character
  recognition with pre-trained models. In: Proceedings of the AAAI Conference
  on Artificial Intelligence, pp 13094--13102

\bibitem[{Liang et~al(2021)Liang, Hu, Wang, He, and Feng}]{liang2021source}
Liang J, Hu D, Wang Y, et~al (2021) Source data-absent unsupervised domain
  adaptation through hypothesis transfer and labeling transfer. IEEE
  Transactions on Pattern Analysis and Machine Intelligence 44(11):8602--8617

\bibitem[{Liu et~al(2021{\natexlab{a}})Liu, HaoChen, Gaidon, and
  Ma}]{liu2021imbalance}
Liu H, HaoChen JZ, Gaidon A, et~al (2021{\natexlab{a}}) Self-supervised
  learning is more robust to dataset imbalance. arXiv preprint arXiv:211005025

\bibitem[{Liu et~al(2022)Liu, Wang, Bao, Xue, Kang, Jiang, Liu, and
  Ren}]{liu2022perceiving}
Liu H, Wang B, Bao Z, et~al (2022) Perceiving stroke-semantic context:
  Hierarchical contrastive learning for robust scene text recognition. In:
  Proceedings of the AAAI Conference on Artificial Intelligence, pp 1702--1710

\bibitem[{Liu et~al(2020)Liu, Zhong, Yuan, Su, and Du}]{liu2020semitext}
Liu J, Zhong Q, Yuan Y, et~al (2020) Semitext: Scene text detection with
  semi-supervised learning. Neurocomputing 407:343--353

\bibitem[{Liu et~al(2021{\natexlab{b}})Liu, Zhang, Hou, Mian, Wang, Zhang, and
  Tang}]{liu2021self}
Liu X, Zhang F, Hou Z, et~al (2021{\natexlab{b}}) Self-supervised learning:
  Generative or contrastive. IEEE transactions on knowledge and data
  engineering 35(1):857--876

\bibitem[{Liu et~al(2019)Liu, Ott, Goyal, Du, Joshi, Chen, Levy, Lewis,
  Zettlemoyer, and Stoyanov}]{liu2019roberta}
Liu Y, Ott M, Goyal N, et~al (2019) Roberta: A robustly optimized bert
  pretraining approach. arXiv preprint arXiv:190711692

\bibitem[{Long et~al(2021)Long, He, and Yao}]{long2021scene}
Long S, He X, Yao C (2021) Scene text detection and recognition: The deep
  learning era. International Journal of Computer Vision 129(1):161--184

\bibitem[{Lucas et~al(2005)Lucas, Panaretos, Sosa, Tang, Wong, Young, Ashida,
  Nagai, Okamoto, Yamamoto et~al}]{lucas2005icdar}
Lucas SM, Panaretos A, Sosa L, et~al (2005) Icdar 2003 robust reading
  competitions: entries, results, and future directions. International Journal
  of Document Analysis and Recognition (IJDAR) 7:105--122

\bibitem[{Luo et~al(2020)Luo, Zhu, Jin, and Wang}]{luo2020learn}
Luo C, Zhu Y, Jin L, et~al (2020) Learn to augment: Joint data augmentation and
  network optimization for text recognition. In: Proceedings of the IEEE/CVF
  Conference on Computer Vision and Pattern Recognition, pp 13746--13755

\bibitem[{Luo et~al(2022)Luo, Jin, and Chen}]{luo2022siman}
Luo C, Jin L, Chen J (2022) Siman: Exploring self-supervised representation
  learning of scene text via similarity-aware normalization. In: Proceedings of
  the IEEE/CVF Conference on Computer Vision and Pattern Recognition, pp
  1039--1048

\bibitem[{Lyu et~al(2023)Lyu, Zhang, Liu, Qiao, Xu, Wu, Yao, Han, Ding, and
  Wang}]{lyu2023maskocr}
Lyu P, Zhang C, Liu S, et~al (2023) Maskocr: Text recognition with masked
  encoder-decoder pretraining. arXiv preprint arXiv:220600311

\bibitem[{Marti and Bunke(2002)}]{marti2002iam}
Marti UV, Bunke H (2002) The iam-database: an english sentence database for
  offline handwriting recognition. International journal on document analysis
  and recognition 5:39--46

\bibitem[{Minaee et~al(2021)Minaee, Boykov, Porikli, Plaza, Kehtarnavaz, and
  Terzopoulos}]{minaee2021image}
Minaee S, Boykov Y, Porikli F, et~al (2021) Image segmentation using deep
  learning: A survey. IEEE transactions on pattern analysis and machine
  intelligence 44(7):3523--3542

\bibitem[{Mishra et~al(2012)Mishra, Alahari, and Jawahar}]{mishra2012scene}
Mishra A, Alahari K, Jawahar C (2012) Scene text recognition using higher order
  language priors. In: BMVC-British machine vision conference, BMVA

\bibitem[{Misra and Maaten(2020)}]{misra2020self}
Misra I, Maaten Lvd (2020) Self-supervised learning of pretext-invariant
  representations. In: Proceedings of the IEEE/CVF conference on computer
  vision and pattern recognition, pp 6707--6717

\bibitem[{Naiemi et~al(2022)Naiemi, Ghods, and Khalesi}]{naiemi2022scene}
Naiemi F, Ghods V, Khalesi H (2022) Scene text detection and recognition: a
  survey. Multimedia Tools and Applications 81(14):20255--20290

\bibitem[{Narang et~al(2020)Narang, Jindal, and Kumar}]{narang2020ancient}
Narang SR, Jindal MK, Kumar M (2020) Ancient text recognition: a review.
  Artificial Intelligence Review 53(8):5517--5558

\bibitem[{Nayef et~al(2019)Nayef, Patel, Busta, Chowdhury, Karatzas, Khlif,
  Matas, Pal, Burie, Liu et~al}]{nayef2019icdar2019}
Nayef N, Patel Y, Busta M, et~al (2019) Icdar2019 robust reading challenge on
  multi-lingual scene text detection and recognition—rrc-mlt-2019. In: 2019
  International conference on document analysis and recognition (ICDAR), IEEE,
  pp 1582--1587

\bibitem[{Nikitha et~al(2020)Nikitha, Geetha, and
  JayaLakshmi}]{nikitha2020handwritten}
Nikitha A, Geetha J, JayaLakshmi D (2020) Handwritten text recognition using
  deep learning. In: 2020 International Conference on Recent Trends on
  Electronics, Information, Communication \& Technology (RTEICT), IEEE, pp
  388--392

\bibitem[{Nikolaidou et~al(2022)Nikolaidou, Seuret, Mokayed, and
  Liwicki}]{nikolaidou2022survey}
Nikolaidou K, Seuret M, Mokayed H, et~al (2022) A survey of historical document
  image datasets. International Journal on Document Analysis and Recognition
  (IJDAR) 25(4):305--338

\bibitem[{Noroozi and Favaro(2016)}]{noroozi2016unsupervised}
Noroozi M, Favaro P (2016) Unsupervised learning of visual representations by
  solving jigsaw puzzles. In: European conference on computer vision, Springer,
  pp 69--84

\bibitem[{Novotny et~al(2018)Novotny, Albanie, Larlus, and
  Vedaldi}]{novotny2018self}
Novotny D, Albanie S, Larlus D, et~al (2018) Self-supervised learning of
  geometrically stable features through probabilistic introspection. In:
  Proceedings of the IEEE Conference on Computer Vision and Pattern
  Recognition, pp 3637--3645

\bibitem[{Oord et~al(2018)Oord, Li, and Vinyals}]{oord2018representation}
Oord Avd, Li Y, Vinyals O (2018) Representation learning with contrastive
  predictive coding. arXiv preprint arXiv:180703748

\bibitem[{Ozbulak et~al(2023)Ozbulak, Lee, Boga, Anzaku, Park, Van~Messem,
  De~Neve, and Vankerschaver}]{ozbulak2023know}
Ozbulak U, Lee HJ, Boga B, et~al (2023) Know your self-supervised learning: A
  survey on image-based generative and discriminative training. arXiv preprint
  arXiv:230513689

\bibitem[{Pajot et~al(2019)Pajot, de~Bezenac, and
  Gallinari}]{pajot2019unsupervised}
Pajot A, de~Bezenac E, Gallinari P (2019) Unsupervised adversarial image
  inpainting. arXiv preprint arXiv:191212164

\bibitem[{Pang et~al(2020)Pang, Yang, Hospedales, Xiang, and
  Song}]{pang2020solving}
Pang K, Yang Y, Hospedales TM, et~al (2020) Solving mixed-modal jigsaw puzzle
  for fine-grained sketch-based image retrieval. In: Proceedings of the
  IEEE/CVF conference on computer vision and pattern recognition, pp
  10347--10355

\bibitem[{Pathak et~al(2016)Pathak, Krahenbuhl, Donahue, Darrell, and
  Efros}]{pathak2016context}
Pathak D, Krahenbuhl P, Donahue J, et~al (2016) Context encoders: Feature
  learning by inpainting. In: Proceedings of the IEEE conference on computer
  vision and pattern recognition, pp 2536--2544

\bibitem[{Penarrubia et~al(2024)Penarrubia, Garrido-Munoz, Valero-Mas, and
  Calvo-Zaragoza}]{penarrubia2024spatial}
Penarrubia C, Garrido-Munoz C, Valero-Mas JJ, et~al (2024) Spatial
  context-based self-supervised learning for handwritten text recognition.
  arXiv preprint arXiv:240411585

\bibitem[{Phan et~al(2013)Phan, Shivakumara, Tian, and
  Tan}]{phan2013recognizing}
Phan TQ, Shivakumara P, Tian S, et~al (2013) Recognizing text with perspective
  distortion in natural scenes. In: Proceedings of the IEEE international
  conference on computer vision, pp 569--576

\bibitem[{Pippi et~al(2023)Pippi, Cascianelli, Baraldi, and
  Cucchiara}]{pippi2023evaluating}
Pippi V, Cascianelli S, Baraldi L, et~al (2023) Evaluating synthetic
  pre-training for handwriting processing tasks. Pattern Recognition Letters
  172:44--50

\bibitem[{Puigcerver(2017)}]{puigcerver2017multidimensional}
Puigcerver J (2017) Are multidimensional recurrent layers really necessary for
  handwritten text recognition? In: 2017 14th IAPR international conference on
  document analysis and recognition (ICDAR), IEEE, pp 67--72

\bibitem[{Qian et~al(2022)Qian, Xu, Hu, Li, and Jin}]{qian2022unsupervised}
Qian Q, Xu Y, Hu J, et~al (2022) Unsupervised visual representation learning by
  online constrained k-means. In: Proceedings of the IEEE/CVF Conference on
  Computer Vision and Pattern Recognition, pp 16640--16649

\bibitem[{Qiao et~al(2023)Qiao, Ji, Yuan, and Bai}]{qiao2023decoupling}
Qiao Z, Ji Z, Yuan Y, et~al (2023) Decoupling visual-semantic features learning
  with dual masked autoencoder for self-supervised scene text recognition. In:
  International Conference on Document Analysis and Recognition, Springer, pp
  261--279

\bibitem[{Qiu et~al(2020)Qiu, Sun, Xu, Shao, Dai, and Huang}]{qiu2020pre}
Qiu X, Sun T, Xu Y, et~al (2020) Pre-trained models for natural language
  processing: A survey. Science China Technological Sciences 63(10):1872--1897

\bibitem[{Radford et~al(2015)Radford, Metz, and
  Chintala}]{radford2015unsupervised}
Radford A, Metz L, Chintala S (2015) Unsupervised representation learning with
  deep convolutional generative adversarial networks. arXiv preprint
  arXiv:151106434

\bibitem[{Ramesh et~al(2021)Ramesh, Pavlov, Goh, Gray, Voss, Radford, Chen, and
  Sutskever}]{ramesh2021zero}
Ramesh A, Pavlov M, Goh G, et~al (2021) Zero-shot text-to-image generation. In:
  International conference on machine learning, Pmlr, pp 8821--8831

\bibitem[{Risnumawan et~al(2014)Risnumawan, Shivakumara, Chan, and
  Tan}]{risnumawan2014robust}
Risnumawan A, Shivakumara P, Chan CS, et~al (2014) A robust arbitrary text
  detection system for natural scene images. Expert Systems with Applications
  41(18):8027--8048

\bibitem[{Rolfe(2016)}]{rolfe2016discrete}
Rolfe JT (2016) Discrete variational autoencoders. arXiv preprint
  arXiv:160902200

\bibitem[{Sharma et~al(2020)Sharma, Kaushik, and Gondhi}]{sharma2020character}
Sharma R, Kaushik B, Gondhi N (2020) Character recognition using machine
  learning and deep learning-a survey. In: 2020 International Conference on
  Emerging Smart Computing and Informatics (ESCI), IEEE, pp 341--345

\bibitem[{Sherstinsky(2020)}]{sherstinsky2020fundamentals}
Sherstinsky A (2020) Fundamentals of recurrent neural network (rnn) and long
  short-term memory (lstm) network. Physica D: Nonlinear Phenomena 404:132306

\bibitem[{Shi et~al(2016{\natexlab{a}})Shi, Bai, and Yao}]{shi2016end}
Shi B, Bai X, Yao C (2016{\natexlab{a}}) An end-to-end trainable neural network
  for image-based sequence recognition and its application to scene text
  recognition. IEEE transactions on pattern analysis and machine intelligence
  39(11):2298--2304

\bibitem[{Shi et~al(2016{\natexlab{b}})Shi, Wang, Lyu, Yao, and
  Bai}]{shi2016robust}
Shi B, Wang X, Lyu P, et~al (2016{\natexlab{b}}) Robust scene text recognition
  with automatic rectification. In: Proceedings of the IEEE conference on
  computer vision and pattern recognition, pp 4168--4176

\bibitem[{Shi et~al(2017)Shi, Yao, Liao, Yang, Xu, Cui, Belongie, Lu, and
  Bai}]{shi2017icdar2017}
Shi B, Yao C, Liao M, et~al (2017) Icdar2017 competition on reading chinese
  text in the wild (rctw-17). In: 2017 14th iapr international conference on
  document analysis and recognition (ICDAR), IEEE, pp 1429--1434

\bibitem[{Singh et~al(2019)Singh, Natarajan, Shah, Jiang, Chen, Batra, Parikh,
  and Rohrbach}]{singh2019towards}
Singh A, Natarajan V, Shah M, et~al (2019) Towards vqa models that can read.
  In: Proceedings of the IEEE/CVF conference on computer vision and pattern
  recognition, pp 8317--8326

\bibitem[{Singh et~al(2021)Singh, Pang, Toh, Huang, Galuba, and
  Hassner}]{singh2021textocr}
Singh A, Pang G, Toh M, et~al (2021) Textocr: Towards large-scale end-to-end
  reasoning for arbitrary-shaped scene text. In: Proceedings of the IEEE/CVF
  conference on computer vision and pattern recognition, pp 8802--8812

\bibitem[{Singh and Karayev(2021)}]{singh2021full}
Singh SS, Karayev S (2021) Full page handwriting recognition via image to
  sequence extraction. In: Document Analysis and Recognition--ICDAR 2021: 16th
  International Conference, Lausanne, Switzerland, September 5--10, 2021,
  Proceedings, Part III 16, Springer, pp 55--69

\bibitem[{Smys et~al(2020)Smys, Chen, and Shakya}]{smys2020survey}
Smys S, Chen JIZ, Shakya S (2020) Survey on neural network architectures with
  deep learning. Journal of Soft Computing Paradigm (JSCP) 2(03):186--194

\bibitem[{Souibgui et~al(2023)Souibgui, Biswas, Mafla, Biten, Forn{\'e}s,
  Kessentini, Llad{\'o}s, Gomez, and Karatzas}]{souibgui2023text}
Souibgui MA, Biswas S, Mafla A, et~al (2023) Text-diae: A self-supervised
  degradation invariant autoencoder for text recognition and document
  enhancement. In: proceedings of the AAAI conference on artificial
  intelligence, pp 2330--2338

\bibitem[{Sun et~al(2019)Sun, Ni, Chng, Liu, Luo, Ng, Han, Ding, Liu, Karatzas
  et~al}]{sun2019icdar}
Sun Y, Ni Z, Chng CK, et~al (2019) Icdar 2019 competition on large-scale street
  view text with partial labeling-rrc-lsvt. In: 2019 International Conference
  on Document Analysis and Recognition (ICDAR), IEEE, pp 1557--1562

\bibitem[{Tao et~al(2023)Tao, Zhu, Su, Huang, Li, Zhou, Qiao, Wang, and
  Dai}]{tao2023siamese}
Tao C, Zhu X, Su W, et~al (2023) Siamese image modeling for self-supervised
  vision representation learning. In: Proceedings of the IEEE/CVF Conference on
  Computer Vision and Pattern Recognition, pp 2132--2141

\bibitem[{Tendle and Hasan(2021)}]{tendle2021study}
Tendle A, Hasan MR (2021) A study of the generalizability of self-supervised
  representations. Machine Learning with Applications 6:100124

\bibitem[{Tong et~al(2023)Tong, Chen, Ma, and Lecun}]{tong2023emp}
Tong S, Chen Y, Ma Y, et~al (2023) Emp-ssl: Towards self-supervised learning in
  one training epoch. arXiv preprint arXiv:230403977

\bibitem[{Touvron et~al(2021)Touvron, Cord, Douze, Massa, Sablayrolles, and
  J{\'e}gou}]{touvron2021training}
Touvron H, Cord M, Douze M, et~al (2021) Training data-efficient image
  transformers \& distillation through attention. In: International conference
  on machine learning, PMLR, pp 10347--10357

\bibitem[{Valero-Mas et~al(2024)Valero-Mas, Gallego, and
  Rico-Juan}]{valero2024overview}
Valero-Mas JJ, Gallego AJ, Rico-Juan JR (2024) An overview of ensemble and
  feature learning in few-shot image classification using siamese networks.
  Multimedia Tools and Applications 83(7):19929--19952

\bibitem[{Vaswani et~al(2017)Vaswani, Shazeer, Parmar, Uszkoreit, Jones, Gomez,
  Kaiser, and Polosukhin}]{vaswani2017attention}
Vaswani A, Shazeer N, Parmar N, et~al (2017) Attention is all you need.
  Advances in neural information processing systems 30

\bibitem[{Veit et~al(2016)Veit, Matera, Neumann, Matas, and
  Belongie}]{veit2016coco}
Veit A, Matera T, Neumann L, et~al (2016) Coco-text: Dataset and benchmark for
  text detection and recognition in natural images. arXiv preprint
  arXiv:160107140

\bibitem[{Vincent et~al(2008)Vincent, Larochelle, Bengio, and
  Manzagol}]{vincent2008extracting}
Vincent P, Larochelle H, Bengio Y, et~al (2008) Extracting and composing robust
  features with denoising autoencoders. In: Proceedings of the 25th
  international conference on Machine learning, pp 1096--1103

\bibitem[{Voigtlaender et~al(2016)Voigtlaender, Doetsch, and
  Ney}]{voigtlaender2016handwriting}
Voigtlaender P, Doetsch P, Ney H (2016) Handwriting recognition with large
  multidimensional long short-term memory recurrent neural networks. In: 2016
  15th international conference on frontiers in handwriting recognition
  (ICFHR), IEEE, pp 228--233

\bibitem[{Wan et~al(2020)Wan, Zhang, Zhang, Luo, and Yao}]{wan2020vocabulary}
Wan Z, Zhang J, Zhang L, et~al (2020) On vocabulary reliance in scene text
  recognition. In: Proceedings of the IEEE/CVF Conference on Computer Vision
  and Pattern Recognition, pp 11425--11434

\bibitem[{Wang et~al(2021)Wang, Wang, Wang, Torr, and Lin}]{wang2021solving}
Wang G, Wang K, Wang G, et~al (2021) Solving inefficiency of self-supervised
  representation learning. In: Proceedings of the IEEE/CVF International
  Conference on Computer Vision, pp 9505--9515

\bibitem[{Wang et~al(2011)Wang, Babenko, and Belongie}]{wang2011end}
Wang K, Babenko B, Belongie S (2011) End-to-end scene text recognition. In:
  2011 International conference on computer vision, IEEE, pp 1457--1464

\bibitem[{Wang et~al(2020)Wang, Wei, Dong, Bao, Yang, and
  Zhou}]{wang2020minilm}
Wang W, Wei F, Dong L, et~al (2020) Minilm: Deep self-attention distillation
  for task-agnostic compression of pre-trained transformers. Advances in neural
  information processing systems 33:5776--5788

\bibitem[{Wang and Gupta(2015)}]{wang2015unsupervised}
Wang X, Gupta A (2015) Unsupervised learning of visual representations using
  videos. In: Proceedings of the IEEE international conference on computer
  vision, pp 2794--2802

\bibitem[{Wang et~al(2023)Wang, He, Wang, Wang, Zou, and Wu}]{wang2023survey}
Wang XF, He ZH, Wang K, et~al (2023) A survey of text detection and recognition
  algorithms based on deep learning technology. Neurocomputing 556:126702

\bibitem[{Wei et~al(2020)Wei, Wang, Shen, and Yuille}]{wei2020co2}
Wei C, Wang H, Shen W, et~al (2020) Co2: Consistent contrast for unsupervised
  visual representation learning. arXiv preprint arXiv:201002217

\bibitem[{Xiang et~al(2023)Xiang, Zou, Nawaz, Huang, Zhang, and
  Yu}]{xiang2023deep}
Xiang H, Zou Q, Nawaz MA, et~al (2023) Deep learning for image inpainting: A
  survey. Pattern Recognition 134:109046

\bibitem[{Xie et~al(2022)Xie, Zhang, Cao, Lin, Bao, Yao, Dai, and
  Hu}]{xie2022simmim}
Xie Z, Zhang Z, Cao Y, et~al (2022) Simmim: A simple framework for masked image
  modeling. In: Proceedings of the IEEE/CVF conference on computer vision and
  pattern recognition, pp 9653--9663

\bibitem[{Yamaguchi et~al(2021)Yamaguchi, Kanai, Shioda, and
  Takeda}]{yamaguchi2021image}
Yamaguchi S, Kanai S, Shioda T, et~al (2021) Image enhanced rotation prediction
  for self-supervised learning. In: 2021 IEEE International Conference on Image
  Processing (ICIP), IEEE, pp 489--493

\bibitem[{Yang et~al(2022{\natexlab{a}})Yang, Liao, Lu, Wang, Zhu, Luo, Tian,
  and Bai}]{yang2022reading}
Yang M, Liao M, Lu P, et~al (2022{\natexlab{a}}) Reading and writing:
  Discriminative and generative modeling for self-supervised text recognition.
  In: Proceedings of the 30th ACM International Conference on Multimedia, pp
  4214--4223

\bibitem[{Yang et~al(2024)Yang, Yang, Liao, Zhu, and Bai}]{yang2024class}
Yang M, Yang B, Liao M, et~al (2024) Class-aware mask-guided feature refinement
  for scene text recognition. Pattern Recognition 149:110244

\bibitem[{Yang et~al(2021)Yang, Lu, Wang, Yin, Florencio, Wang, Zhang, Zhang,
  and Luo}]{yang2021tap}
Yang Z, Lu Y, Wang J, et~al (2021) Tap: Text-aware pre-training for text-vqa
  and text-caption. In: Proceedings of the IEEE/CVF conference on computer
  vision and pattern recognition, pp 8751--8761

\bibitem[{Yang et~al(2022{\natexlab{b}})Yang, Yu, He, Sun, Mao, and
  Mian}]{yang2022fully}
Yang Z, Yu H, He Y, et~al (2022{\natexlab{b}}) Fully convolutional
  network-based self-supervised learning for semantic segmentation. IEEE
  Transactions on Neural Networks and Learning Systems

\bibitem[{Yenduri et~al(2023)Yenduri, Srivastava, Maddikunta, Jhaveri, Wang,
  Vasilakos, Gadekallu et~al}]{yenduri2023generative}
Yenduri G, Srivastava G, Maddikunta PKR, et~al (2023) Generative pre-trained
  transformer: A comprehensive review on enabling technologies, potential
  applications, emerging challenges, and future directions. arXiv preprint
  arXiv:230510435

\bibitem[{Yousef et~al(2020)Yousef, Hussain, and Mohammed}]{yousef2020accurate}
Yousef M, Hussain KF, Mohammed US (2020) Accurate, data-efficient,
  unconstrained text recognition with convolutional neural networks. Pattern
  Recognition 108:107482

\bibitem[{Yu and Zhang(2021)}]{yu2021english}
Yu E, Zhang Z (2021) English billboard text recognition using deep learning.
  In: Journal of Physics: Conference Series, IOP Publishing, p 012003

\bibitem[{Yu et~al(2021)Yu, Li, Koh, Zhang, Pang, Qin, Ku, Xu, Baldridge, and
  Wu}]{yu2021vector}
Yu J, Li X, Koh JY, et~al (2021) Vector-quantized image modeling with improved
  vqgan. arXiv preprint arXiv:211004627

\bibitem[{Zbontar et~al(2021)Zbontar, Jing, Misra, LeCun, and
  Deny}]{zbontar2021barlow}
Zbontar J, Jing L, Misra I, et~al (2021) Barlow twins: Self-supervised learning
  via redundancy reduction. In: International conference on machine learning,
  PMLR, pp 12310--12320

\bibitem[{Zhang et~al(2020)Zhang, Ding, Peng, Fu, and Wang}]{zhang2020street}
Zhang C, Ding W, Peng G, et~al (2020) Street view text recognition with deep
  learning for urban scene understanding in intelligent transportation systems.
  IEEE Transactions on Intelligent Transportation Systems 22(7):4727--4743

\bibitem[{Zhang et~al(2022{\natexlab{a}})Zhang, Zhang, Song, Yi, Zhang, and
  Kweon}]{zhang2022survey}
Zhang C, Zhang C, Song J, et~al (2022{\natexlab{a}}) A survey on masked
  autoencoder for self-supervised learning in vision and beyond. arXiv preprint
  arXiv:220800173

\bibitem[{Zhang et~al(2022{\natexlab{b}})Zhang, Zhang, Pham, Niu, Qiao, Yoo,
  and Kweon}]{zhang2022dual}
Zhang C, Zhang K, Pham TX, et~al (2022{\natexlab{b}}) Dual temperature helps
  contrastive learning without many negative samples: Towards understanding and
  simplifying moco. In: Proceedings of the IEEE/CVF conference on computer
  vision and pattern recognition, pp 14441--14450

\bibitem[{Zhang et~al(2022{\natexlab{c}})Zhang, Zhang, Zhang, Pham, Yoo, and
  Kweon}]{zhang2022does}
Zhang C, Zhang K, Zhang C, et~al (2022{\natexlab{c}}) How does simsiam avoid
  collapse without negative samples? a unified understanding with
  self-supervised contrastive learning. arXiv preprint arXiv:220316262

\bibitem[{Zhang et~al(2021)Zhang, Nan, Wei, Li, Zhu, McKeown, Nallapati,
  Arnold, and Xiang}]{zhang2021supporting}
Zhang D, Nan F, Wei X, et~al (2021) Supporting clustering with contrastive
  learning. arXiv preprint arXiv:210312953

\bibitem[{Zhang et~al(2023)Zhang, Lin, Xu, Chen, and
  Zhang}]{zhang2023relational}
Zhang J, Lin T, Xu Y, et~al (2023) Relational contrastive learning for scene
  text recognition. In: Proceedings of the 31st ACM International Conference on
  Multimedia, pp 5764--5775

\bibitem[{Zhang et~al(2019{\natexlab{a}})Zhang, Qi, Wang, and
  Luo}]{zhang2019aet}
Zhang L, Qi GJ, Wang L, et~al (2019{\natexlab{a}}) Aet vs. aed: Unsupervised
  representation learning by auto-encoding transformations rather than data.
  In: Proceedings of the IEEE/CVF Conference on Computer Vision and Pattern
  Recognition, pp 2547--2555

\bibitem[{Zhang et~al(2016)Zhang, Isola, and Efros}]{zhang2016colorful}
Zhang R, Isola P, Efros AA (2016) Colorful image colorization. In: Computer
  Vision--ECCV 2016: 14th European Conference, Amsterdam, The Netherlands,
  October 11-14, 2016, Proceedings, Part III 14, Springer, pp 649--666

\bibitem[{Zhang et~al(2019{\natexlab{b}})Zhang, Zhou, Jiang, Song, Li, Zhou,
  Wang, Wang, Liao, Yang et~al}]{zhang2019icdar}
Zhang R, Zhou Y, Jiang Q, et~al (2019{\natexlab{b}}) Icdar 2019 robust reading
  challenge on reading chinese text on signboard. In: 2019 international
  conference on document analysis and recognition (ICDAR), IEEE, pp 1577--1581

\bibitem[{Zhang et~al(2022{\natexlab{d}})Zhang, Wang, Jin, Ren, and
  Xue}]{zhang2022cmt}
Zhang X, Wang J, Jin L, et~al (2022{\natexlab{d}}) Cmt-co: Contrastive learning
  with character movement task for handwritten text recognition. In:
  Proceedings of the Asian Conference on Computer Vision, pp 3104--3120

\bibitem[{Zhang et~al(2022{\natexlab{e}})Zhang, Wang, Wang, Jin, Luo, and
  Xue}]{zhang2022chaco}
Zhang X, Wang T, Wang J, et~al (2022{\natexlab{e}}) Chaco: character
  contrastive learning for handwritten text recognition. In: International
  Conference on Frontiers in Handwriting Recognition, Springer, pp 345--359

\bibitem[{Zhang et~al(2022{\natexlab{f}})Zhang, Zhu, Yao, Sun, Li, and
  Yu}]{zhang2022context}
Zhang X, Zhu B, Yao X, et~al (2022{\natexlab{f}}) Context-based contrastive
  learning for scene text recognition. In: Proceedings of the AAAI Conference
  on Artificial Intelligence, pp 3353--3361

\bibitem[{Zhang et~al(2017)Zhang, Gueguen, Zharkov, Zhang, Seifert, and
  Kadlec}]{zhang2017uber}
Zhang Y, Gueguen L, Zharkov I, et~al (2017) Uber-text: A large-scale dataset
  for optical character recognition from street-level imagery. In: SUNw: Scene
  Understanding Workshop-CVPR, p~5

\bibitem[{Zhang et~al(2019{\natexlab{c}})Zhang, Nie, Liu, Xu, Zhang, and
  Shen}]{zhang2019sequence}
Zhang Y, Nie S, Liu W, et~al (2019{\natexlab{c}}) Sequence-to-sequence domain
  adaptation network for robust text image recognition. In: Proceedings of the
  IEEE/CVF conference on computer vision and pattern recognition, pp 2740--2749

\bibitem[{Zheng et~al(2021)Zheng, You, Wang, Qian, Zhang, Wang, and
  Xu}]{zheng2021ressl}
Zheng M, You S, Wang F, et~al (2021) Ressl: Relational self-supervised learning
  with weak augmentation. Advances in Neural Information Processing Systems
  34:2543--2555

\bibitem[{Zhou et~al(2023)Zhou, Li, Wang, Chen, and Luo}]{zhou2023novel}
Zhou J, Li G, Wang R, et~al (2023) A novel contrastive self-supervised learning
  framework for solving data imbalance in solder joint defect detection.
  Entropy 25(2):268

\bibitem[{Zhou et~al(2022)Zhou, Liu, Qiao, Xiang, and Loy}]{zhou2022domain}
Zhou K, Liu Z, Qiao Y, et~al (2022) Domain generalization: A survey. IEEE
  Transactions on Pattern Analysis and Machine Intelligence 45(4):4396--4415

\bibitem[{Zhuang et~al(2022)Zhuang, Ren, Li, and Liang}]{zhuang2022text}
Zhuang J, Ren Y, Li X, et~al (2022) Text-level contrastive learning for scene
  text recognition. In: 2022 International Conference on Asian Language
  Processing (IALP), IEEE, pp 231--236

\end{thebibliography}

\end{document}